\documentclass[english,journal]{IEEEtran}

\usepackage{graphicx}                        %
\usepackage{evolvingai}                     %
\usepackage{textcomp}                        %
\usepackage{gensymb}                         %
\usepackage[numbers, sort&compress]{natbib}  %
\usepackage[percent]{overpic}                %
\usepackage{url}                             %
\usepackage[T1]{fontenc}                     %
\usepackage{amssymb}
\usepackage{amsmath}
\usepackage{amsfonts}
\usepackage{babel}
\usepackage{calc}                            %
\usepackage{setspace}                        %
\usepackage{siunitx}                         %
\usepackage{numprint}
\usepackage{afterpage}
\npthousandsep{,}

\addeditor{jh}{olive} 

\newcommand{\myext}[1]{#1.pdf}

\newcommand{\myfig}[2]{\includegraphics[width=#1\textwidth]{\myext{#2}}}

\newcommand{\labfigo}[4]{%
\begin{overpic}[width=#2\textwidth]{#4}%
\put(-1,#3){(#1)}%
\end{overpic}}

\newcommand{\toplabel}[1]{\begin{minipage}[t]{13pt}\vspace{0pt}#1\end{minipage}}

\newcommand{\topfig}[3]{
\begin{minipage}[t]{#1\textwidth}
\vspace{0pt} \centering
\myfig{#2}{#3}
\end{minipage}}

\newcommand{\botfig}[4]{
\begin{minipage}[t][#2][b]{#1\textwidth}
\vspace{0pt} \centering
\myfig{#3}{#4}
\end{minipage}}

\newcommand{\topfigcap}[4]{
\begin{minipage}[t]{#1\textwidth}
\vspace{0pt} \centering
\myfig{#2}{#3}\\
#4
\end{minipage}}

\newcommand{\labfig}[4]{\toplabel{(#1)}\topfig{#2}{#3}{#4}}

\begin{document}

\newcommand{\papertitle}{Evolving Multimodal Robot Behavior via Many Stepping Stones with the Combinatorial Multi-Objective Evolutionary Algorithm}
\title{\papertitle}

\author{Joost Huizinga and Jeff Clune%
\thanks{J. Huizinga and J. Clune are with the Evolving Artificial Intelligence Laboratory, University of Wyoming, Laramie,
WY, 82071 USA and Uber AI Labs, San Francisco, CA, 94104 USA e-mail: jeffclune@uwyo.edu.}}

\maketitle

\begin{abstract}
An important challenge in reinforcement learning is to solve multimodal problems, where agents have to act in qualitatively different ways depending on the circumstances.
Because multimodal problems are often too difficult to solve directly, it is often helpful to define a curriculum, which is an ordered set of sub-tasks that can serve as the stepping stones for solving the overall problem. 
Unfortunately, choosing an effective ordering for these subtasks is difficult, and a poor ordering can reduce the performance of the learning process. 
Here, we provide a thorough introduction and investigation of the Combinatorial Multi-Objective Evolutionary Algorithm (CMOEA), which allows all combinations of subtasks to be explored simultaneously. 
We compare CMOEA against three algorithms that can similarly optimize on multiple subtasks simultaneously: NSGA-II, NSGA-III and $\epsilon$-Lexicase Selection.
The algorithms are tested on a simulated multimodal robot locomotion problem with six subtasks as well as a simulated robot maze navigation problem with a hundred subtasks. 
On these problems, CMOEA either outperforms or is competitive with the controls. 
As a separate contribution, we show that adding a linear combination over all objectives can improve the ability of the control algorithms to solve these multimodal problems.
Lastly, we show that CMOEA can leverage auxiliary objectives more effectively than the controls on the multimodal locomotion task.
In general, our experiments suggest that CMOEA is a promising algorithm for solving multimodal problems.
\end{abstract}


\section{Introduction}

A pervasive challenge in reinforcement learning
is to have agents autonomously learn many qualitatively different behaviors, 
generally referred to as \emph{multimodal behavior}~\cite{li2014evolving, schrum2014evolving}.
Problems that require such multimodal behavior, which we will refer to as \emph{multimodal problems} (also known as \emph{modal problems}~\cite{spector2012assessment}), are ubiquitous in real-world applications. A self-driving car will have to respond differently depending on whether it is on a highway, in a city, in a rural area, or in a traffic jam. A robot on a search-and-rescue operation will have to behave differently depending on whether it is searching for a victim or bringing a survivor to safety. Even a simple trash-collecting robot will have to behave differently depending on whether it is searching for trash, picking it up, or looking for a place to recharge.

Because multimodal problems require an agent to learn multiple different behaviors, they can be difficult to solve directly. 
A key insight for solving multimodal problems comes from how natural animals, including humans, learn complex tasks. 
Rather than learning all aspects of the combined task at once, we learn simpler, related tasks first. Later, the skills learned in these earlier tasks can be combined and adjusted in order to learn the more complex task at hand. These related tasks thus form the \emph{stepping stones} towards solving the complete task. Methods that incrementally increase the difficulty of tasks have been successfully applied in animal training~\cite{skinner1958reinforcement, peterson2004day}, gradient-descent based machine learning~\cite{elman1993learning, bengio2009curriculum}, and evolutionary algorithms~\cite{gomez1997incremental, lewis1992genetic, lewis1993genetic, harvey1994seeing, larsen2005evolving, lessin2013open,mouret2008incremental}. 
Unfortunately, defining a proper set of subtasks and an effective ordering is a non-trivial problem, and choosing poor subtasks or presenting them in the wrong order can severely reduce the effectiveness of the learning process~\cite{bongard2008behavior, auerbach2009robot}.

\begin{figure*}[tbp!]
\centering
\includegraphics[width=0.7\textwidth]{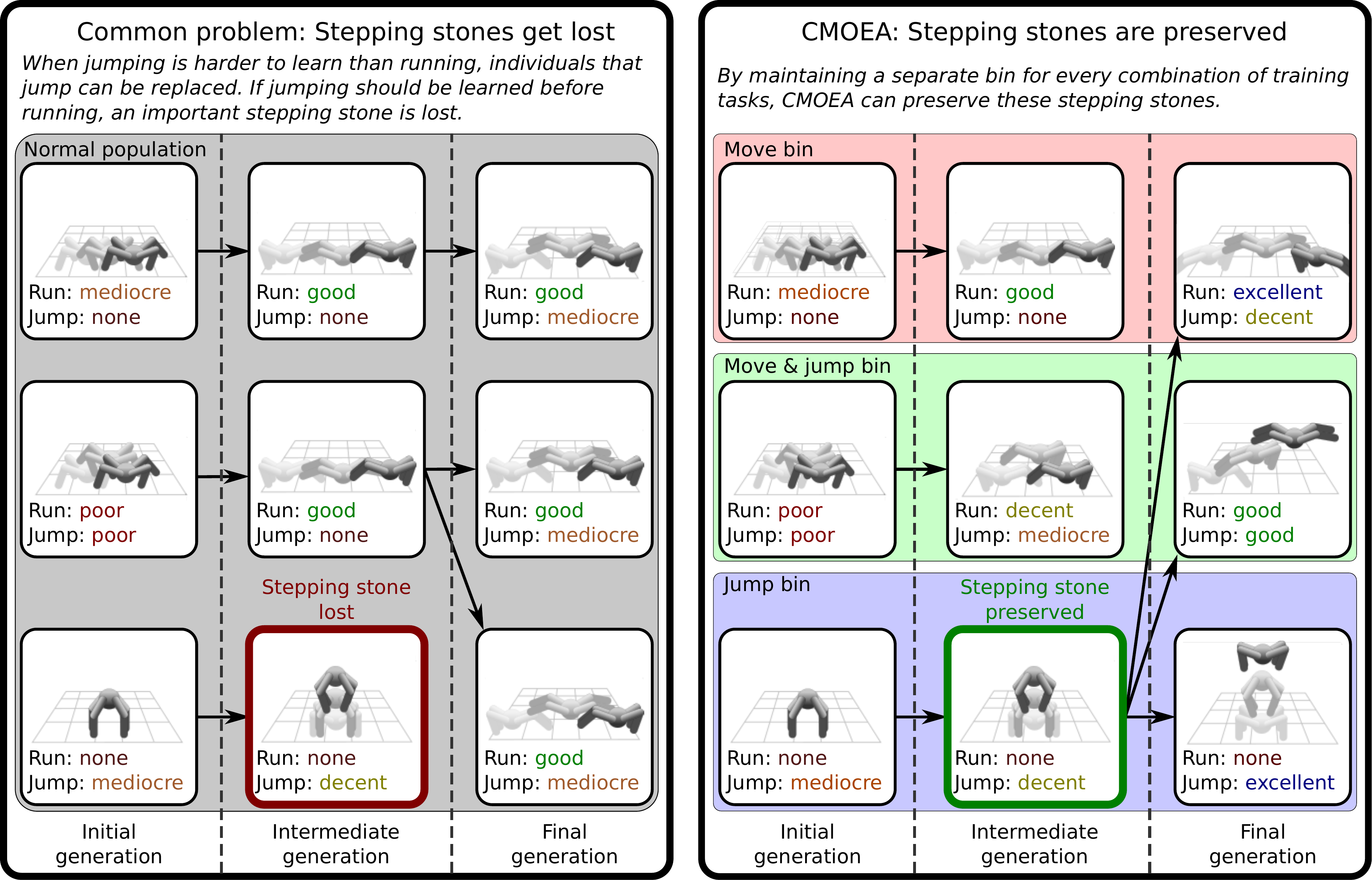}
\caption{\textbf{CMOEA can preserve stepping stones that may be lost in other EAs.} In this hypothetical example, a four-legged robot has to learn multimodal behavior that involves both running and jumping. Running is initially much easier to learn than jumping, but learning to jump well first is an important stepping stone in order to become excellent at both tasks. Arrows indicate ancestor-descendant relationships that can span many generations. 
\textbf{(Left)} Example of losing an important stepping stone. Initial generation: Some individuals are better at running while others are better at jumping, but all individuals are evaluated roughly equally by the fitness function. Intermediate generations: Because running is easier to learn than jumping, individuals that are good at running are rated more favorably than individuals that are average at jumping, and those specialized in jumping are not selected for in future generations. Final generation: All individuals have converged to the same local optima, where they are good at running, but only mediocre at jumping.
\textbf{(Right)} Example of how CMOEA can preserve important stepping stones. Initial generation: Individuals that specialize in different combinations of tasks are assigned to different bins. Intermediate generations: Individuals that are average at jumping do not compete against individuals that are good at running and they are thus preserved within the population. Final generation: Because jumping turned out to be an important stepping stone, the descendants of individuals that initially specialized in jumping have increased performance on all combinations of tasks.
}
\label{fig:concept}
\end{figure*}

Population-based \emph{Evolutionary Algorithms} (EAs) may provide a unique opportunity to combat the problem of choosing and ordering subtasks, because the population as a whole can try many different ways of learning the subtasks and evolutionary selection can preserve the methods that work.
For example, imagine training a robot that has to be able to both run and jump, 
but in order to learn both tasks, it is imperative that it learns how to jump first before it learns how to run.
In an evolutionary algorithm, one lineage of robots may start out being better at running,
while another lineage may initially be better at jumping.
If learning to jump first is an essential stepping stone towards learning to both run and jump,
the lineage that started by being good at running will never learn both tasks, 
but the lineage that started by being good at jumping will, thus solving the problem.
However, without proper tuning, most evolutionary algorithms are prone to converge towards the task that is easiest to learn at first, as learning this task will result in the most rapid increase in fitness. 
For example, if learning to run is much easier than learning to jump, the lineage specialized in running may outcompete the lineage of those specialized in jumping before they have the chance to adapt (Fig.~\ref{fig:concept} left).
As these jumping individuals were an important stepping stone for learning how to both run and jump, this stepping stone is now lost from the population, and because the population is now dominated by runners, it is unlikely that the jumping-only behavior will ever be visited again.

It is important to note that we usually have no way of knowing in advance what the important stepping stones are~\cite{woolley2011deleterious, nguyen2015innovation, wang2019poet}.
As such, one of the best ways of preserving stepping stones may be to maintain as many forms of different behavior in the population as possible.
Here we introduce
the Combinatorial Multi-Objective Evolutionary Algorithm (CMOEA)~\cite{huizinga2016aligning},
a multiobjective evolutionary algorithm specifically designed to preserve the stepping stones of multimodal problems. 
CMOEA was briefly described before in~\citet{huizinga2016aligning}, but here we provide a more thorough description of the algorithm, an extension that allows the algorithm to be applied to at least 100 objectives, and a more detailed experimental investigation that includes three different controls and one additional problem domain.
CMOEA divides the population into separate bins, one for each combination of subtasks, and ensures that there is only competition within bins, rather than between bins. This way, individuals that excel at any combination of subtasks are preserved as potential stepping stones for solving the overall problem (Fig.~\ref{fig:concept}, right). 

We compare CMOEA against three multiobjective evolutionary algorithms: the widely applied NSGA-II algorithm~\cite{deb2002nsga}, the more recent NSGA-III algorithm~\cite{deb2013evolutionary}, and the $\epsilon$-Lexicase Selection algorithm, which was also specifically designed to solve multimodal problems~\cite{la2016epsilon} and can handle problems with at least 100 subtasks~\cite{moore2017lexicase}. 
We compare the algorithms on both a simulated multimodal robot locomotion problem with 6 subtasks and on a simulated robot maze-navigation problem with 100 subtasks, and show that CMOEA either outperforms or is competitive with the control treatments.
Note that, while both presented problems involve a robotic agent, CMOEA is not restricted to robotics problems. Tasks that require multimodal behavior can also be found in other domains, such as video games~\cite{schrum2014evolving, li2014evolving} and function approximation~\cite{spector2012assessment}, and CMOEA should be similarly applicable in those domains. In fact, the maze navigation problem presented in this paper is very similar to navigation problems presented in video-game and grid-world domains~\cite{justesen2018illuminating, sutton1990integrated}, providing some evidence that CMOEA is applicable in those domains as well.
As a separate contribution, we show that adding a linear combination over all objectives as an additional objective to 
the control algorithms (which are popular in their own right) can improve their ability to solve multimodal problems.
Lastly, we demonstrate that CMOEA is able to effectively incorporate auxiliary objectives 
that increase the evolvability of individuals by selecting for genotypic and phenotypic modularity.
With these auxiliary objectives, CMOEA substantially outperforms all controls on the simulated multimodal robot locomotion task, 
while the controls do not benefit as much or even perform worse when these auxiliary objectives are added.
These results indicate that CMOEA is a promising, state-of-the-art algorithm for solving multimodal problems.

\section{Background}
\label{sec:background}

Because multimodal problems are ubiquitous in many practical applications, a wide range of strategies have been developed for solving them.
Many methods are based on the idea of \emph{incremental evolution}, where complex problems are solved incrementally, one step at a time~\cite{gomez1997incremental}. 

One incremental method is \emph{staged evolution}, where the evolutionary process is divided into separate stages, each with its own objective function~\cite{lewis1993genetic, harvey1994seeing}. The process starts in the first, easiest stage, and the population is moved to the next, more difficult stage when the first stage is considered ``solved'' according to stage-specific success criteria. Staged evolution requires each stage, the order in which the stages are presented, and the success criteria to be defined manually, but the design of the controller being evolved can be fully determined by evolution. 
Staged evolution can be made more smooth if the environment has parameters that allow for fine-grained adjustments, such as the speed of the prey in a predator-prey simulation~\cite{gomez1997incremental}. Such fine-grained staging has also been referred to as \emph{environmental complexification}~\cite{mouret2008incremental}.
Closely related to staged evolution is \emph{fitness shaping}, where the fitness function is either dynamically or statically shaped to provide a smooth gradient towards a final goal~\cite{nolfi1995evolving2, schrum2010evolving}.
Fitness shaping provides the benefits of staging without the need to define the stages explicitly, but it does increase the complexity of the fitness function, which needs to be carefully designed by hand.

Another successful incremental-evolution method is \emph{behavioral decomposition}~\cite{mouret2008incremental}, 
where a separate controller is optimized for each task
and higher level controllers can build upon lower level controllers~\cite{larsen2005evolving, lessin2013open},
similar to hierarchical reinforcement learning~\cite{barto2003recent}.
Variants of behavioral decomposition techniques frame the process as a cooperative multiagent system, 
where each agent optimizes the controller for a particular subtask,
such that a group of these agents can combine their controllers and solve the task as a whole~\cite{pieter2004evolutionary, agogino2008efficient}.
One downside of these behavioral decomposition methods is that it becomes the experimenters responsibility to decide which tasks should get their own controller, 
which controllers build upon which other controllers, and in what order controllers should be trained, 
all of which are difficult decisions that can severely impact the effectiveness of the method.
In addition, because the controllers are optimized separately, there is no opportunity for the optimization process to reuse information between controllers or find atomic controllers that work well together, meaning computational time may be wasted due to having to reinvent the same partial solutions in several different controllers.

A completely different way of approaching multimodal problems is to focus on \emph{behavioral diversity}~\cite{mouret2009overcoming}. 
Behavioral diversity based approaches attempt to promote intermediate stepping stones
by rewarding individuals for being different from the rest of the population. 
As a result, a population will naturally diverge towards many different behaviors, 
each of which may be a stepping stone towards the actually desired behavior.
One canonical example of an algorithm based on behavioral diversity is Novelty Search,
where individuals are selected purely based on how different their behaviors are compared to an archive of individuals from previous generations~\cite{lehman2011abandoning}.
Novelty Search has been shown to be effective in maze navigation and biped locomotion problems~\cite{lehman2011abandoning},
though it is unclear whether these problems required multimodal behavior.
By selecting for behavioral diversity and performance, 
\citet{Mouret2012} were able to evolve robots that would exhibit a form of multimodal behavior
where a robot would alternate between searching for a ball and depositing it in a predefined goal location.
The main drawback of behavioral diversity based approaches is that the space of possible behaviors can be massive, 
meaning that it may contain many ``uninteresting'' behaviors that neither resemble a potential solution nor represent a stepping stone towards any other relevant behavior.
One method for avoiding the problem of having too many ``uninteresting'' solutions is by having a fixed number of behavioral niches~\cite{cully2015robots, nguyen2015innovation}. 
By discretizing the behavior space, \citet{cully2015robots} were able to evolve many different modes of behavior for a hexapod robot,
which formed the basis of an intelligent trial-and-error algorithm that enabled the robot to quickly respond to damage.
Similarly, 
\citet{nguyen2015innovation} were able to evolve a wide range of different looking images by having separate niches for images assigned to different categories by a pre-trained neural network, an algorithm called the \emph{Innovation Engine}. 
CMOEA also builds on the idea of having different niches for different types of solutions,
but instead of defining its niches based on different behaviors or different classes,
it defines its niches based on different combinations of subtasks.

The strategy of solving multimodal problems considered in this paper revolves around framing it as a multiobjective problem, 
where each training task is its own objective~\cite{mouret2008incremental, schrum2010evolving, spector2012assessment}.
We will refer to this strategy as \emph{multiobjective incremental evolution} and, as with staged evolution, 
it requires the problem to be decomposed into a set of subtasks, each with its own fitness function. 
However, in contrast to staged evolution, the subtasks do not have to be explicitly ordered
and there is no need to explicitly define success criteria for each stage.

There are many ways to obtain an appropriate set of subtasks. 
For example, it is possible to use prior knowledge in order to define a separate training task for every mode of behavior that might be relevant for solving the overall problem, such as having separate subtasks for moving and jumping. 
Alternatively, it is also possible to generate different environments and have each environment be its own training task. Provided that the environments are diverse enough (e.g. some environments include objects that need to be jumped over while other environments feature flat ground that needs to be traversed quickly) they can similarly encourage different modes of behavior.
Subtasks could even involve different problem domains, such as image classification for one task and robot locomotion for another task.
The main idea is that, as long as a task is unique and somewhat related to the overall problem, it can be added as an objective to promote multimodal behavior.

That said, while it is relatively straightforward to split a multimodal problem into different subtasks, there is no guarantee that classic multiobjective algorithms will perform well on this set. 
The main reason is that the set of subtasks will often be much larger than the number of objectives generally solved by multiobjective algorithms;
rather than a multiobjective problem, which generally refers to problems with three or fewer objectives~\cite{khare2003performance, deb2005finding, ishibuchi2008evolutionary, wagner2007pareto, fleming2005many}, it becomes a \emph{many-objective problem}, a term coined for problems that require optimization of many more objectives~\cite{deb2005finding, ishibuchi2008evolutionary, wagner2007pareto, fleming2005many}.
For example, the maze navigation problem presented in this paper required at least 100 subtasks in order to promote general maze solving behavior (preliminary experiments with 10 subtasks generalized poorly and even with 100 training mazes generalization is not perfect, SI Sec.~\ref{sec:number_training_mazes}).
Many popular multiobjective algorithms have trouble with such a large number of objectives because they are based on the principle of \emph{Pareto dominance}~\cite{khare2003performance, deb2005finding, ishibuchi2008evolutionary, wagner2007pareto, fleming2005many}. 
According to the definition of Pareto-dominance, an individual $A$ dominates an individual $B$ only if $A$ is not worse on any objective than $B$ and $A$ is better than $B$ on at least one objective~\cite{deb2001multi}.
With the help of this Pareto-dominance relation, these algorithms attempt to approximate the true \emph{Pareto front},
the set of solutions which are non-dominated with respect to all other possible solutions.
However, as the number of objectives grows, the number of individuals in a population that are likely to be non-dominated increases exponentially.
When nearly all individuals in a population are non-dominated, 
a Pareto-dominance based algorithm may lose its ability to apply adequate selection pressure.
There exist many different methods to increase the maximum number of objectives that these Pareto-based algorithms can handle~\cite{deb2012handling, laumanns2002combining, deb2005finding},
and these methods have been shown to be effective up to $10$ objectives if no assumptions about the problem are made~\cite{deb2012handling},
and up to $30$ objectives if the majority of those objectives are redundant~\cite{deb2005finding}.
However, because of the exponential relationship between the number of objectives and the dimensionality of the Pareto front,
it is unlikely that purely Pareto-based methods will be able to scale much further.

There also exist many multiobjective evolutionary algorithms that do not rely on Pareto dominance~\cite{deb2001multi, bentley1998finding, spector2012assessment, drechsler2001multi}.
Such techniques may be especially relevant for multimodal problems because multimodal problems
do not necessarily require an approximation of the true Pareto front.
Instead, multimodal problems simply require adequate performance on all objectives,
which generally means searching for only a small area or point on the true Pareto front.
In theory, searching for a point on a Pareto front can be achieved by simply optimizing a weighted sum of all objectives~\cite{deb2001multi}.
The main problem with such a weighted sum approach is that, even when the desired tradeoff for the optimal solution is known,
the trajectory for finding this optimal solution may not be a straight line,
but may instead require the algorithm to find a number of solutions with different tradeoffs first~\cite{deb2001multi}.
This issue will almost certainly be present in the context of multimodal problems
because different modes of behavior can vary greatly in difficulty,
meaning that ``straight line'' optimization (i.e. attempting to learn all modes simultaneously) is likely to fail. 
As such, multimodal problems may be best tackled by algorithms that do not strictly rely on Pareto dominance,
but that still explore many different tradeoffs during optimization.

\section{Treatments}

\subsection{CMOEA}

\begin{figure}[tb!]
\centering
\includegraphics[width=0.49\textwidth]{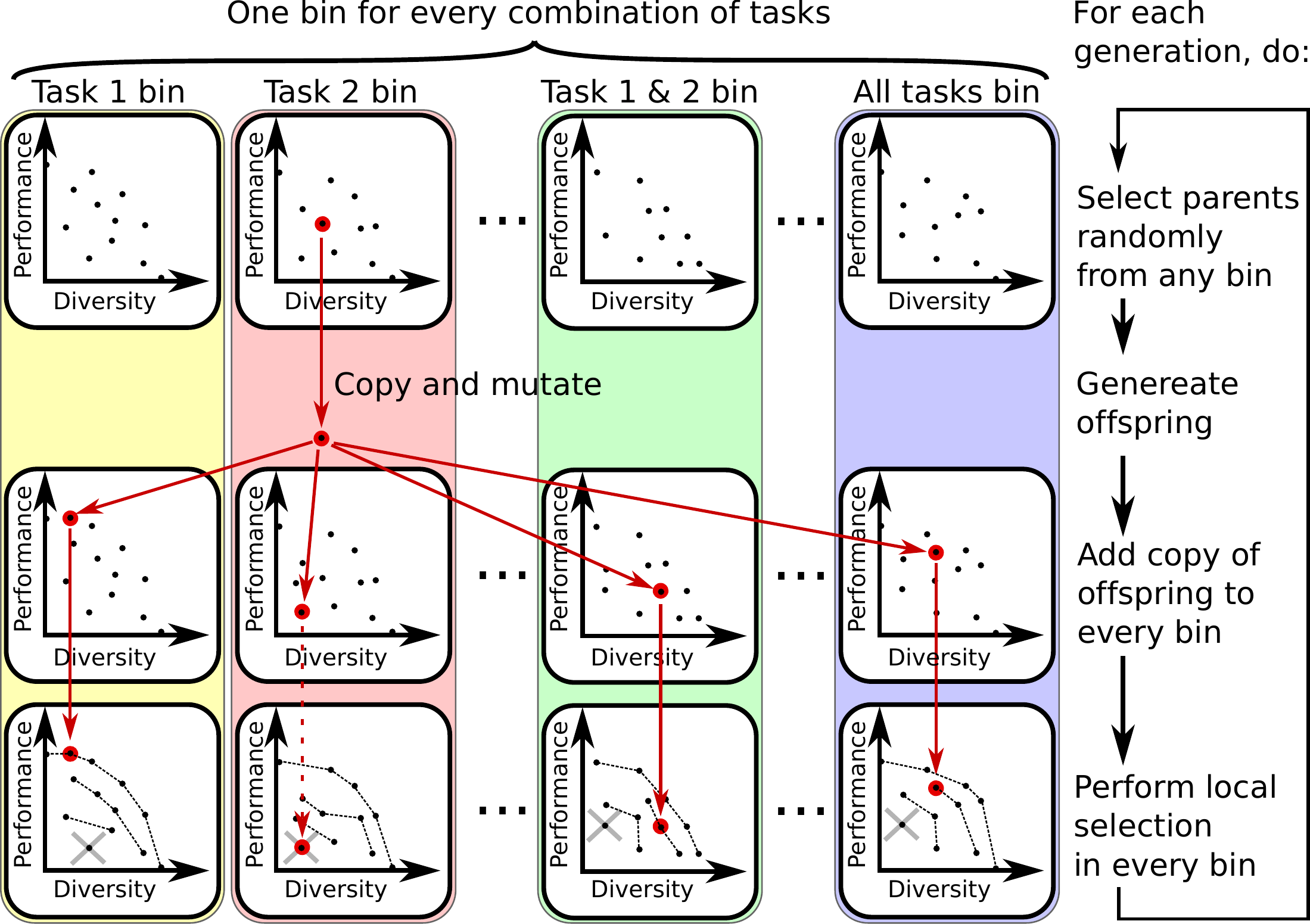}
\caption{\textbf{Overview of CMOEA.} At every generation, CMOEA first selects a number of parents ($1$ in this example) randomly from across all bins. It then creates offspring by copying and mutating those parents and one copy of each offspring is added to every bin. Afterward, a local survivor-selection method determines which individuals remain in each bin. In this example, survivor selection is performed by the non-dominated sorting algorithm from NSGA-II, with performance on the tasks associated with the relevant bin as one objective and behavioral diversity within the bin as the other objective.}
\label{fig:cmoea}
\end{figure}

The goal of CMOEA is to provide a large number of potential evolutionary stepping stones,
thus increasing the probability that some of these stepping stones are on the path to solving the task as a whole.
To do so, we define a \textit{bin} for every combination of subtasks of our problem.
For example, if we have the two subtasks of moving forward and moving backward, there will be one bin for moving forward, 
one bin for moving backward, and one bin for the combination of moving forward and backward.
The algorithm starts by generating and evaluating a predetermined number of random individuals and 
adding a copy of each generated individual to every bin. 
Next, survivor selection is performed within each bin such that, afterward, each bin contains a number of individuals equal to some predetermined bin size.
For each bin, selection happens only with respect to the subtasks associated with that bin.
After this initialization procedure, the algorithm will perform the following steps at each generation: 
(1) select a number of parents randomly from across all bins, 
(2) generate one child for each selected parent by copying and mutating that parent (no crossover is performed in the version presented in this paper), 
(3) add a copy of each child to every bin, and 
(4) perform survivor selection within each bin (Fig.~\ref{fig:cmoea}).

For survivor selection to work on a bin with multiple subtasks,
we need some way of comparing individuals who may have different performance values on each of these tasks.
While there exist many selection procedures specifically designed to work with multiple objectives~\cite{fonseca1993genetic, horn1994niched, srinivas1994muiltiobjective, deb2001multi},
these multiobjective selection procedures tend to have difficulty with many objectives (see Sec.~\ref{sec:background}),
which is exactly the problem CMOEA was designed to solve.
As such, within bins, CMOEA combines performance values on different tasks into a single fitness value by taking their arithmetic mean or by multiplying them.
Multiplication can be more effective than taking the arithmetic mean because it requires individuals to obtain at least some non-zero performance on every relevant subtask within a bin, rather than being able to specialize in a subset of those subtasks while neglecting the others.
Note that the same properties can be obtained by taking the geometric mean,
but multiplication is computationally more efficient to calculate
and both methods result in the same relative ordering.
Multiplication does require clipping or shifting values in a way that avoids negatives,
as negative values could completely alter the meaning of the combined performance metric.
That said, it is generally considered good practice to normalize performance values
regardless of whether values are combined by taking the arithmetic mean or through multiplication,
as overly large or overly negative values can negatively impact the effectiveness of both aggregation methods.

While CMOEA does not prescribe any particular selection procedure for the survivor selection step within each bin,
we implement the multiobjective \emph{behavioral diversity} method by~\citet{mouret2009overcoming}.
In this method, the multiobjective evolutionary algorithm NSGA-II~\cite{deb2002nsga} selects for both performance and behavioral diversity,
which allows it to avoid local optima and fitness plateaus~\cite{mouret2009overcoming}.
We apply it as a within-bin selection procedure because it ensures that each bin maintains individuals that solve the same subtasks in different ways.
In addition, this method outperformed a method based on novelty search with local competition~\cite{lehman2011evolving}, which is another algorithm that optimizes for both performance and diversity (SI~\ref{sec:bin_selection}).
For any particular bin, the performance objective is the main objective associated with that bin (e.g. move forward, move backward, move forward $\times$ move backward, etc.).
The behavioral diversity of an individual is calculated by first measuring some relevant feature of the behavior of an individual, 
called the \emph{behavior descriptor}, and then calculating the mean distance to the behavior descriptors of all other individuals in the same bin. 
As such, the larger this distance, the more unique the behavior of the individual is with respect to the other individuals in that bin.
Behavioral diversity metrics differ per domain, and details can be found in sections~\ref{sec:robot} and~\ref{sec:maze}.

Maintaining a bin for every single combination subtasks may not be feasible on problems with many subtasks. For example, in the simulated robot maze navigation problem we define $100$ subtasks (see section~\ref{sec:maze_setup} for details), which would result in $2^{100} - 1 = \num{1.27e 30}$ different bins. While it may seem that such exponential scaling would prevent CMOEA from being applied to larger problems, 
it is important to note that what we really require from CMOEA is that it provides a sufficiently large number of different stepping stones.
As long as there is a sufficiently large number of directions in which improvements can be discovered, 
evolution is unlikely to get stuck in local optima, and can thus continue to make progress. 
As such, if the number of bins is large, only a subset of bins may be necessary to provide the required stepping stones.
One method for constructing such a subset, demonstrated in our maze navigation experiment (section~\ref{sec:maze_setup}), starts by including the bin for each individual subtask as well as the bin for the combination of all subtasks, and then adds random bins to the subset until a desired number of bins is reached.
The code for CMOEA, as well as for all experiments and control treatments, is available at: \url{www.evolvingai.org/cmoea}.

To assess the performance of CMOEA relative to other algorithms, we compare CMOEA against three successful multiobjective algorithms, namely NSGA-II~\cite{deb2002nsga}, NSGA-III~\cite{deb2013evolutionary} and $\epsilon$-Lexicase Selection~\cite{la2016epsilon}. To verify the usefulness of having many CMOEA bins, we also compare CMOEA against a variant that only has a single bin with all of the subtasks. A description of each of these treatments is provided below.

\subsection{Single Bin CMOEA}
\label{sec:controls}

To verify that having many CMOEA bins actually provides a practical benefit,
we run a control that features only a single CMOEA bin, called the Single Bin treatment.
The Single Bin treatment is the same as CMOEA, except that it only has one bin, namely the bin that is associated with all subtasks.
To ensure a fair comparison, this bin is resized such that the number of individuals within this bin is equal to the total number of individuals maintained by CMOEA across all bins
(the Single Bin treatment with a population size equal to the number of new individuals created at each generation (\numprint{1000}),
which is a common default in EAs, performed worse, see SI Fig.~\ref{fig:popsize_comp}).
In addition, the Single Bin treatment also implements the Pareto-based tournament selection procedure for parent selection from NSGA-II, 
which is expected to increase the performance of the Single Bin treatment, and thus ensures a comparison against the best possible implementation of this treatment~\cite{deb2002nsga}.

\subsection{NSGA-II}

NSGA-II~\cite{deb2002nsga} is a Pareto-based multiobjective evolutionary algorithm that, because of its popularity~\cite{deb2002nsga, huizinga2016aligning,schrum2010evolving,khare2003performance,mouret2009overcoming,Mouret2012,deb2005finding,
ishibuchi2008evolutionary,wagner2007pareto,deb2012handling,
laumanns2002combining,clune2013originModularity,mengistu2016evolutionary,ellefsen2015neural}, functions as a recognizable benchmark.
Briefly (see~\citet{deb2002nsga} for details), NSGA-II works by sorting a mixed population of parents and children into ranked fronts where each individual is non-dominated with respect to all individuals in the same and lower ranked fronts. 
During selection, NSGA-II iteratively adds individuals, starting from the highest ranked front and moving towards the lowest ranked front, 
until a sufficient number of individuals have been selected to populate the next generation.
The last front in this process generally can not be added to the next generation in its entirety, meaning that a tie-breaker is necessary. The tie-breaker in NSGA-II is a crowding score that prefers individuals whose neighbors on each objective are far apart from each other~\cite{deb2002nsga}.

\subsection{NSGA-III}

NSGA-III differs from NSGA-II only in the tie-breaker that determines which individuals are selected from the last front that is copied to the next generation. While this is a small change algorithmically,  the tie-breaking behavior can greatly affect performance on many-objective problems because, on these problems, the majority of the population may actually be on the same front, meaning the tie-breaker can be the most important selective pressure applied by the algorithm. The tie-breaking behavior of NSGA-III is described in detail in~\citet{deb2013evolutionary}, so here we describe it only briefly. In order to deal with differently scaled objectives, NSGA-III first normalizes all objective scores. It then defines a number of reference lines in this normalized space that are evenly distributed over different trade offs between objectives (e.g. with two objectives and three lines, the trade offs will be $[0, 1]$, $[.5, .5]$, and $[1, 0]$). It subsequently assigns each individual to the reference line that is closest to it. From there, NSGA-III first selects from the individuals that are closest to each line, never selecting the same individual twice. If that step did not provide sufficient individuals, it randomly selects an individual from a line that has thus far provided the fewest individuals for the next generation, ignoring lines that have no individuals associated with them (i.e. NSGA-III will alternate selecting one individual from each line).

We have implemented NSGA-III as described in~\citet{deb2013evolutionary}, with the exception that we normalize by the highest value found along each dimension of the objective space, rather than by the intercepts found based on the extreme points of the population. The reason for this change is that, in preliminary experiments, the extreme points 
often did not span the entire space, thus making it impossible to calculate the intercepts. In addition, always performing normalization by the highest value outperformed an alternative where we only normalize by highest value when the intercepts could not be calculated (SI~\ref{sec:nsga_iii_normalization}). Following the standard implementation, the number of reference lines was chosen to be as close as possible to the population size~\cite{deb2013evolutionary}. 

NSGA-III provides the possibility to add additional reference lines corresponding to trade offs of interest. In all experiments we added a reference line for the combination of all objectives, as this is the trade off of interest.

\subsection{$\epsilon$-Lexicase Selection}
\label{sec:e_lexicase_selection}

$\epsilon$-Lexicase Selection is a state-of-the-art algorithm for solving multimodal problems and it thus presents another good benchmark to compare against~\cite{la2016epsilon}.
The $\epsilon$-Lexicase Selection algorithm is an extension of the Lexicase Selection algorithm~\cite{spector2012assessment},
where each individual is selected by first choosing a random order for the objectives, 
and then selecting the individual that is the best according to the lexicographical ordering that results from the randomly ordered objectives
(e.g. if the random order of objectives is \{Forward, Backward\}, first all individuals that have the maximum performance on the Forward task will be selected 
and then the performance on the Backward task will serve as a tiebreaker).
Because the order of objectives is randomized for every individual being selected, 
Lexicase Selection will select specialists on each of the objectives first, 
with ties being broken by performance on other objectives.

While Lexicase Selection works well when subtask performance is measured in a discrete way (e.g. when performance is passing or failing a particular test case), it tends to break down when the subtask performance is measured in a continuous space, such as when performing regression. The reason for such failure is that the probability of having ties becomes close to zero in a continuous space, meaning Lexicase Selection will regress to selecting the individual that is best at the first objective in the random order, while ignoring all other objectives. As a result, Lexicase Selection on  continuous subtasks effectively selects one specialist for each subtask, but it will not produce solutions with some performance on all subtasks.

The $\epsilon$-Lexicase Selection algorithm resolves this issue by defining an $\epsilon$ margin such that a solution is considered tied on a particular objective if it scores within $\epsilon$ of a particular threshold on that objective. In~\cite{cava2018probabilistic} three versions of $\epsilon$-Lexicase Selection are presented: \emph{static $\epsilon$-Lexicase Selection}, \emph{semi-dynamic $\epsilon$-Lexicase Selection}, and \emph{dynamic $\epsilon$-Lexicase Selection}. In static $\epsilon$-Lexicase Selection, $\epsilon$ is either chosen in advance or calculated based on the Median Absolute Deviation (MAD) on the objective among the entire population and the threshold on each objective is either based on a preset maximum or equal to the highest score on that objective among the entire population. In semi-dynamic $\epsilon$-Lexicase Selection, $\epsilon$ is still static, but the threshold is now the highest score among the individuals that are still considered for selection (e.g. if three individuals were considered tied on the first objective, the threshold for the second objective will come from these three individuals, rather than from the entire population). Lastly, dynamic $\epsilon$-Lexicase Selection calculates both the threshold and $\epsilon$ based on the individuals still considered for selection, rather than across the entire population. To chose a version to compare against (including regular Lexicase Selection), we performed extensive preliminary experiments for each problem, and we chose the version and parameters that worked best in those preliminary experiments (SI~\ref{sec:lexicase_variants}). For the multimodal locomotion problem the best performing variant was semi-dynamic $\epsilon$-Lexicase Selection with a fixed $\epsilon=0.05$. For the simulated robot maze navigation problem most variants performed equally well, and we chose dynamic $\epsilon$-Lexicase Selection for comparison.

In previous work, $\epsilon$-Lexicase Selection was implemented as part of an evolutionary algorithm where evolution happens by selecting a number of parents equal to the population size, 
and then having the children of these parents replace the old population~\cite{la2016epsilon, cava2018probabilistic}.
To ensure a fair comparison with CMOEA, which maintains a population size that can be much larger than the number of offspring created at each generation, 
we have similarly decoupled the size of the population and the number of offspring per generation for $\epsilon$-Lexicase Selection.
In our implementation, a predetermined number of parents are selected to produce an equal number of offspring, 
and then survivors are selected from among the combined population of parents and offspring until the number of remaining individuals equals the intended population size.
We apply $\epsilon$-Lexicase Selection to both parent selection and survivor selection.

\subsection{Combined-Target Objective}

In order to perform well on a multimodal problem, in addition to preserving important stepping stones, an evolutionary algorithm will also need to preserve solutions that make progress on the multimodal problem as a whole.
In the problems presented in this paper, progress on the multimodal problem as a whole is defined as a linear combination over all objectives.
After decomposing the multimodal problem into sub-problems, due to practical limits on population size, there is no guarantee that a multiobjective evolutionary algorithm will preserve individuals that make progress on such a linear combination. However, many of them will preserve the best individual on each objective, as they are guaranteed to be on the Pareto-front and they tend to be preferred by measures intended to preserve a diverse Pareto-front (e.g. crowding score~\cite{deb2002nsga} or reference lines \cite{deb2013evolutionary}). As such, if a linear combination on all objectives would be added as an additional objective, which we will call the \emph{Combined-Target} (CT) objective, individuals that have increased performance on the multimodal problem as a whole may have a much higher chance of being preserved, which can thus increase the performance of many multiobjective evolutionary algorithms when solving multimodal problems. In this paper, we test the CT objective with each of our controls, and we show that it improves their performance in almost all cases. We did not test the CT objective with CMOEA because CMOEA already has a bin specifically dedicated towards optimizing this objective.

\section{Experiments}

\subsection{Settings and plots}

For all experiments, the number of individuals created at every generation was \numprint{1000}.
Because the population size of CMOEA can not be set directly, as it is partially determined by the number of subtasks,
it has to be tuned by setting the bin size.
We set the bin size such that each bin would be large enough to allow for some diversity within each bin while keeping the total population size computationally tractable.
For a fair comparison, the population size for all other treatments was subsequently set to be equal to the total population size maintained by CMOEA
(we tested some of the controls with a population size equal to the number of individuals created at every generation, which is a common default in EAs, but those treatments performed worse, see SI Sec.~\ref{sec:pop_size}). All experiments involved the evolution of a network with NEAT mutation operators~\cite{stanley2002evolving}, extended with deletion operators as described in previous work~\cite{huizinga2016aligning} (i.e. we implemented delete node and delete connection mutations in addition to the mutation operators already available in NEAT), and the treatments did not implement crossover. Experiment-specific settings are described in the relevant sub-sections.
 
All line plots show the median over 30 runs with different random seeds. Unless stated otherwise, shaded areas indicate the $95\%$ bootstrapped confidence interval of the median obtained by resampling \numprint{5000} times and lines are smoothed by a median filter with a window size of $21$. The performance of a run at a particular generation is defined as the highest performance among the individuals at that generation. Symbols in the bar below each plot indicate that the difference between the indicated distributions is statistically significant (according to the Mann-Whitney-U test with $\alpha=0.05$). Because these significance indicators result in a large number of statistical tests, we have applied a Bonferroni correction based on the number of treatments being compared. We did not apply a Bonferonni correction based on the actual number of tests, because we believe that the large number of tests performed between the same treatments actually helps in identifying spurious positives (see SI~\ref{sec:plot_significance_indicators} for details). Unless otherwise specified, all statistical comparisons are performed with the Mann-Whitney-U test.

\subsection{Simulated multimodal robot locomotion domain}
\label{sec:robot}

\subsubsection{Multimodal locomotion domain experimental setup}

\begin{figure}[tb]
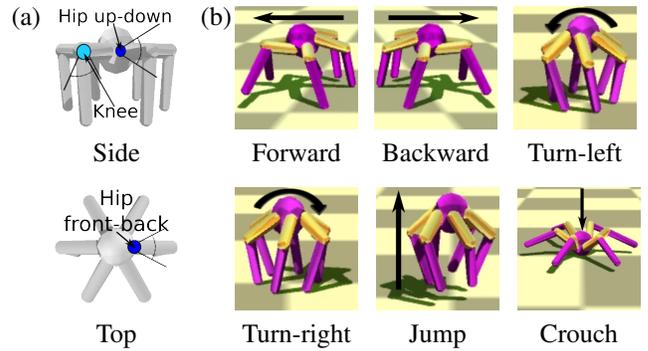

\centering
\toplabel{(a)}
\begin{minipage}[t]{0.1\textwidth}
\topfigcap{0.9}{1.0}{figures/Robot/Side-alternate3}{Side} \vspace{10pt} \\
\topfigcap{0.9}{1.0}{figures/Robot/Top-alternate2}{Top}
\end{minipage}
\toplabel{(b)}%
\begin{minipage}[t]{0.3\textwidth}
\topfigcap{0.3}{1.0}{figures/Robot/Forward-fast}{Forward}
\topfigcap{0.3}{1.0}{figures/Robot/Backward-fast}{Backward}
\topfigcap{0.3}{1.0}{figures/Robot/turnleft}{Turn-left} \vspace{10pt} \\
\topfigcap{0.3}{1.0}{figures/Robot/turnright}{Turn-right}
\topfigcap{0.3}{1.0}{figures/Robot/Jump-fast}{Jump}
\topfigcap{0.3}{1.0}{figures/Robot/Crouch-fast}{Crouch}
\end{minipage}
\caption{\textbf{Six-tasks robot and problem.} 
\textbf{(a)} The hexapod robot has 6 knee joints with one degree of freedom and 
6 hip joints with 2 degrees of freedom (up-down, front-back). 
\textbf{(b)} The six tasks that need to be learned by the robot.
}
\label{fig:robot_task}
\end{figure}

\begin{figure}[tb]
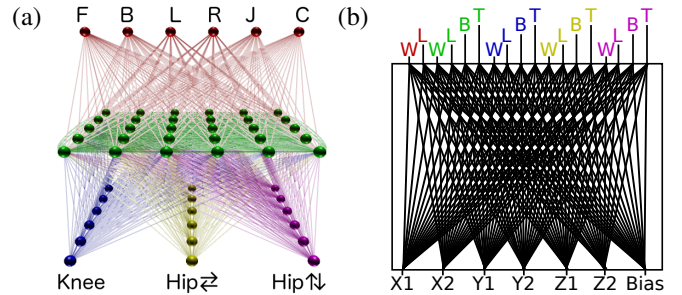

\centering
\toplabel{(a)}%
\botfig{0.20}{110pt}{1.0}{figures/substrate}
\toplabel{(b)}
\botfig{0.20}{110pt}{1.0}{figures/cppn-mss_io_tall}
\caption{\textbf{The spatial network layout, MSS planes, and associated CPPN for the multimodal locomotion task.} \textbf{(a)} Spatial layout of the network for the multimodal locomotion task. Neurons are shown in a cube that extends from -1 to 1 in all directions and neurons are placed such that the extreme neurons lie on the boundaries of this cube.
The letter above each of the six input neurons specifies with which task that neuron is associated: forward (F), backward (B), turn-left (L), turn-right (R), jump (J), and crouch (C). 
Besides these task-indicator neurons, the network has no other inputs. 
The color of every node and connection indicates to which MSS plane it belongs and it matches the color of the CPPN outputs that determine its parameters. \textbf{(b)} The CPPN for the multimodal locomotion task. Colored letters above the CPPN indicate the following outputs: weight output (W), link-expression output (L), bias output (B), and time-constant output (T). There is no bias or time-constant output in the CPPN for the red MSS plane because that plane governs the input neurons of the CTRNN, which do not have bias or time-constant parameters. Inputs to the CPPN are the three coordinates for the source (x1, y1, z1) and target (x2, y2, z2) neurons and a bias input with the constant value of 1.}
\label{fig:layout}
\end{figure}

We have tested the performance of CMOEA on two different problems.
The first is a simulated robotics problem known as the six-tasks problem~\cite{huizinga2016aligning}, 
where a simulated hexapod robot has to learn to perform six different tasks (move forward, move backward, turn left, turn right, jump, and crouch) 
depending on its inputs (Fig.~\ref{fig:robot_task}). 
Neural network controllers (Fig.~\ref{fig:layout}) are evaluated by performing a separate trial for each task, 
with the information about which task to perform being presented to the inputs.
How performance and behavioral diversity are measured on this task follow~\citet{huizinga2016aligning} and are described in SI Sec.~\ref{sec:robot_exp_details}.
The robot was simulated with the Bullet\footnote{\url{https://pybullet.org}} simulator.

Because this problem features six subtasks, CMOEA maintains $2^6-1=63$ bins (one for each combination of subtasks except the permutation with zero subtasks). For this problem, the size of each bin was set to be $100$, meaning that all controls had a population size of \numprint{6300}. 
Following previous work~\cite{huizinga2016aligning}, the controller was a Continuous-Time Recurrent Neural Network (CTRNN)~\cite{beer1992evolving} encoded by a Compositional Pattern-Producing Network (CPPN)~\cite{stanley2009hypercube}, which was extended with a Multi-Spatial Substrate (MSS)~\cite{Pugh2013} and a Link Expression Output~\cite{verbancsics2011constraining}.
For details about the aforementioned algorithms and extensions, we refer the reader to the cited papers. 
Parameters for the CTRNN and the evolutionary algorithm were the same as in~\citet{huizinga2016aligning} and are listed in the SI for convenience (SI Sec.~\ref{sec:robot_exp_details}).

\begin{figure}[tb!]
\centering
\includegraphics[width=0.48\textwidth]{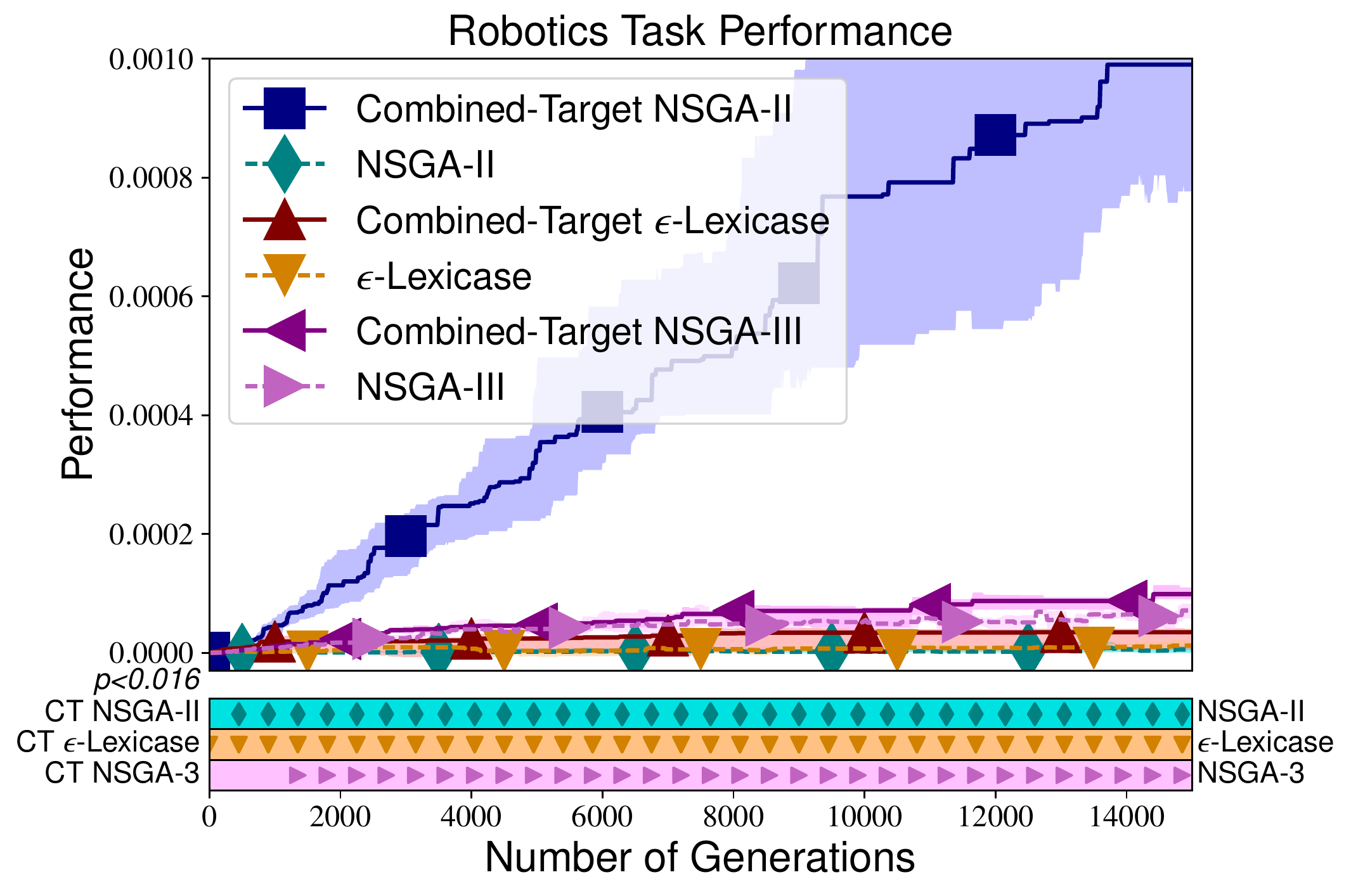}
\caption{\textbf{The CT variant of each treatment performs significantly better than that same treatment without the CT objective.} 
Note that performance values appear to be extremely low because it is the product of six numbers between 0 and 1,
but an individual with a fitness greater than 0.001 generally demonstrates some basic competency on all six tasks, 
while and individual with a fitness smaller than 0.0003 does not (videos are available at: www.evolvingai.org/cmoea).
}
\label{fig:naive}
\end{figure}

\subsubsection{Multimodal locomotion domain results} 
The first thing to note is that CT NSGA-II significantly and substantially outperformed regular NSGA-II (Fig.~\ref{fig:naive}).
Similar, though smaller, effects can be observed for NSGA-III and $\epsilon$-Lexicase Selection (Fig.~\ref{fig:naive}).
While this figure only shows results for semi-dynamic $\epsilon$-Lexicase Selection with $\epsilon=0.05$, the positive effect of the CT objective was observed in all versions that we tested (SI Fig.~\ref{fig:elexicase_combined_target}).
These results demonstrate that the CT objective is an effective method for alleviating the effects that the high-dimensional 
Pareto-front has on NSGA-II on this particular problem, and that it can aid other multiobjective evolutionary algorithms as well.
This enhancement for multiobjective evolutionary algorithms is an independent contribution of this paper.

Second, $\epsilon$-Lexicase Selection and NSGA-III, with or without the CT objective, perform far worse than CT NSGA-II.
It is unclear why these algorithms perform worse on this multimodal locomotion task. For NSGA-III it may be the case that, as it is better than NSGA-II at maintaining a well distributed Pareto front~\cite{deb2013evolutionary}, solutions that perform well on the multimodal problem as a whole are not found as quickly as with NSGA-II. This suggests that evolving multimodal behavior by splitting the desired behavior into subtasks is not a regular multiobjective problem and that a high performance on classic multiobjective benchmarks does not necessarily indicate a high performance on multimodal problems. $\epsilon$-Lexicase Selection may suffer from a similar issue and also has a second problem where the automatic calculation of $\epsilon$ does not work as well on this multimodal locomotion problem as there are many tasks on which performance is originally 0, meaning the Median Average Deviation becomes 0 as well. As a result, the version of $\epsilon$-Lexicase Selection that performed best on this domain was a semi-dynamic version with a fixed $\epsilon$ (SI~\ref{sec:lexicase_variants}), but such a fixed $\epsilon$ may not be as effective as an automatically adjusted $\epsilon$ because the optimal $\epsilon$ may change over generations. Other methods for calculating $\epsilon$ may work better, but we believe that those experiments are outside the scope of this paper.

Because the CT versions of all of NSGA-II, NSGA-III and $\epsilon$-Lexicase Selection performed better than their regular counterparts, 
we consider only the CT versions for the remainder of the multimodal locomotion task results.

\begin{figure}[tb!]
\centering
\includegraphics[width=0.48\textwidth]{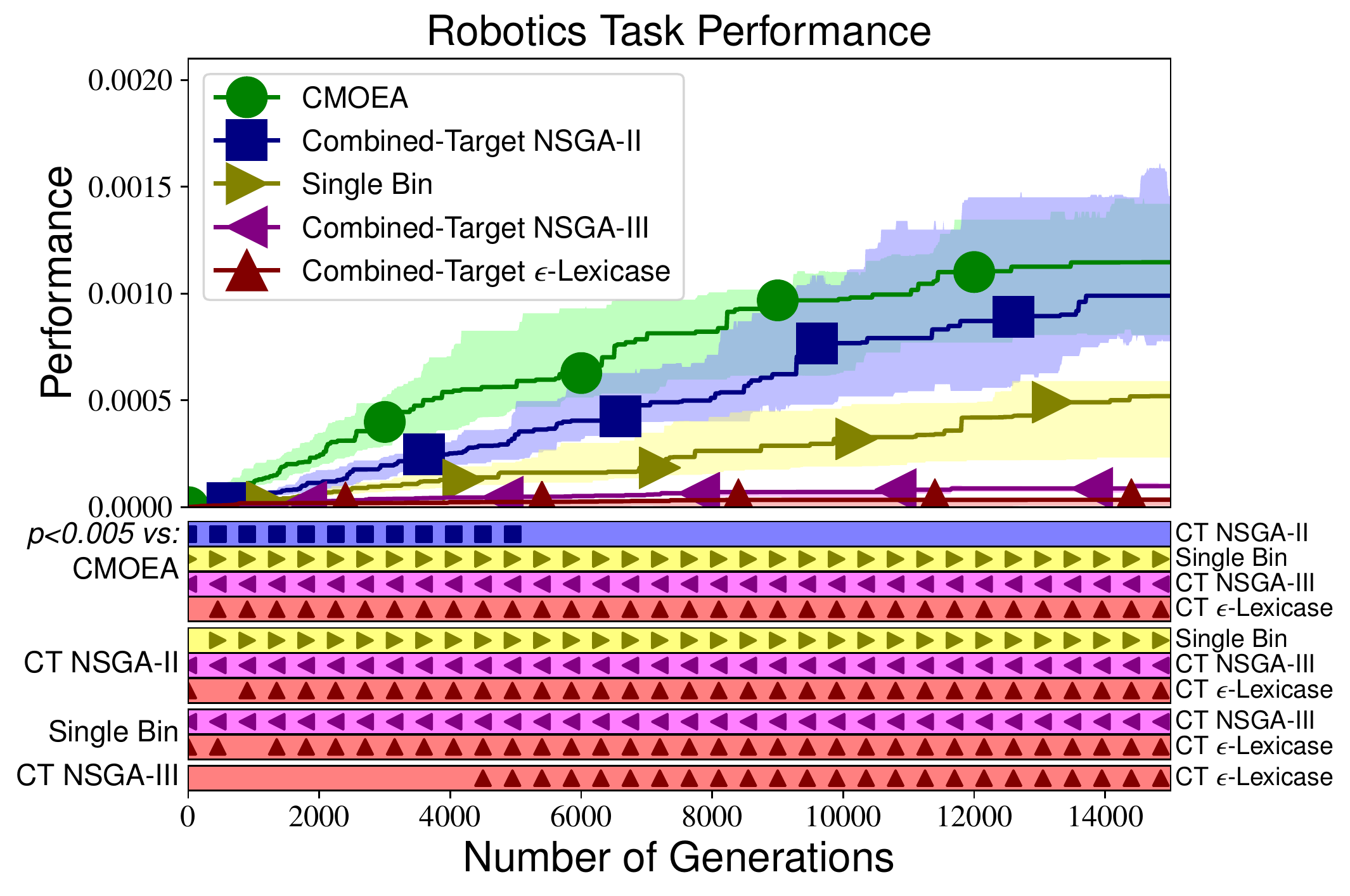}
\caption{\textbf{CMOEA performs significantly better than the control treatments in early generations.} After roughly \numprint{5000} generations, the difference between CMOEA and CT NSGA-II is no longer significant, though the difference between CMOEA and the other treatments remains significant. 
}
\label{fig:main_robots}
\end{figure}

When we compare the controls with CMOEA, we see that CMOEA performs significantly better than any of the controls for the first \numprint{5000} generations (Fig.~\ref{fig:main_robots}). After those \numprint{5000} generations, the significant difference between CMOEA and CT NSGA-II disappears, though the difference between CMOEA and the other treatments remains significant. CT NSGA-II also performs significantly better than the other treatments besides CMOEA. The fact that the Single Bin treatment performs substantially and significantly worse than both CMOEA and CT NSGA-II is indicative of the importance of having multiple different objectives. 
While the Single Bin treatment dedicates all of its resources to the combination of the six objectives, such a strategy did not lead to the best performance on this problem, presumably because it fails to find all the necessary stepping stones required to learn these behaviors.
Instead, both CMOEA and CT NSGA-II dedicate a substantial amount of resources towards optimizing subsets of these objectives 
and these subsets then form the stepping stones towards better overall performance.

\begin{figure}[tb!]
\centering
\includegraphics[width=0.48\textwidth]{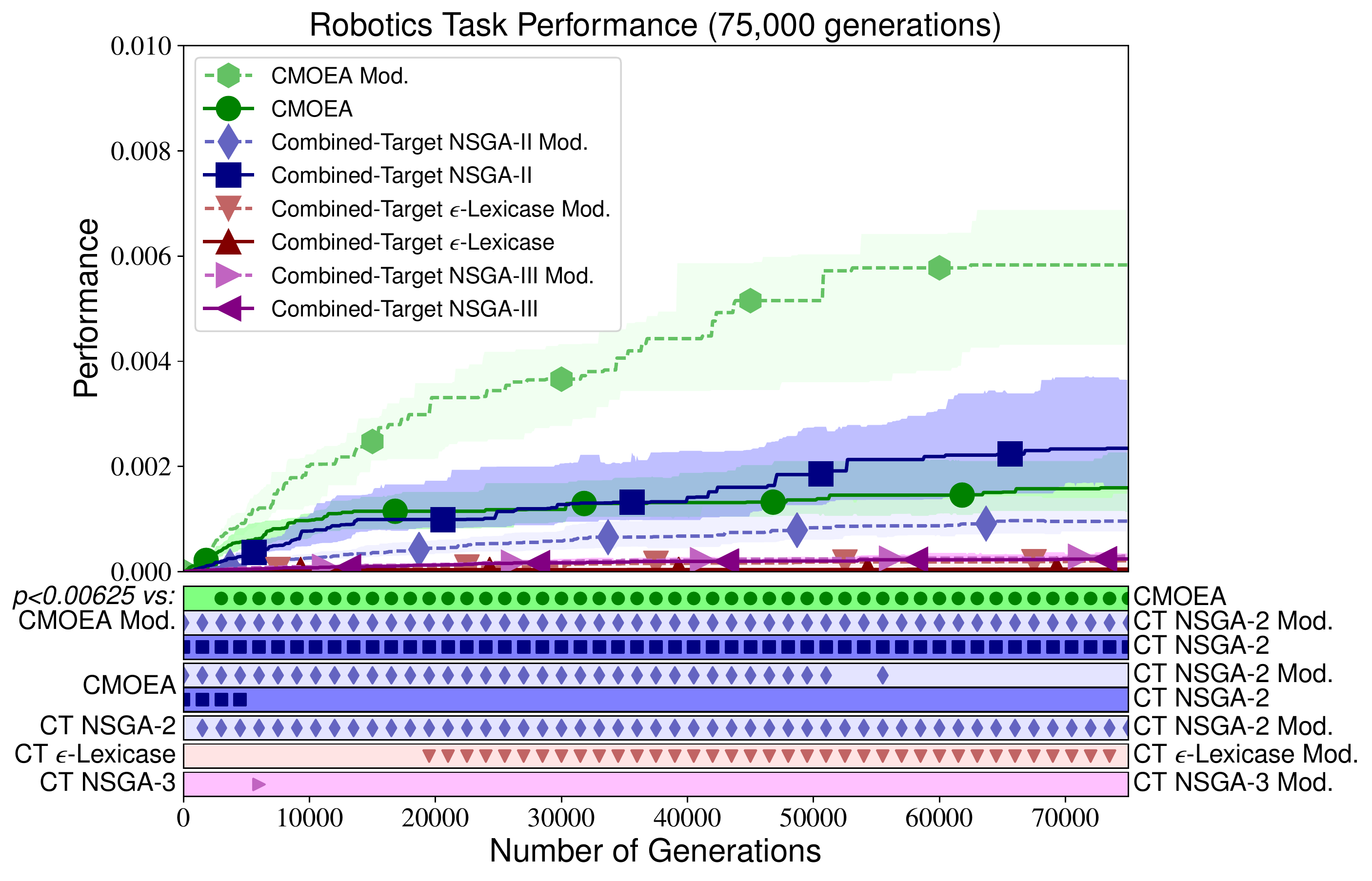}
\caption{\textbf{While CT NSGA-II outperforms CMOEA in extended runs, 
CMOEA performed significantly and substantially better than CT NSGA-II when auxiliary objectives were added.} 
The auxiliary objectives significantly improve the performance of CT $\epsilon$-Lexicase Selection, 
have no observable effect on the performance of CT NSGA-III, and significantly reduce the performance CT NSGA-II. A magnification of CT $\epsilon$-Lexicase Selection and NSGA-III with and without modularity is provided in the SI to visualize the difference (SI Fig.~\ref{fig:lexicase_blowup}).}
\label{fig:extended}
\end{figure}

Given that both CMOEA and CT NSGA-II were still gaining performance after \numprint{15000} generations and that it was unclear which algorithm would perform better in the long run, we extended the experiments with these treatments up to \numprint{75000} generations (Fig.~\ref{fig:extended}). 
While the difference is relatively small, CT NSGA-II starts outperforming CMOEA after about \numprint{40000} generations, though the difference is not significant after the (potentially overly conservative) Bonferroni correction. 
This result
suggests that, similar to CMOEA, CT NSGA-II is capable of maintaining the evolutionary stepping stones required for performing well on this task. In addition, given that CT NSGA-II maintains only seven different objectives (the six main objectives and the CT objective), it is likely that CT NSGA-II is capable of dedicating more resources to individuals that perform well on the CT objective than CMOEA, thus explaining why CT NSGA-II eventually outperforms CMOEA. If this is true, the performance of CMOEA can possibly be improved by increasing the relative population size of the bin responsible for the combination of all subtasks. That said, even without such optimization, CMOEA remains competitive with CT NSGA-II for the majority of the \numprint{75000} generations.

Previous work has shown that it may be helpful to have auxiliary objectives~\cite{jaderberg2016reinforcement, mirowski2016learning} that influence the structure of the evolved neural networks,
such as by promoting modularity or hierarchy~\cite{clune2013originModularity, mengistu2016evolutionary, ellefsen2015neural}.
In particular, the paper that briefly introduced CMOEA~\cite{huizinga2016aligning} demonstrated that selecting for genotypic and phenotypic modularity increases the performance of CMOEA on the six-tasks robot locomotion problem. 
In this paper, the modularity of the genotype (i.e. the CPPN network) and the phenotype (i.e. the neural controller itself) was measured through a computationally efficient approximation~\cite{Newman2006} of the modularity-Q score for directed networks~\cite{Leicht2008}, and these two modularity scores were subsequently added as additional objectives to be maximized within each CMOEA bin.
Modularity may be beneficial on this problem because, once the phenotypic network has developed modules, those modules can be involved in different types of behavior (e.g. there may be a separate module for moving and a separate module for turning), allowing those behaviors to be optimized separately.
However, because the network is indirectly encoded~\cite{stanley2007compositional}, a modular phenotype alone may not be sufficient to allow those modules to be separately optimized, as a local change in the genotype may cause a global change in the phenotype.
Supporting this hypothesis, previous work demonstrated that performance only increases with simultaneous selection for both genotypic and phenotypic modularity, and not with selection for either genotypic or phenotypic modularity alone~\cite{huizinga2016aligning}.

Note that these two auxiliary modularity objectives are different from the subtask objectives in that they are completely unrelated to the problem that needs to be solved,
meaning that individuals can gain performance on these objectives without making any progress towards the overall goal.
Instead, these objectives promote genotypic and phenotypic structures that increase the evolvability of individuals,
thus possibly increasing the potential of these individuals in later generations.
As was shown in~\cite{huizinga2016aligning}, being able to effectively make use of these auxiliary objectives can greatly improve the effectiveness of an algorithm.
As such, we examined whether our controls could similarly benefit from the objectives of maximizing the Q-score of the genotype network and the phenotype network, which we will refer to as our auxiliary objectives.

For CMOEA, these two auxiliary objectives were added to the NSGA-II selection procedure within every bin, thus ensuring that every individual maintained by CMOEA would be subject to selection for genotypic and phenotypic modularity~\cite{huizinga2016aligning}. 
While adding these additional objectives causes the number of objectives within each bin to increase to four, which could have been too many for the NSGA-II-based selection procedure, we hypothesize that the Pareto front within each bin does not collapse because the objectives are not fully competing (i.e. increasing the modularity of the network or its genome does not necessarily lead to a reduction in performance or diversity). We refer the reader to~\cite{huizinga2016aligning} for a more in-depth discussion of this result.

Our controls do not have a selection procedure that is equivalent to CMOEA's within-bin selection procedure.
As such, the two most straightforward ways of adding the modularity objectives to the controls is to either include the modularity metric as a weighted sum with the primary objectives or add the modularity objectives alongside the primary objectives. However, because the modularity metrics have a different scale with respect to the primary performance objectives (e.g. the modularity metrics tend to be greater than 0.2 while the combined performance objective does not become greater than 0.01), a naive weighted sum is unlikely to work without extensive tuning of the weighting. As such, we believe that adding genotypic and phenotypic modularity as separate objectives to the controls is the method that is likely to perform best, and which thus represents the most fair comparison.

CT NSGA-II performed significantly worse when the modularity objectives were added (Fig.~\ref{fig:extended}). A likely cause for this effect is that, because these auxiliary objectives are completely separate from the main objectives, the algorithm maintains individuals that are champions at having a modular genotype or a modular phenotype, but not in combination with actually performing well on any of the main objectives.
CT $\epsilon$-Lexicase Selection performs slightly (but significantly) better when modularity is added, suggesting that the algorithm is better at finding tradeoffs between modularity and performance than CT NSGA-II. The modularity objectives have no observable effect on the performance of CT NSGA-III. It is unclear why this is the case, but it is possible that, with 9 objectives, there are too few reference lines that actually describe relevant tradeoffs between modularity and performance.
CMOEA ensures a proper tradeoff between modularity and performance by having the auxiliary objectives be present in every bin, thus forcing all individuals to invest in being modular, regardless of which subtasks they solve. 
Because it may be harder to increase performance at low performance values than at higher performance values, care should be taken when comparing the relative impact of the modularity objectives. That said, the benefit of the modularity objectives on the performance of CMOEA is substantially larger than that of the controls, suggesting that CMOEA may be more effective when it comes to utilizing auxiliary objectives.

Note that, instead of adding the selection pressure for genotypic and phenotypic modularity to every CMOEA bin, 
it is possible to allocate additional bins for individuals that combine genotypic and phenotypic modularity with performance
(e.g. having one bin for jumping alone and another bin for jumping $\times$ genotypic and phenotypic modularity).
The main downside of such an approach is that it requires the tradeoff between modularity and performance to be explicitly defined. However, because such an approach would increase the number of potential stepping-stones that are preserved, it is possible that doing so would improve the performance of CMOEA even when the modularity and performance objectives are not properly balanced. 
Examining the effect of adding additional bins that select for a linear combination of modularity and performance remains a topic for future research.

\subsection{ Simulated robot maze navigation domain}
\label{sec:maze}

\subsubsection{Maze domain experimental setup}
\label{sec:maze_setup}

The second problem is a simulated robot maze navigation task, where a wheeled robot is put into a randomly generated maze and has to navigate to a goal location.
In contrast to the six-tasks problem discussed before, we do not define the different modalities of the problem explicitly. 
Instead, we have the modalities arise naturally from the problem instances.

\begin{figure}[tbp]
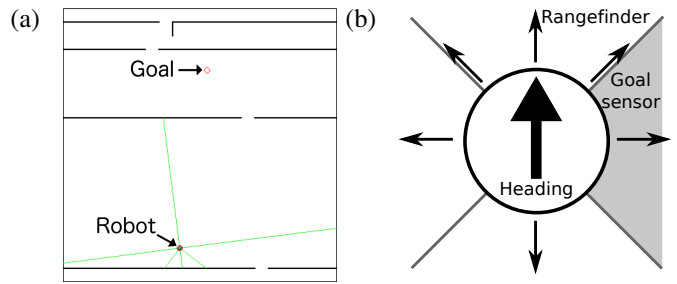

\centering
\toplabel{(a)}
\topfig{0.2}{1.0}{{"figures/maze_with_arrows"}}
\toplabel{(b)}
\topfig{0.2}{1.0}{{"figures/maze_robot_schematic"}}
\caption{\textbf{Example maze and robot schematic.} \textbf{(a)} Example maze generated by the maze generation algorithm. The green lines represent the rangefinder sensors of the robot. \textbf{(b)} The schematic of the maze exploration robot (adapted from~\cite{lehman2011abandoning}).}
\label{fig:maze_task}
\end{figure}

The mazes were generated according to the maze generation algorithm from \cite{meyerson2016learning} (originally introduced in~\cite{reynolds2010maze}), where a grid-based space is repeatedly divided by a wall with a single gap. The mazes for our experiments were generated by dividing a 20 by 20 grid 5 times with the goal placed in the center of a cell randomly selected from the grid. The grid was subsequently converted into a continuous space where each cell in the grid represented an area of 20 by 20 units. The robot had a circular collision body with a radius of 4 units, giving it plenty of space to move within a cell, and always started at the center of the maze facing north. Walls were 2 units wide and the gaps within each wall were 20 units wide. This maze-generation algorithm resulted in mazes with a house-like quality, where the space was divided into separate rooms connected by doorways and the goal positioned somewhere within one of those rooms (Fig.~\ref{fig:maze_task}a). The robot was simulated with the Fastsim simulator~\cite{Mouret2012, fastsim}.

The robot had two different types of sensors: range-finder sensors, which detect the distance to the closest wall in a certain direction, and goal sensors, which indicate whether the goal lies within a specific quadrant relative to the robot (Fig.~\ref{fig:maze_task}b). In contrast to previous work~\cite{meyerson2016learning, lehman2011abandoning}, our goal sensors did not work through walls. As a result, the problem had two different modes: in the first mode the robot has to traverse different rooms in order to find the room containing the goal, and in the second mode the robot has to move towards a goal that is located in the same room. These two behaviors are different modes because they require the robot to operate in different ways. 
The most straight-forward method for traversing all rooms is probably a wall-following strategy, 
as it implicitly implements the classic maze solving strategy of always choosing the left-most or right-most path.
However, moving towards the goal requires the robot to leave the wall and exhibit homing behavior instead, as the goal may not be located next to a wall.

In this problem, rather than selecting for each mode of behavior explicitly, we simply generate a large number of mazes that may, by chance, emphasize different modes of behavior. For example, in some random mazes, the robot will start in the same room as the goal, meaning that all it has to do is move straight towards the goal without any wall-following behavior. In other mazes, the robot may be in a different room from the goal, but the goal may be located right next to a wall. 
In those mazes, wall-following behavior alone can guide the robot to the goal, without any homing behavior being required. Lastly, some mazes will put the robot and the goal in different rooms, and put the goal somewhere in the center of a room, thus requiring both wall-following and homing behavior to be navigated successfully.

This experimental setup is especially relevant because it reflects a practical way of applying CMOEA.
While it may be hard to define in advance exactly all the different behavioral modes that are important to solve a particular problem, 
it is usually much easier to define different instances of the same problem.
As with our mazes, different instances of the same problem may emphasize different modalities and, 
as a result, these different instances may provide effective scaffolding for learning to solve the problem as a whole. 

In this experiment, the problem as a whole is not to solve a particular maze, or even any specific set of mazes, but rather to solve these house-like mazes in general. As such, any solution evolved to solve a particular set of mazes has to be tested on a set of unseen mazes to assess its generality. To do this, for every run, we generated a training set of 100 mazes to calculate the fitness of individuals during evolution and we generated a test set of \numprint{1000} mazes to assess the generality of individuals. 
The generality of solutions was evaluated every 100 generations for plotting purposes, 
but this information was not available to the algorithms.

Given that there are $100$ mazes in our training set, there are $100$ different subtasks.
As such, it is not feasible to maintain a bin for every combination of subtasks. 
Instead, we define a maximum number of bins (\numprint{1000} with bin size $10$, meaning each bin contains $10$ individuals, in this experiment) 
to which we assign different sets of subtasks.
First, we assign the combination of all subtasks to one of our bins, as this combination represents the problem we are trying to solve. 
Second, we assign every individual training task to a bin, as those provide the most obvious starting points for our algorithm. 
Lastly, the remaining bins, which we will call \emph{dynamic bins}, are assigned random combinations of subtasks. 
To create a random set of subtasks, we included each training task with a probability of $50\%$, 
meaning most bins were associated with about half of the total number of subtasks.
We choose this method for its simplicity, 
even though a different approach could have offered a smoother gradient from bins that govern only a few subtasks to bins that govern many subtasks.
Examining the effectiveness of smoothed bin selection methods is a topic for future work.

To make sure the algorithm does not get stuck because it was initialized with poor sets of subtasks, we randomly reassign the subtasks associated with one of the dynamic bins every generation. While this does mean that many of the individuals previously assigned to that bin will now be replaced (as the selection criteria may be completely different), some research has suggested that such extinction events may actually help an evolutionary process in various ways~\cite{lehman2015enhancing, krink2001self, kashtan2009extinctions}. We did not attempt to find the optimal rate at which to reassign the subtasks of the dynamic bins, but we found that the arbitrary choice of one bin per generation performed well in this particular domain.

Performance of an individual on a maze was defined as its distance to the goal divided by the maximum possible distance to the goal for that maze (i.e. the distance from the goal to the furthest corner of the maze). 
A fitness of one was awarded as soon as the body of the robot was on top of the goal, at which point the maze was considered solved.
Performance on a combination of mazes was calculated as the mean performance over those mazes.
We did not calculate the multiplicative performance on this problem because individuals did not over-specialize on easy mazes,
probably because no additional fitness could be gained after a maze was solved.
Every simulation lasted for \numprint{2500} time-steps or until the maze was solved. 
The wheels of the robot had a maximum speed of 3 units per time-step and the speed of each wheel was determined by scaling the output of the relevant output neuron to the $[-3, 3]$ range. 

The robot controllers were directly-encoded recurrent neural networks with 10 inputs (6 for the rangefinders and 4 for the goal sensors), 2 outputs (one for each wheel), and sigmoid activation functions. 
Neural network and EA settings followed~\citet{stanton2016curiosity} and are listed in SI Sec.~\ref{sec:maze_exp_details}.

\subsubsection{Maze domain results}
\label{sec:maze_results}

By the end of \numprint{1000} generations, all treatments evolved a well-known, general maze-solving solution, which is to pick any wall and follow it in one direction until the goal is reached (see video on \url{www.evolvingai.org/cmoea}). Here, the solution is also multimodal, as individuals switch from wall-following behavior to goal-homing behavior when they see the goal.

In preliminary experiments, the CT objective helped for both $\epsilon$-Lexicase Selection and NSGA-II, but slightly hurt the performance of NSGA-III in this domain (SI~\ref{sec:nsga_iii_combined_target}). As such, we include only CT $\epsilon$-Lexicase Selection, CT NSGA-II, and regular NSGA-III in the experiments presented here.

\begin{figure*}[tb!]
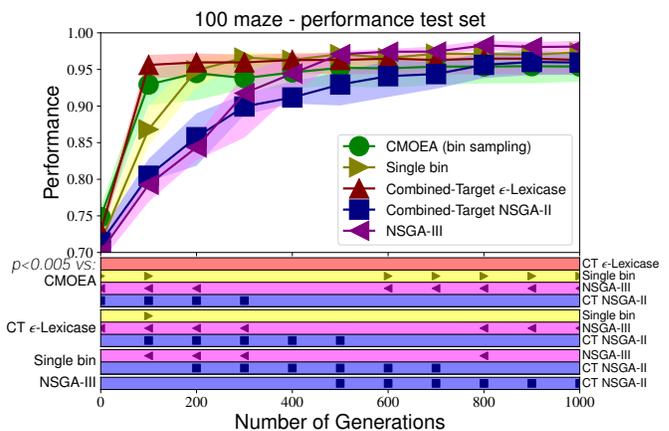

\centering
\labfigo{a}{0.49}{81}{{"figures/100_maze_training"}.pdf}
\labfigo{b}{0.49}{81}{{"figures/100_maze_test"}.pdf}
\caption{\textbf{On the maze domain, CMOEA and CT $\epsilon$-Lexicase Selection significantly outperformed the other treatments during early generations, though NSGA-III eventually has the highest performance on the test set.} \textbf{(a)} 
On the training set, the median performance of each treatment, except CT NSGA-II, converged to 1. CT $\epsilon$-Lexicase Selection and CMOEA are the first to converge, followed by the Single Bin treatment and finally NSGA-III.
The inset shows a zoom in of the indicated area,
but instead of showing a confidence interval of the median 
it shows the interquartile range, which makes it clear that the significant difference between CMOEA and the Single Bin treatment around 700 and 900 generations is because the Single Bin has more runs that converged to a performance of 1.
\textbf{(b)} Median test-set performance of the individual with the highest training-set performance from each replicate, with ties broken arbitrarily. Because performance on the test set is only evaluated every 100 generations, lines are not smoothed by a median filter. On the test set, CMOEA and CT $\epsilon$-Lexicase Selection significantly outperformed the other treatments at generation 100. NSGA-III started outperforming all other treatments from generation 800 and onward, though its difference with the single bin treatment is not significant.}
\label{fig:100_mazes}
\end{figure*}

\begin{figure*}[tb!]
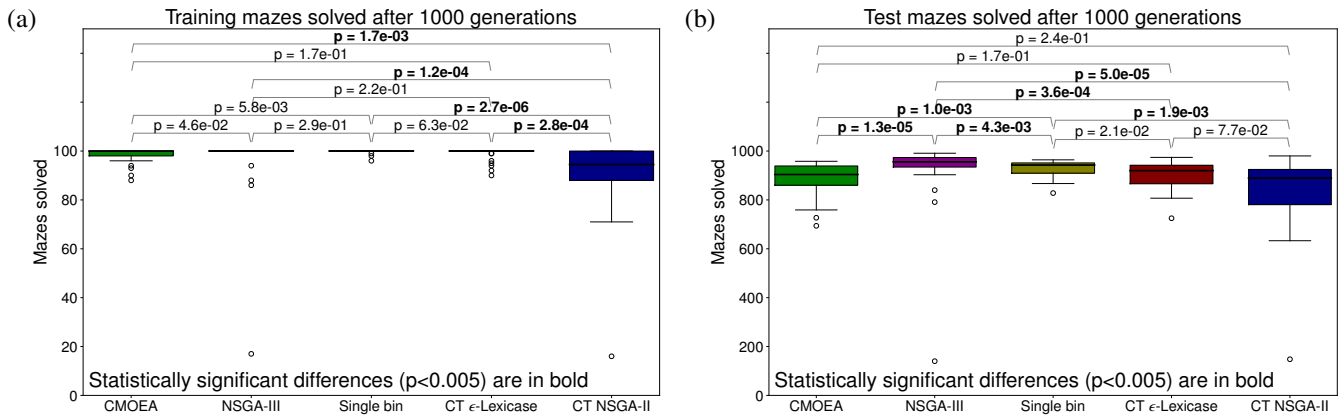

\centering
\labfigo{a}{0.49}{61}{{"figures/boxplot_training_mazes_solved"}.pdf}
\labfigo{b}{0.49}{61}{{"figures/boxplot_test_mazes_solved"}.pdf}
\caption{\textbf{All treatments except CT NSGA-II perform roughly equally in terms of number of mazes solved on the training set, but NSGA-III outperforms all other treatments on the test set.}
\textbf{(a)} The number of training mazes solved by the individual with the highest training-set performance from each replicate, with ties broken arbitrarily. 
On the training set, CT NSGA-II solved significantly fewer mazes than the other treatments. There is no significant difference between any of the other treatments.
\textbf{(b)} The number of test mazes solved by the individual with the highest training-set performance, with ties broken arbitrarily. 
On the test set, NSGA-III solved significantly more mazes than any other treatment, closely followed by the Single Bin treatment. There are no significant differences between CT $\epsilon$-Lexicase Selection, CMOEA, or CT NSGA-II.}
\label{fig:mazes_solved}
\end{figure*}

With respect to performance on the training set, CMOEA and CT $\epsilon$-Lexicase Selection performed significantly better than any of the other controls during the first 100 generations, and they both quickly converged near the optimal performance of 1, with CT $\epsilon$-Lexicase Selection converging slightly faster than CMOEA (Fig.~\ref{fig:100_mazes}a). 
The Single Bin treatment converged slightly slower than CMOEA and CT $\epsilon$-Lexicase Selection, but it reached a similar near-perfect performance after about 200 generations, indicating that the additional bins were helpful during early generations, but not required for solving this problem. After 600 generations, the Single Bin treatment actually performs significantly better than CMOEA, not in terms of its median performance, but in terms of the number of runs that obtain perfect performance (Fig.~\ref{fig:100_mazes}a inset). This difference is also apparent in terms of the number of mazes solved after \numprint{1000} generations (Fig.~\ref{fig:mazes_solved}a), as all but three Single Bin runs solved all training mazes perfectly, while the success rate of CMOEA was not as high. Both NSGA-III and Combined Target NSGA-II were slower in finding near-optimal solutions, demonstrating the debilitating effect of 100 objectives on these Pareto-dominance based methods. That said, after \numprint{1000} generations NSGA-III solves all training mazes in all but four of its runs, whereas Combined Target NSGA-II solves significantly fewer (Fig.~\ref{fig:mazes_solved}a).

For the most part, observations that held for performance on the training set also held for performance on the test set, both in terms of performance (Fig.~\ref{fig:100_mazes}b) and in terms of mazes solved (Fig.~\ref{fig:mazes_solved}b). That is, CMOEA and CT $\epsilon$-Lexicase Selection converged faster than the other treatments, but the other treatments caught up eventually (Fig.~\ref{fig:100_mazes}b). After \numprint{1000} generations, solutions found by NSGA-III have the highest performance on the test set (Fig.~\ref{fig:100_mazes}b) and solve significantly more mazes than the other treatments (Fig.~\ref{fig:mazes_solved}b).

In general, solutions found by NSGA-III generalized best relative to their performance on the training set;
NSGA-III is not significantly different from CMOEA, Single Bin, and Combined Target $\epsilon$-Lexicase Selection on the training set, but it outperforms those treatments on the test set (Figs.~\ref{fig:100_mazes}b and~\ref{fig:mazes_solved}b). It is unclear why this is the case, but it is possible that NSGA-III maintains a larger diversity of solutions than the other treatments because it attempts to maintain a well distributed 100-dimensional Pareto front. Maintaining such a larger diversity may have increased the survival of individuals that generalize well, thus resulting in slower convergence but better generalization.

Overall, CT $\epsilon$-Lexicase Selection is the treatment that converges fastest on this problem, while NSGA-III has the best test-set performance after \numprint{1000} generations. However, both CT $\epsilon$-Lexicase Selection and NSGA-III perform substantially worse than CMOEA on the multimodal locomotion problem. Conversely, while CT NSGA-II performed better than CMOEA on the multimodal locomotion problem, it does not perform as well on the maze navigation problem. As such, while CMOEA is not the best performing algorithm on either problem, it is competitive with the best algorithms on both problems, thus making it the most generally effective algorithm on these two problems.

Lastly, note that the Single Bin treatment performs well on this problem, as it converged faster than NSGA-III and generalized better than CT $\epsilon$-Lexicase Selection, suggesting that the maze navigation problem does not actually require an algorithm specialized in solving multimodal problems. That said, the maze navigation problem still highlighted different convergence and generalization properties of the tested algorithms and was useful for demonstrating how CMOEA can potentially scale to at least $100$ subtasks.

We hypothesize that CMOEA converges slower than CT $\epsilon$-Lexicase Selection because, on this maze navigation problem, it is relatively easy to find near perfect solutions on the training set. Once the population gets close to the global optimum of the search problem, CMOEA will spend a lot of computational resources in areas of the search space that are no longer relevant to the problem being solved, whereas CT $\epsilon$-Lexicase Selection maintains a pressure on the entire population to solve the remaining mazes. However, we argue that this is only a minor disadvantage for most practical problems, as it is unlikely that an evolutionary algorithm will actually get near the true global optimum for a real-world problem. 
In those problems, diversity and different stepping stones are likely to remain relevant for the entirety of an evolutionary run. 
However, even if this is not the case, one can switch from CMOEA with many bins to CMOEA with a single bin when there is a belief that additional bins are no longer beneficial.
For example, one could switch after a predetermined number of generations or when performance gains slow down.
Alternatively, one could estimate the contribution of each bin separately by measuring the number of generations since a child from this bin managed to survive in a different bin, and slowly remove the number of bins over time.
Either way, these strategies would allow CMOEA to maintain selection pressure, even when close to the global optimum.
Analyzing the effectiveness of such a version of CMOEA is a fruitful topic for future research.

\section{Conclusion}

Many real-world problems require multimodal behavior, from self-driving cars, which need to act differently depending on where they are, to medical robots, which require a wide range of different behaviors to perform different operations. Unfortunately, complex multimodal behavior may be difficult to learn directly and classic evolutionary optimization algorithms tend to rely on manual staging or shaping in order to learn such tasks. 
Such manual staging or shaping of a task requires extensive domain knowledge because finding the correct stepping stones and the order in which they should be traversed is a difficult problem, making it hard to estimate whether any particular staging or shaping strategy is truly optimal for the problem at hand.
In this paper, we have introduced the Combinatorial Multi-Objective Evolutionary Algorithm (CMOEA), an algorithm specifically designed to solve complex multimodal problems automatically, without having to explicitly define the order in which the problem should be learned.

We have shown that CMOEA is effective at solving two different tasks: (1) a simulated multimodal robot locomotion task and (2) a simulated robot maze navigation task. 
We have also introduced the Combined-Target (CT) objective, which improves the performance of NSGA-II, NSGA-III and $\epsilon$-Lexicase Selection when evolving multimodal behavior.
On the multimodal locomotion task, CMOEA outperforms CT NSGA-III, CT $\epsilon$-Lexicase Selection, and a variant of CMOEA with only a single bin, and it is competitive with CT NSGA-II.
On the maze domain, CMOEA converges faster than NSGA-III, CT NSGA-II, and Single Bin CMOEA, though it does not generalize as well, and it is competitive with CT $\epsilon$-Lexicase Selection. Thus, at least on these two problems, it is more generally effective than any other single algorithm.
Lastly, we have shown that CMOEA is more effective at incorporating auxiliary objectives that increase the evolvability of individuals, and these auxiliary objectives enable CMOEA to substantially outperform all controls on the multimodal locomotion task.

\section{Acknowledgments}
We thank Christopher Stanton, Roby Velez, Nick Cheney, and Arash Norouzzadeh for their comments and suggestions. This work was funded by NSF CAREER award 1453549.

\bibliographystyle{unsrtnat}

\bibliography{references_abbreviations_short,references}{}

\clearpage

\renewcommand{\thesection}{S\arabic{section}}
\renewcommand{\thesubsection}{\thesection.\arabic{subsection}}

\newcommand{\beginsupplementary}{%
		\setcounter{table}{0}
	\renewcommand{\thetable}{S\arabic{table}}%
		\setcounter{figure}{0}
	\renewcommand{\thefigure}{S\arabic{figure}}%
		\setcounter{section}{0}
}
\newcommand{\suptitle}{Supplementary materials for:\\\papertitle}

\newcommand{\toptitlebar}{
	\hrule height 4pt
	\vskip 0.25in
	\vskip -\parskip%
}
\newcommand{\bottomtitlebar}{
	\vskip 0.29in
	\vskip -\parskip
	\hrule height 1pt
	\vskip 0.09in%
}

\beginsupplementary

\newcommand{\maketitlesupp}{
\newpage
\twocolumn[
	\begin{@twocolumnfalse}
	\null
	\vskip .375in
	\begin{center}
		{\Large \bf \suptitle \par}
		\vspace*{24pt}
		{
			\large
			\lineskip .5em
			\par
		}
		\vskip .5em
		\vspace*{12pt}
	\end{center}
\end{@twocolumnfalse}
]
}

\maketitlesupp

\section{Experimental details}

\subsection{Plot significance indicators}
\label{sec:plot_significance_indicators}

Most line plots presented in this paper and its SI show ``significance indicators'' below each plot, which indicate whether there exists a statistically significant difference (Mann-Whitney-U test with $\alpha=0.05$) between two treatments. Together with the $95\%$ bootstrapped confidence intervals, they are intended to provide the reader with an indication of where apparent differences between two treatments may  represent an actual difference in the underlying distributions. Depending on the number of generations, about two dozen tests are performed for each comparison between two treatments.

Under the assumption that these tests are independent, performing this many statistical tests greatly increases the probability of observing false positives. 
However, such false positives can be easily detected, because they will generally be isolated points without neighbors. 
We do not hide such false positives from the reader, but we only consider a difference to be statistically significant if, under the assumption that each test is independent, there exists a consecutive series of significance indicators such that the probability that all of them are false positives is smaller than our chosen $\alpha$ of $0.05$ (which is two consecutive positives for up to 20 tests and three consecutive positives for up to 410 tests).
However, it is important to note that the statistical tests are not independent; if there exists a significant difference between two treatments at time $t$, it is likely there will be a significant difference at time $t+1$, and if there was no significant difference between two treatments at time $t$ then it is likely that there will not be a significant difference at time $t+1$.
In the strongest case, where two consecutive tests will always give the same result, multiple consecutive positives do not indicate a reduced probability of there being a false positive, but the additional tests also do not increase the probability of a false positive (e.g. either non of the tests are false positives, or all of them are).
For our experiments the behavior will be somewhere in the middle, but in either case, given the precaution of looking for a sufficient number of consecutive positives, these additional tests should not increase the probability of false positives to be higher than the chosen $\alpha$. 

In addition to performing many consecutive tests, we also compare many different treatments. In the main paper, to reduce the probability of false positives as a result of comparing many different treatments, we apply a per-figure Bonferroni correction based on the number of treatments being compared. We did not apply a similar correction to the SI, as these graphs are intended to explain our decision making progress, rather than be results on their own. As such, in the SI, the significance indicators should only be considered as a visual aid, rather than as a proper indicator of significance.

\subsection{Simulated multimodal robot locomotion experiment}
\label{sec:robot_exp_details}

Below is a description of the settings for the multimodal locomotion experiment. All settings are from~\cite{huizinga2016aligning}. 

\subsubsection{Performance evaluation}

Performance for the different robotics tasks is calculated in six separate trials, one for each task. 
During each trial, the neural network input associated with the task being evaluated is set to 1, and the other inputs are set to 0. 
At the start of each trial, the robot is moved to its starting position of $[0, 1, 0]$.
Performance values on the forward ($p_{f}$), backward ($p_{b}$), and crouch ($p_{c}$) tasks are calculated as:

\begin{equation*}
p_{f} = \frac{x_T}{12.5}
\end{equation*}

\begin{equation*}
p_{b} = \frac{-x_T}{12.5}
\end{equation*}

\begin{equation*}
p_{c} = \frac{1}{T}\sum_{t=1}^T(1 - ||\vec{c}_t - [0, 0, 0]||)
\end{equation*}

Where $\vec{c}_t = [x_t, y_t, z_t]$ is the center of mass of the robot at time-step $t$, and $T$ is the total number of time-steps in an evaluation. $12.5$ is a normalizing constant that was estimated based on the maximum performance reached on this objective in preliminary, single-objective experiments. Performance values on the turning tasks, turn-left ($p_{l}$) and turn-right ($p_{r}$), are calculated as:

\begin{equation*}
p_{l} = \frac{25}{T-1}\sum_{t=2}^T(\angle_l(\vec{x}_t, \vec{x}_{t-1}) u_t) + min(1 - \frac{1}{T} \sum_{t=1}^T||\vec{c}_t - \vec{c}_0||, 0)
\end{equation*}

\begin{small}
\begin{equation*}
p_{r} = \frac{25}{T-1}\sum_{t=2}^T(-\angle_l(\vec{x}_t, \vec{x}_{t-1}) u_t) + min(1 - \frac{1}{T} \sum_{t=1}^T||\vec{c}_t - \vec{c}_0||, 0)
\end{equation*}
\end{small}

Where $\vec{x}_{t}$ is a vector pointing in the forward direction of the robot at time $t$, $\angle_l(\vec{x_1}, \vec{x_2})$ is the left angle between $\vec{x_1}$ and $\vec{x_2}$, and $u_t$ is $1$ when the robot is upright (the angle between the robot's up vector and the $y$ axis is less than $\frac{\pi}{3}$) and $0$ otherwise. In short, turning fitness is defined as the degrees turned while being upright, with a penalty for moving more than one unit away from the start. $25$ is a normalizing constant that was estimated based on the maximum performance reached on this objective in preliminary, single-objective experiments. Lastly, jump performance ($p_{j}$) is defined as:

\begin{equation*}
p_{j} =  \left\{
  \begin{array}{lr}
  y_{max} & : y_{max} \cdot u_{T} \leq 0.5\\
  y_{max} + 1 - ||\vec{c}_T - \vec{c}_0||& : y_{max} \cdot u_{T} > 0.5
  \end{array}
\right.
\end{equation*}

Where $y_{max}$ is defined as $max_{t=1}^{T/2}(1 - ||\vec{c}_t - [0, 2, 0]||)$. During the first half of the evaluation, this equation rewards the robot for jumping towards a $[0, 2, 0]$ target coordinate. During the second half of the evaluation, provided that the robot was able to jump at least half-way towards the target coordinate and that it is upright at the end of the trial, it can obtain additional fitness by returning to the starting position. This second half was added to encourage a proper landing. For bins with multiple subtasks, performance values are multiplied to obtain the fitness of individuals. The number of time steps was $400$ for the forward and backward tasks and $200$ for the other subtasks.

\subsubsection{Behavioral diversity}

To calculate the behavior descriptor for each individual, we first recorded 6 training-task vectors by setting the input for one of the subtasks to 1, and then binarizing the values of the 18 actuators over 5 time-steps by setting all values $>0$ to 1 and other values to $0$, which resulted in 6 binary vectors of 90 elements each. 
We then created a seventh \emph{majority vector} by taking the element-wise sum of the 6 training-task vectors,
and binarizing the result such that values $>3$ were set to 1 and others were set to $0$.
Lastly, we XORed the majority vector with every training-task vector and concatenated the 6 resulting vectors to create the behavior descriptor.
Distances between behavior descriptors were calculated with the hamming distance.

\subsubsection{Parameters}

\begin{table}
\centering
\begin{tabular}{|l|l|}
\hline Parameter                              & Value  \\
\hline Population size                        & \numprint{6300} \\
\hline CMOEA bin size                         & $100$ \\
\hline CMOEA number of bins                   & $63$ \\
\hline Add connection prob.                   & $9\%$  \\
\hline Delete connection prob.                & $8\%$  \\
\hline Add node prob.                         & $5\%$  \\
\hline Delete node prob.                      & $4\%$  \\
\hline Change activation function prob.       & $10\%$ \\
\hline Change weight prob.                    & $10\%$ \\
\hline Polynomial mutation $\eta$             & $10$ \\
\hline Minimum weight CPPN                    & $-3$   \\
\hline Maximum weight CPPN                    & $3$    \\
\hline Minimum weight and bias CTRNN          & $-2$   \\
\hline Maximum weight and bias CTRNN          & $2$    \\
\hline Minimum time-constant CTRNN            & $1$   \\
\hline Maximum time-constant CTRNN            & $6$    \\
\hline Activation function CTRNN              & $\sigma(x)=tanh(5x)$ \\
\hline
\end{tabular}
\vspace{3pt}
\caption{Parameters of the multimodal robotics task.}
\label{tab:robotics_parameters}
\end{table}

\begin{table}
\centering
{
\renewcommand{\arraystretch}{1.5}
\begin{tabular}{|l|l|}
\hline Function                               & Definition  \\
\hline Sine             					     & $\sigma(x)=sin(x)$  \\
\hline Sigmoid                                & $\sigma(x)=\frac{2}{1+e^{-x}} - 1$  \\
\hline Gaussian                               & $\sigma(x)=e^{-x^2}$  \\
\hline Linear (clipped)                       & $\sigma(x)=\frac{clip(x, -3, 3)}{3}$  \\
\hline
\end{tabular}
}
\vspace{3pt}
\caption{CPPN activation functions for the multimodal robotics task.}
\label{tab:activation_functions}
\end{table}

The network for the robotics task was represented by the HyperNEAT encoding, 
meaning that a CPPN genotype detemined the weights of the neural network controller~\cite{stanley2009hypercube}.
The CPPN was evolved with the following NEAT mutation operators: add connection, delete connection, add node, delete node, change weight, and change activation function (probabilities are listed in table~\ref{tab:robotics_parameters}). 
The change weight and change activation function mutations were per connection and per node, respectively. 
Weights were mutated with the polynomial mutation operator~\cite{deb2001page124multi}. 
The possible activation functions for the CPPN were: sine, sigmoid, Gaussian and linear, where the linear function was scaled and clipped. 
See table~\ref{tab:activation_functions} for the definitions of each activation function.
Nodes did not have an explicit bias, but a bias input was provided to the CPPN.
After mutation, all weights were clipped so they would not fall outside the minimum and maximum values (see table~\ref{tab:robotics_parameters}). 
Initial CPPNs were fully connected without hidden neurons and with their weights and activation functions uniformly drawn from their allowable range.
The CPPN had separate outputs for the weights, the biases, and the time-constants of the CTRNN, 
and those outputs were scaled to fit the minimum and maximum values of the respective CTRNN parameter (see table~\ref{tab:robotics_parameters}).
For the CTRNN, the activation function of the hidden neurons was scaled to $[0, 1]$ to ensure inhibited neurons would not propagate signals.

\subsection{Simulated robot maze navigation experiment}
\label{sec:maze_exp_details}

Below are the settings for the maze navigation experiment. Evolutionary algorithm and neural network settings are from~\cite{stanton2016curiosity}.

\subsubsection{Performance evaluation}

As mentioned in the main paper (Sec.~\ref{sec:maze_setup}), performance of an individual on a maze was defined as its distance to the goal divided by the maximum possible distance to the goal for that maze, with a performance of 1 awarded if the robot would hit the goal itself.
The equation is:

\begin{equation*}
p =  \left\{
  \begin{array}{ll}
  1 - (dist / maxDist) & : dist \geq radius\\
  1 & : otherwise
  \end{array}
\right.
\end{equation*}

Here, $dist$ is the distance between the robot and the goal at the end of the simulation, $maxDist$ is the distance between the goal and the furthest corner, and $radius$ is the radius of the circular robot. 
A maze would be considered solved as soon as $dist < radius$, and the simulation would end immediately when this condition was met. 

\subsubsection{Behavioral diversity}

The behavior descriptor for a single maze was defined as the $(x, y)$ coordinate of the individual at the end of the simulation. 
The behavior descriptor of an individual over all mazes was a one dimensional vector composed of the final $(x, y)$ coordinates over all mazes. 
Distance between behavioral descriptors was defined as the Manhattan distance between those vectors.

\subsubsection{Parameters}

\begin{table}
\centering
\begin{tabular}{|l|l|}
\hline Parameter                              & Value  \\
\hline Population size                        & \numprint{10000} \\
\hline CMOEA bin size                         & $10$ \\
\hline CMOEA number of bins                   & \numprint{1000} \\
\hline Add connection probability             & $15\%$  \\
\hline Delete connection probability          & $5\%$  \\
\hline Rewire connection probability          & $15\%$  \\
\hline Add node probability                   & $5\%$  \\
\hline Delete node probability                & $5\%$  \\
\hline Change bias probability                & $10\%$ \\
\hline Change weight probability              & $10\%$ \\
\hline Polynomial mutation $\eta$             & $15$ \\
\hline Minimum weight                         & $-1$   \\
\hline Maximum weight                         & $1$    \\
\hline Activation function                    & $\sigma(x) = \frac{1}{e^{-5x} + 1}$ \\
\hline
\end{tabular}
\vspace{3pt}
\caption{Parameters of the maze navigation task.}
\label{tab:maze_parameters}
\end{table}

In the maze experiment, the controller was a directly encoded recurrent neural network. 
The controller was evolved with to the following NEAT mutation operators: add connection, remove connection, rewire connection, add node, remove node, change weight, and change bias (probabilities are listed in table~\ref{tab:maze_parameters}).
The change weight and change bias mutations were per connection and per node, respectively. 
Weights and biases were mutated with the polynomial mutation operator~\cite{deb2001page124multi}.
After mutation, weights and biases were clipped to lie within their allowable range.
To determine whether a rewire connection mutation would be applied, the operator would iterate over all connections, 
apply the rewire mutation with the indicated probability (Tab.~\ref{tab:maze_parameters}), 
and stop iterating as soon as the mutation was applied once.
The ordering of the connections in this process was arbitrary.
When applied, it would change either the source ($50\%$) or the target (the other $50\%$) of the connection, and randomly draw a new source or target from the available candidates.
Multiple connections with the same source and target would not be allowed.
Initial networks were created with between 10 and 30 hidden neurons and between 50 and 250 connections and their weights and biases were uniformly drawn from the allowable range.

\section{Additional analysis}

\subsection{Main paper figure 7 magnification}
\label{sec:lexicase_blowup}

\begin{figure}[tb!]
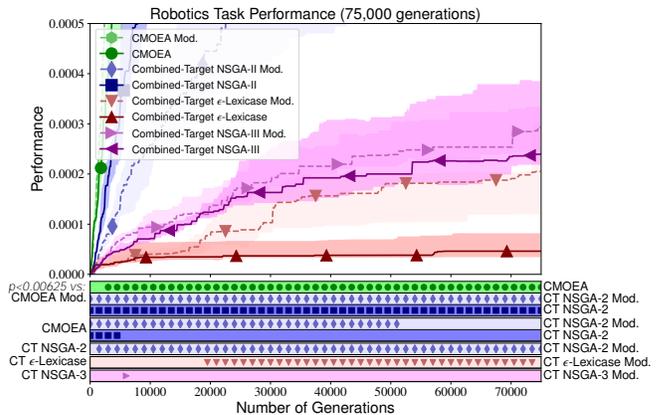

\centering
\myfig{0.48}{"figures/cmoea_guided_nsga_with_without_mod_blowup"}
\caption{\textbf{Combined-Target $\epsilon$-Lexicase Selection performs significantly better with selection for genotypic and phenotypic modularity, while modularity does not seem to have an observable effect on the performance of NSGA-III.} This figure is a magnification of figure~\ref{fig:extended} from the main paper.}
\label{fig:lexicase_blowup}
\end{figure}

Because the performance of Combined-Target $\epsilon$-Lexicase Selection and NSGA-III is low relative to the other treatments, the effect that selection for modularity has on these treatments is difficult to see in the original figure (main paper Fig.~\ref{fig:extended}). Here we provide a magnification of that figure (Fig.~\ref{fig:lexicase_blowup}).
The figure shows that, while the extra objectives of maximizing genotypic and phenotypic modularity have little effect on NSGA-III, Combined-Target $\epsilon$-Lexicase Selection performs significantly better after roughly \numprint{20000} generations.

\section{Preliminary experiments}

\subsection{Number of training mazes}
\label{sec:number_training_mazes}

\begin{figure*}[tb!]
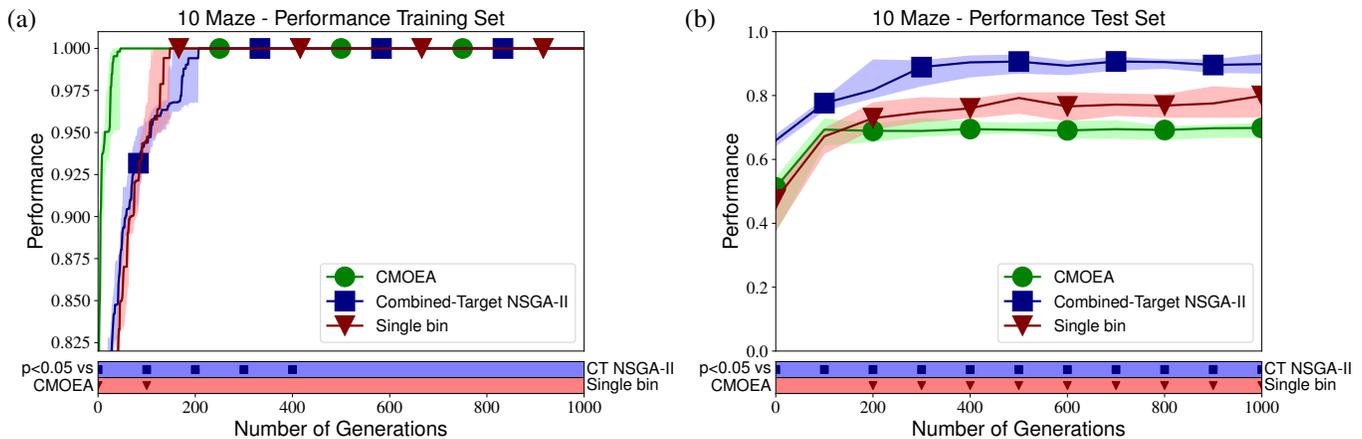

\centering
\labfigo{a}{0.49}{63}{"figures/10_maze_training"}
\labfigo{b}{0.49}{63}{"figures/10_maze_test"}
\caption{\textbf{A training set of 10 mazes does not lead to general maze-solving behavior on the test set.} \textbf{(a)} On the training set of 10 mazes, all treatments quickly converge to the optimal value of 1, suggesting that all treatments can solve all training mazes. \textbf{(b)} On the test set of \numprint{1000} mazes, none of the treatments are able to reach a performance of 1, indicating that they are unable to solve all \numprint{1000} test mazes and suggesting that the 10 training mazes were insufficient to evolve general maze solving behavior. Data for each treatment is from 30 independent runs.}
\label{fig:10_mazes}
\end{figure*}

In order to evolve general maze solving behavior, it is necessary to have a sufficiently large training set that allows individuals to learn the general behaviors necessary to solve mazes. In initial experiments, we tested CMOEA (without bin-sampling and with a bin size of 10), the Single Bin control, and Combined-Target NSGA-II with a training set of only 10 mazes (Fig.~\ref{fig:10_mazes}).
All treatments reach perfect performance on the 10 training mazes before 250 generations, but the highest test-set performance is around 0.9, 
demonstrating that the treatments do not perfectly generalize to other mazes. 
Interestingly, CMOEA is the first treatment to solve all 10 training mazes, yet it is has the lowest performance on the test set, while Combined-Target NSGA-II is the last treatment to solve all 10 training mazes, but it obtains the highest performance on the test set.
This result suggests that, while attempting to maintain a Pareto-front over all objectives slows down progress on the combination of all objectives,
such an approaches also preserves more general strategies than the bin-wise approach implemented in CMOEA.
We argue that this issue can be resolved by providing CMOEA with a larger number of training mazes,
as is presented in the main paper (Sec.~\ref{sec:maze_results}), as this makes it harder to overfit to the training set.

\begin{figure}[tb!]
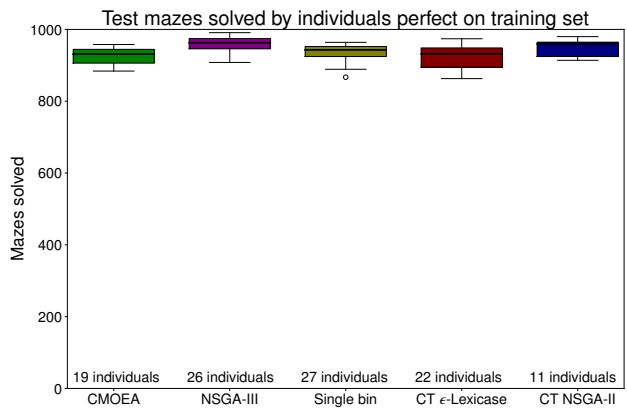

\centering
\myfig{0.48}{"figures/boxplot_train_perfect_test_mazes_solved"}
\caption{\textbf{Even with a training set of 100 mazes, individuals do not perfectly generalize to \numprint{1000} mazes.} Plot shows the number of test mazes solved by the individuals from each treatment that were capable of solving all mazes. Below each box is the number of individuals from the relevant treatment that were able to perfectly solve all 100 training mazes.}
\label{fig:perfect_train_on_test}
\end{figure}

Because there exist simple strategies that should generalize well to all mazes (e.g. general wall following behavior combined with homing behavior),
we would expect that an evolutionary algorithm should be able to find such a strategy given a sufficiently large number of training mazes.
That said, even when we increased the number of training mazes to 100, individuals that perfectly solved all 100 mazes still did not generalize to all \numprint{1000} unseen mazes from the test set, regardless of which algorithm produced those individuals (Fig.~\ref{fig:perfect_train_on_test}).
Visualizing these individuals on the test mazes that they were unable to solve revealed that most failures happened because of rare sensor values, such as being in the corner of an unusually large room or seeing the goal through the doorway of an adjacent room.
These rare sensor values caused inefficient behavior that resulted in the robot not being able to reach the goal in time. 
A video that includes some of these failed test cases is available at: \url{www.evolvingai.org/cmoea}. 
It is likely that an even larger number of training mazes could help these individuals learn how to deal with these corner cases, but doing so is a topic for future work.

\subsection{CMOEA bin selection}
\label{sec:bin_selection}

In early experiments, we examined two survivor selection methods for within CMOEA bins. 
To ensure that CMOEA bins would not be populated by near-identical copies of the same individual,
both selection methods included mechanisms that would be able to preserve within-bin diversity by maintaining individuals that would solve the same combination of tasks in different ways. The selection methods were: (1) NSGA-II's non-dominated sorting with behavioral diversity as a secondary objective~\cite{deb2002nsga}, explained in detail in the main paper, and (2) a selection method inspired by Novelty Search with Local Competition~\cite{lehman2011evolving}. In this second variant, whenever an individual had to be removed in order to reduce the number of individuals in a bin back to the predefined bin size, the algorithm would find the two individuals in the bin that were closest to each other in terms of their behavior (calculated with the same distance metric used when calculating behavioral diversity), and out of those two it would remove the individual with the lowest fitness. As such, this method would promote a diverse set of individuals with fitness values that were high with respect to their behavioral neighborhood.

\begin{figure*}[tbp!]
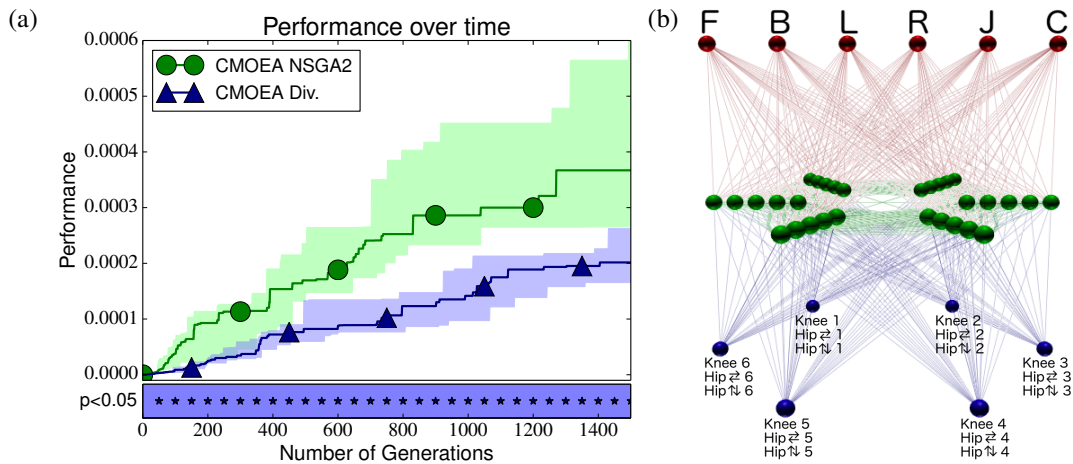

\centering
\labfig{a}{0.43}{1.0}{"figures/nsga_vs_div"}
\labfig{b}{0.28}{1.0}{"figures/circleLayoutLabeled"}
\caption{\textbf{(a) CMOEA combined with NSGA-II performed significantly better than CMOEA combined with a Novelty-Search-with-Local-Competition based method in preliminary experiments.} Data for each treatment is from 30 independent runs. (b) The neuron layout for this preliminary experiment was different from the neuron layout of the main experiment. The neurons are depicted in a cube extending from -1 to 1 in all directions. Inputs are positioned as described in the main paper, but the hidden layer consists of 6 rows of 5 neurons, where the rows form the radially distributed spokes of a circle perpendicular to the y-axis with a radius of 1. The first neuron in each row is positioned 0.5 units from the center, and the last neuron is positioned at 1.0 units from the center. Output neurons are positioned similarly, except that all output neurons are positioned at 1 unit from the center. Note that the output neurons for the knee and hip joints have overlapping positions; their positions are differentiated through the Multi-Spatial Substrate technique, which means that they have separate CPPN outputs~\cite{Pugh2013}.}
\label{fig:nsga_vs_local_comp}
\end{figure*}

In these experiments, NSGA-II's non-dominated sorting algorithm performed significantly better than the selection method based on Novelty Search with Local Competition (Fig.~\ref{fig:nsga_vs_local_comp}a). One possible reason for this result is that the Novelty-Search-with-Local-Competition based method would lead to a higher diversity at the cost of a lower average performance inside each bin. Given that CMOEA already has its bins as a method of maintaining diversity, within bin performance may be more important than within bin diversity for the purpose of solving multimodal problems. 

It is important to note that, for these experiments, the network layout was different from the layout used in the main paper (Fig.~\ref{fig:nsga_vs_local_comp}b). Specifically, input and output neurons were positioned in a radially symmetric pattern corresponding to the physical location of the sensors and actuators of the robot.
Other preliminary experiments suggested that the grid based layout presented in the main paper performed better in general, so we did not perform any further experiments with the radial layout.
However, as we have no reason to suspect that network layout would interact with the within-bin survivor-selection method, we did not repeat the bin-selection experiments with the grid-based layout.

\begin{figure*}[tbp!]
\centering
\includegraphics[width=0.48\textwidth]{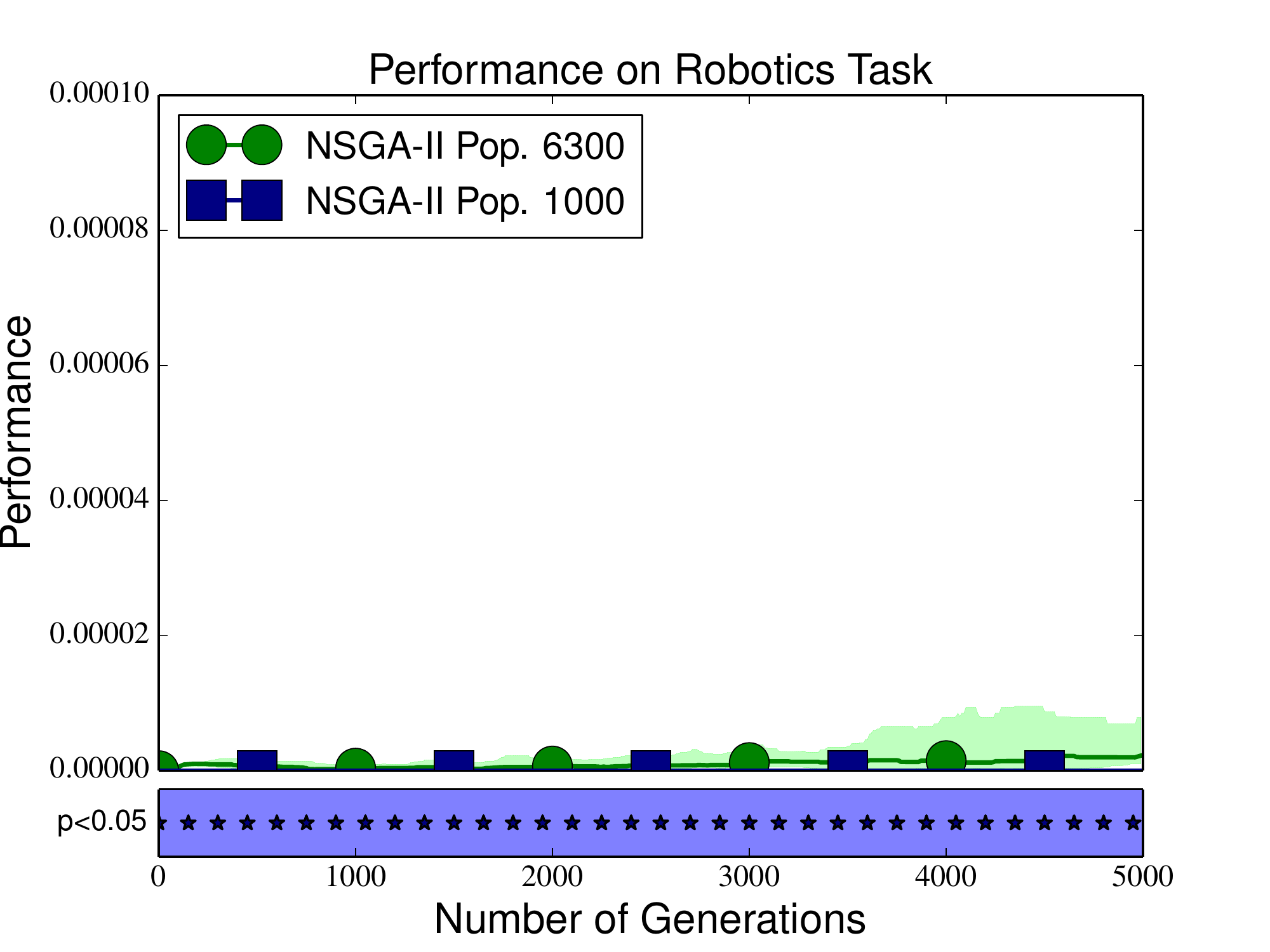}
\includegraphics[width=0.48\textwidth]{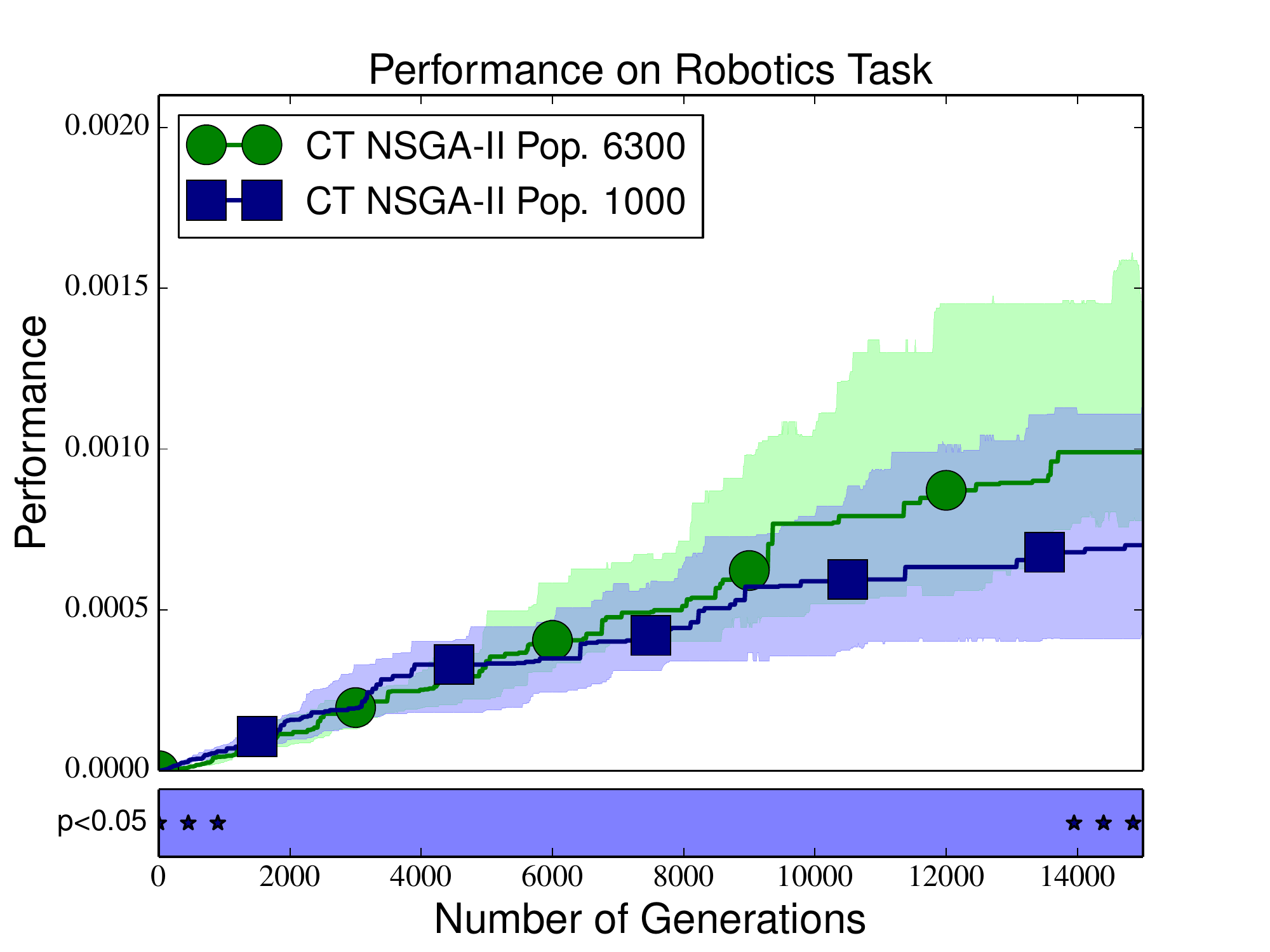}
\includegraphics[width=0.48\textwidth]{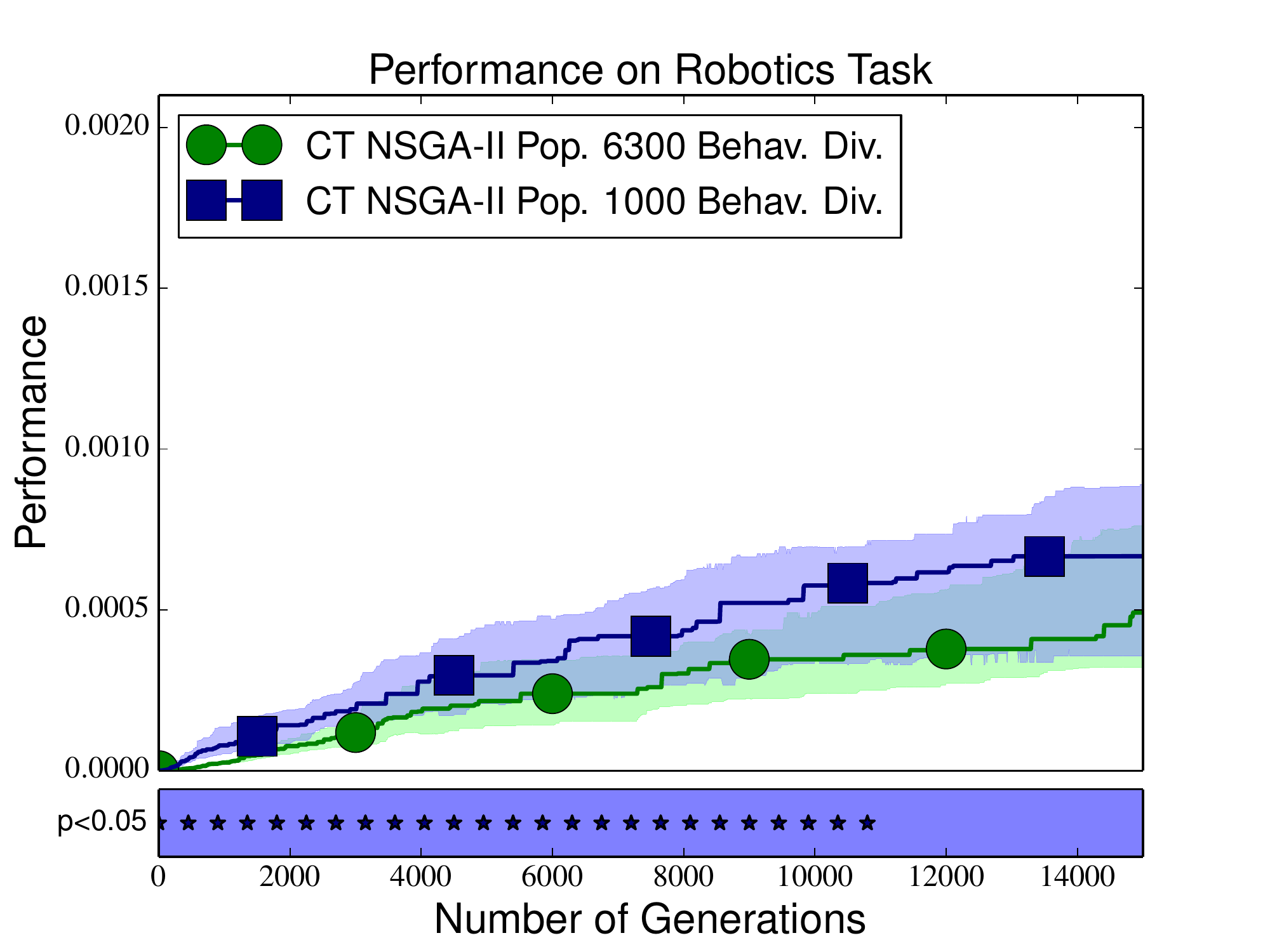}
\includegraphics[width=0.48\textwidth]{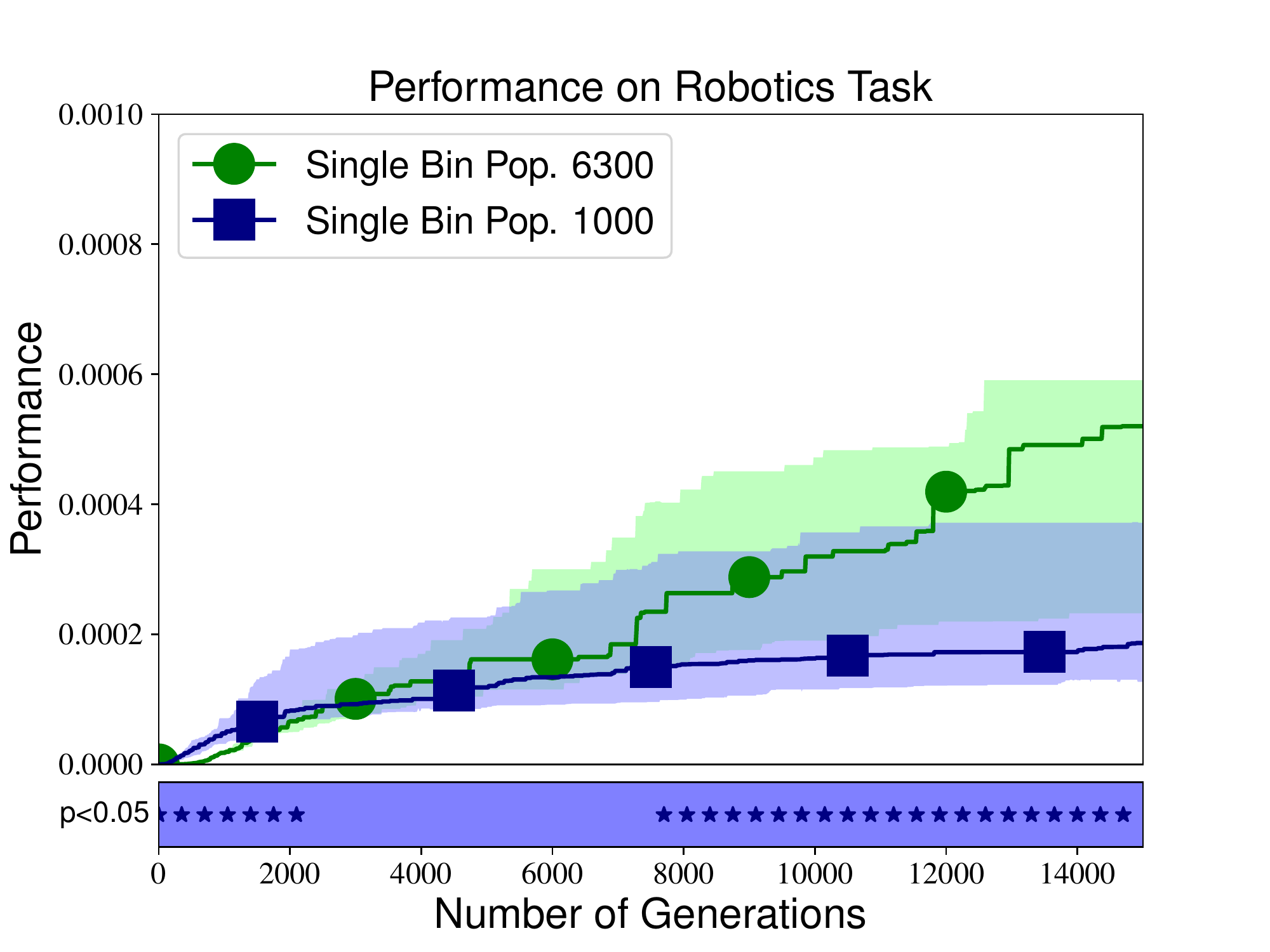}
\caption{\textbf{In general, the NSGA-II based control treatments perform slightly better with a larger population size.} In the legend, CT NSGA-II stands for Combined-Target NSGA-II. The NSGA-II, Combined-Target NSGA-II, and Single Bin treatments all perform significantly better when their population is increased from \numprint{1000} (same as the number of offspring created at each generation) to \numprint{6300} (same as CMOEA on the six-tasks problem). The only exception is Combined-Target NSGA-II combined with a behavioral diversity objective, where a population size of \numprint{1000} seems to perform better than a population size of \numprint{6300}. Data for each treatment is from 30 independent runs.}
\label{fig:popsize_comp}
\end{figure*}

\begin{figure*}[tbp!]
\centering
\includegraphics[width=0.49\textwidth]{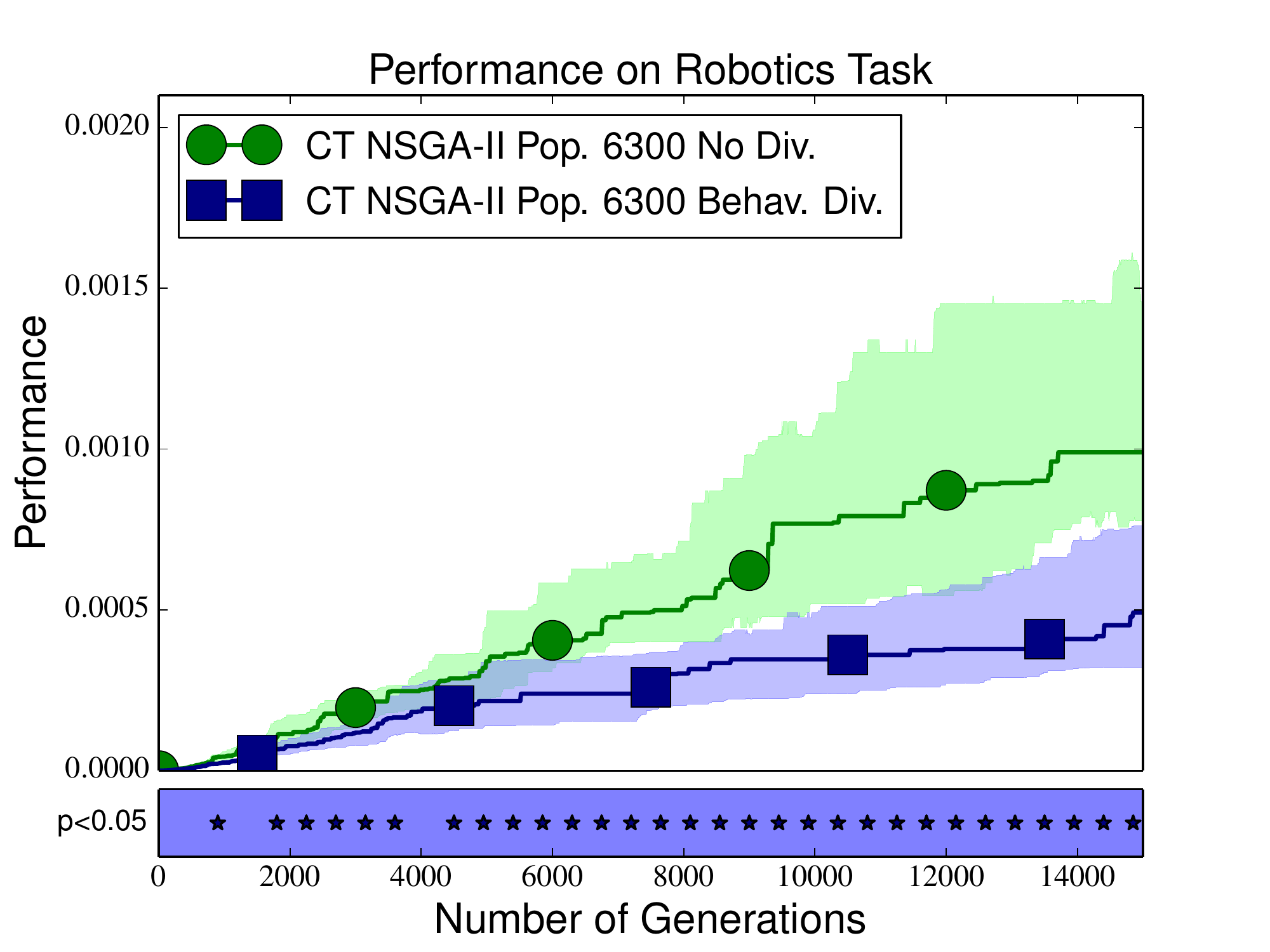}
\includegraphics[width=0.49\textwidth]{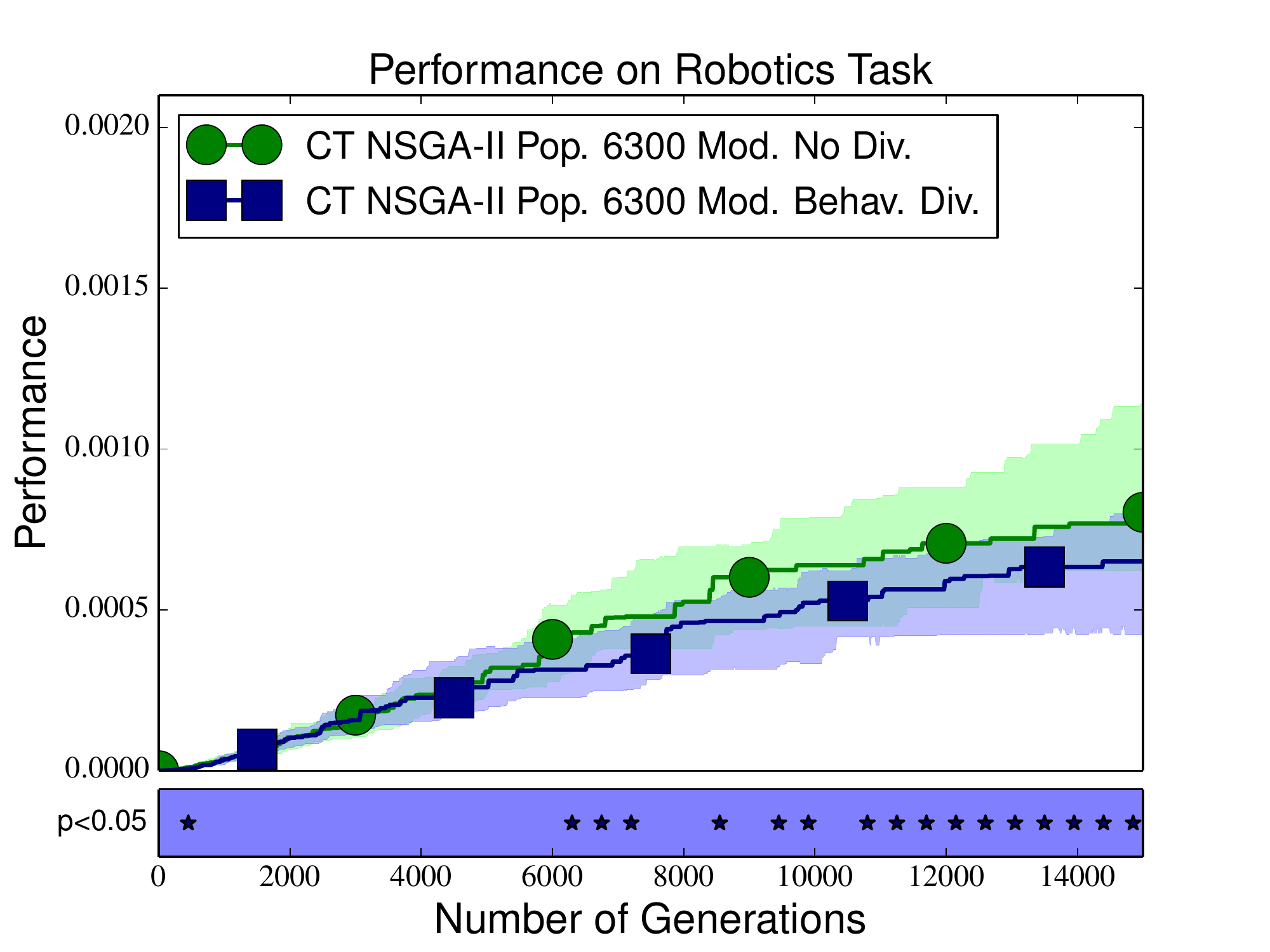}
\includegraphics[width=0.49\textwidth]{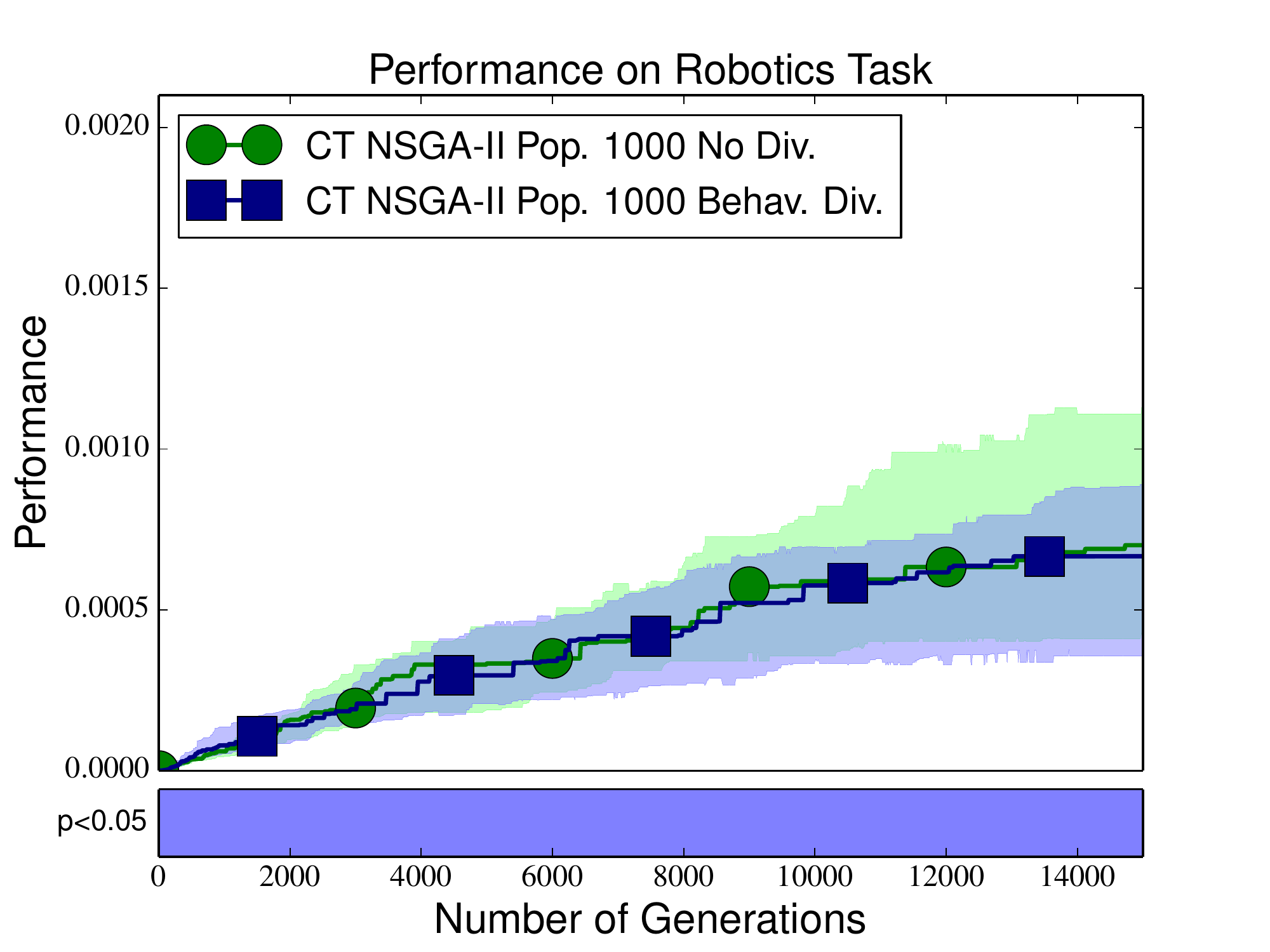}
\includegraphics[width=0.49\textwidth]{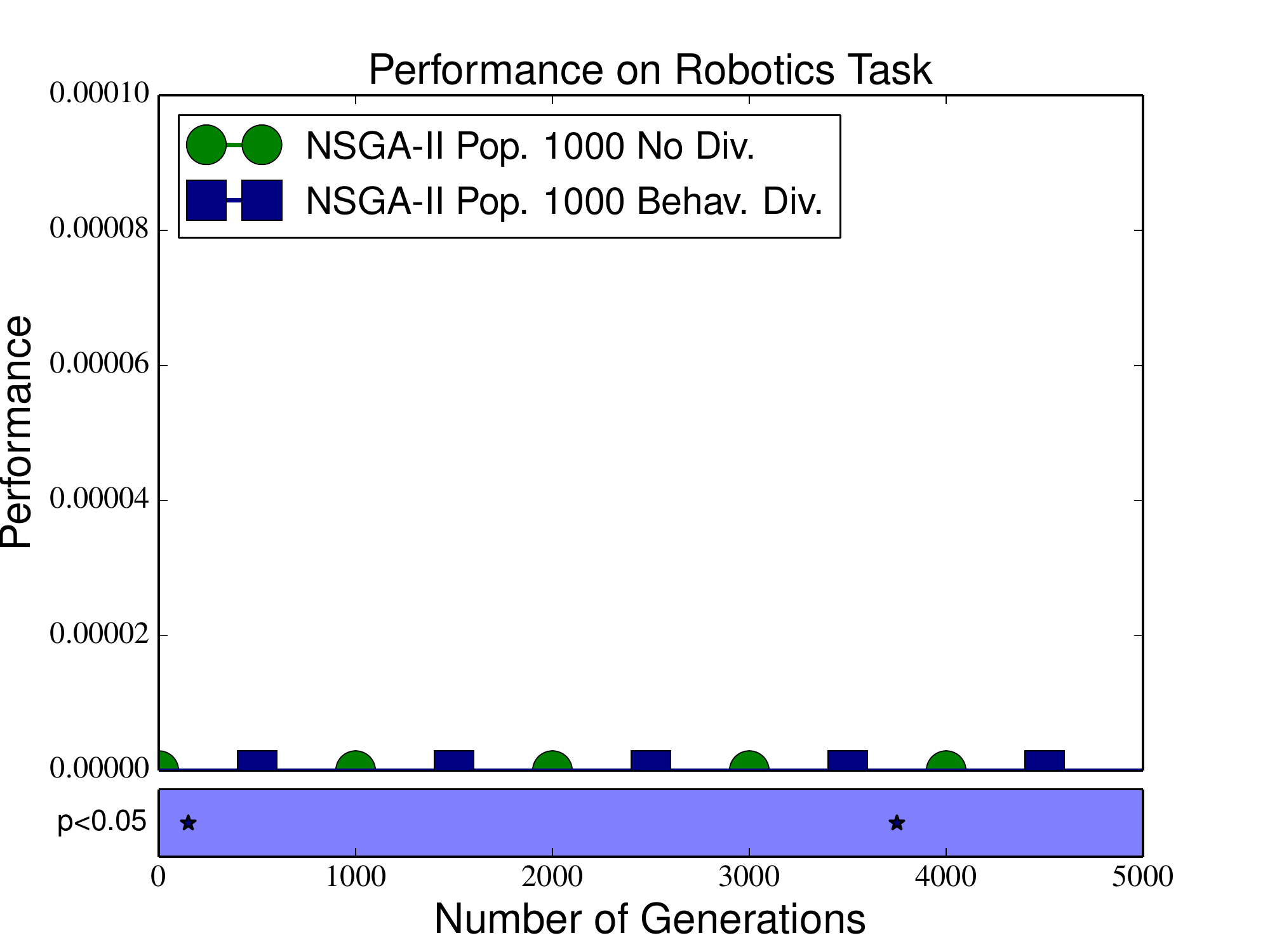}
\caption{\textbf{Behavioral diversity either has no significant effect or actively hurts the performance of the many-objective NSGA-II control treatments.} In the legend, CT NSGA-II stands for Combined-Target NSGA-II. For Combined-Target NSGA-II with a population size of \numprint{6300}, adding behavioral diversity significantly reduces the performance on the six-tasks robotics problem. This effect is also present when genotypic and phenotypic modularity (the CT NSGA-II Pop. \numprint{6300} Mod. treatments) are added as additional objectives. For Combined-Target NSGA-II and regular NSGA-II with a population size of \numprint{1000}, the addition of behavioral diversity as an objective has no significant effect. Data for each treatment is from 30 independent runs.}
\label{fig:nsga_div}
\end{figure*}

\subsection{NSGA-II Population size}
\label{sec:pop_size}

In many evolutionary algorithms, including NSGA-II, the population size defines not just the number of individuals maintained by the algorithm at any point in time, but also the number of new individuals produced at every generation.
This is not a practical choice for CMOEA, however, because the size of the population is a function of the bin size and the number of objectives to be optimized, which is often too large to be a feasible choice for the number of new individuals to create at every generation.
As such, we have similarly decoupled the population size from the number of individuals created at each generation in our control treatments,
and allowed our control treatments to have a population size that is larger than the number of offspring created at each generation.
In preliminary experiments, we verified that choosing a larger population size did not have unintended negative effects on our NSGA-II based control treatments. We did not repeat these experiments for NSGA-III or any of the Lexicase Selection variants to reduce the overall computational cost of our experiments, but we have no reason to believe that these algorithms would respond dramatically differently from NSGA-II.

Increasing the population size in NSGA-II has two potential effects. 
First, it increases the number of Pareto-optimal individuals that are maintained, thus providing a better estimate of the Pareto-front at every generation. 
Based on this observation, a larger population size could increase the effectiveness of NSGA-II, 
as a better estimate of the Pareto-front implies a more diverse set of individuals that can serve as the stepping stones towards optimal solutions.
However, a larger population size also means that sub-optimal individuals have a higher chance of surviving in the population, 
thus diluting the pool of parents that supply offspring for the next generation.
Including more sub-optimal parents in the population can slow down the evolutionary process, and thus hurt the performance of NSGA-II.

Given the large number of objectives presented in our research, we hypothesized that it would require a large population size before non Pareto-optimal individuals would start dominating the population, and thus that increasing the population size should increase NSGA-II's performance on our problems.
This hypothesis was confirmed by our preliminary experiments, which show that most control treatments with a population size of \numprint{6300} outperform the same control treatment with a population size of \numprint{1000} on the six-tasks robotics problem (Fig.~\ref{fig:popsize_comp}). 
The one exception is when Combined-Target NSGA-II is combined with behavioral diversity, as Combined-Target NSGA-II Behav. Div. with a population size of \numprint{1000} outperforms Combined-Target NSGA-II Behav. Div. with a population size of \numprint{6300}. 
The reason for this effect is unclear but, because behavioral diversity tends to reduce the effectiveness of Combined-Target NSGA-II (Sec.~\ref{sec:behav_div}), we decided not to include behavioral diversity in our NSGA-II controls, meaning that this effect was not important for the results presented in this paper. 
In light of these results, the population size for all control treatments presented in the main paper was set to be equal to the population size of CMOEA.

\subsection{NSGA-II Behavioral diversity}
\label{sec:behav_div}

Previous work has demonstrated that adding behavioral diversity as an additional objective to NSGA-II can greatly increase its performance on problems with one or two objectives~\cite{mouret2011novelty}. However, it was unclear whether these benefits would also be present on problems with six or more objectives. While a behavioral diversity objective could aid the evolutionary process on a many-objective problem by increasing the diversity of the population, and thus increasing the number of potential stepping stones, it is also possible that adding yet another dimension to the already high-dimensional space of a many-objective problem would only hurt the performance of the algorithm. To examine whether behavioral diversity would increase the performance of NSGA-II on a many-objective problem, we ran preliminary experiments with behavioral diversity added to different variants of NSGA-II on the six-tasks robotics problem.

The results show that adding behavioral diversity significantly hurts the performance of Combined-Target NSGA-II with a population size of \numprint{6300}, 
both with and without modularity objectives (Fig.~\ref{fig:nsga_div}). 
Furthermore, behavioral diversity has no observable effect on regular NSGA-II or Combined-Target NSGA-II with a population size of \numprint{1000}. These results suggest that behavioral diversity does not increase the performance of NSGA-II when applied to many-objective optimization problems. As such, the NSGA-II based controls presented in the main paper are implemented without behavioral diversity.

\begin{figure*}[p!]
\centering
\includegraphics[height=5cm]{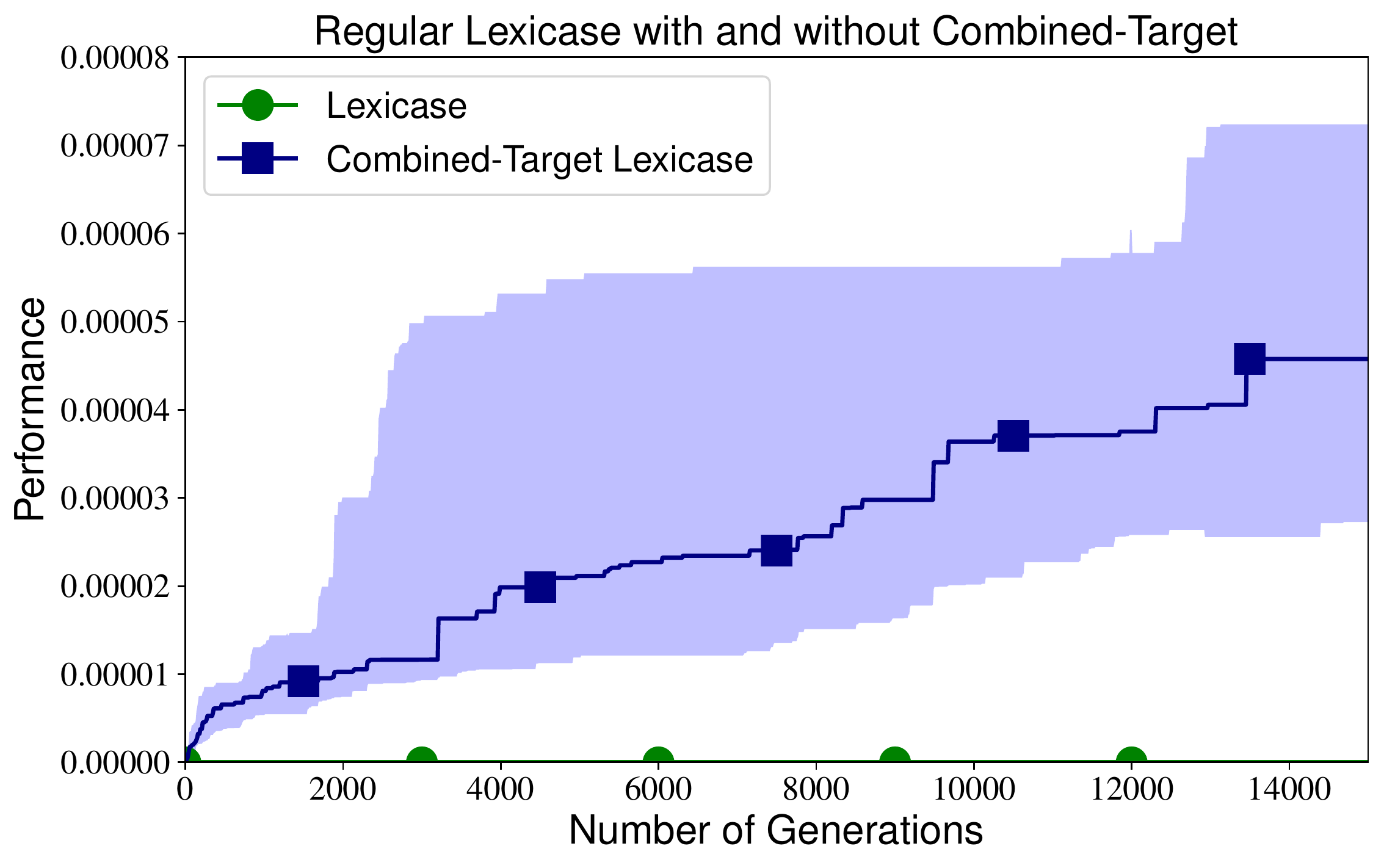}
\includegraphics[height=5cm]{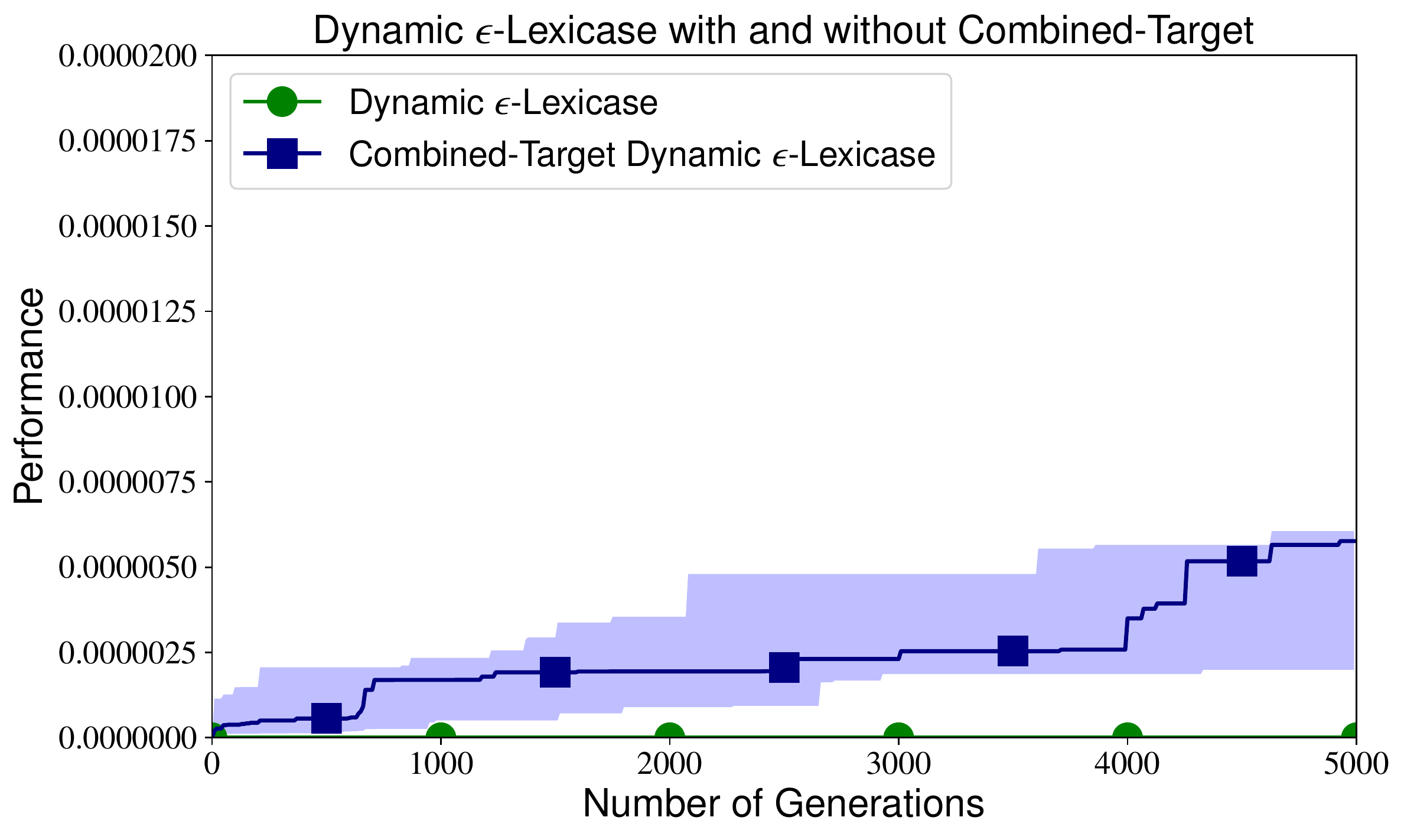}
\includegraphics[height=5cm]{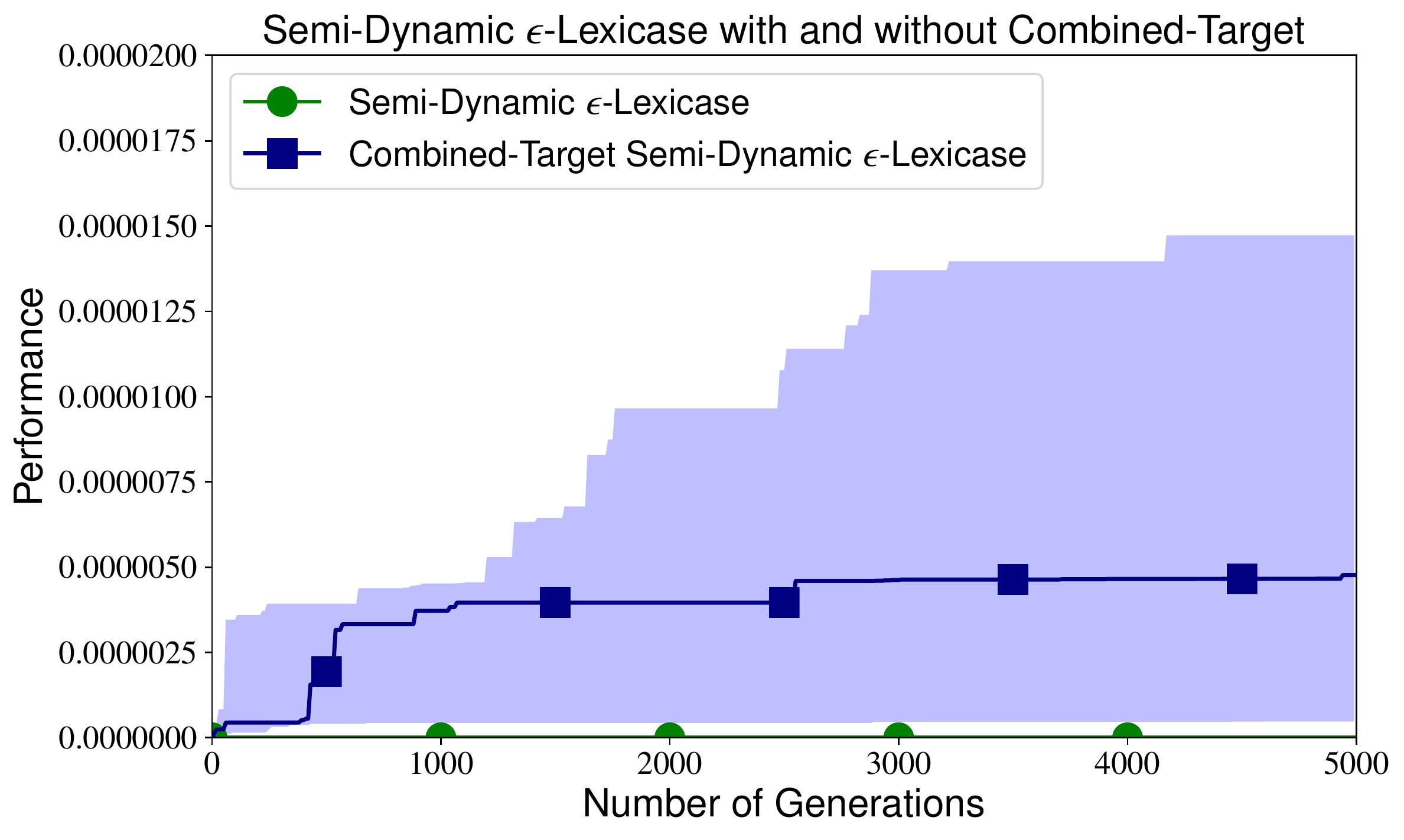}
\includegraphics[height=5cm]{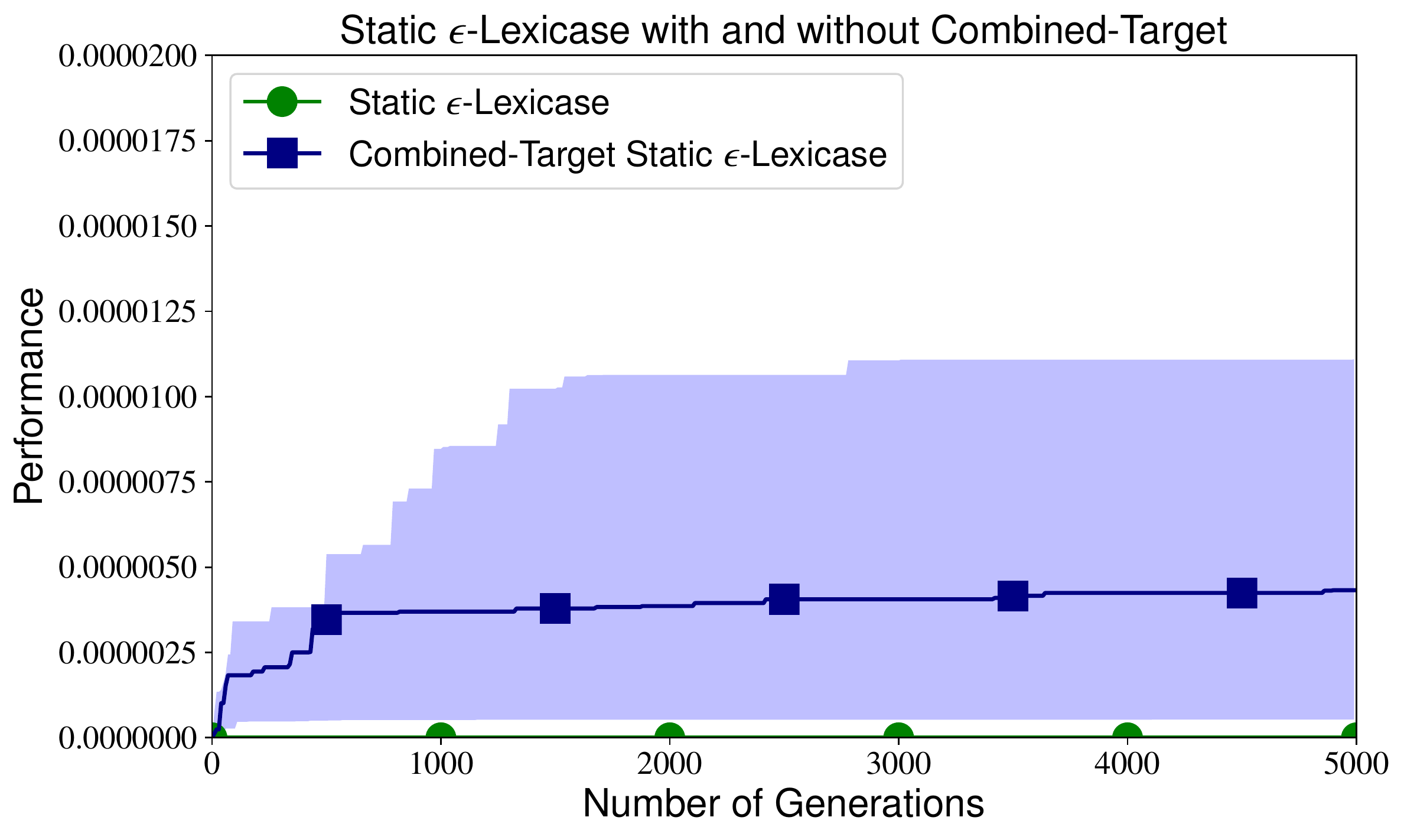}
\caption{\textbf{Adding the Combined-Target objective to Lexicase Selection improves performance for all variants that we tested.} Lexicase Selection variants without the Combined-Target objective all maintain a score of 0 for \numprint{5000} generations. We performed 5 separate runs for each treatment and we do not show significance indicators as those are not informative with only 5 samples.}
\label{fig:elexicase_combined_target}
\end{figure*}

\begin{figure*}[p!]
\centering
\includegraphics[height=5cm]{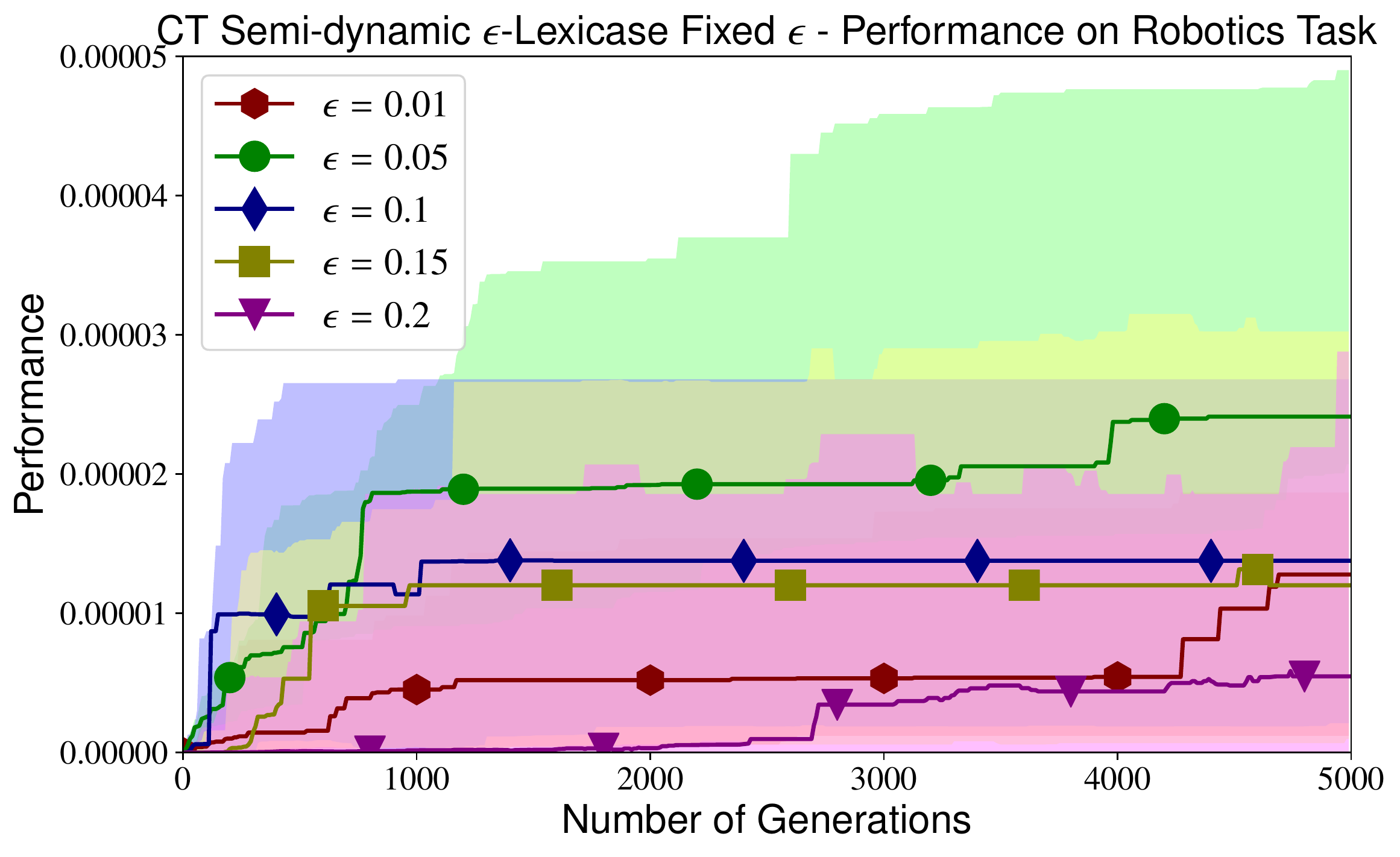}
\includegraphics[height=5cm]{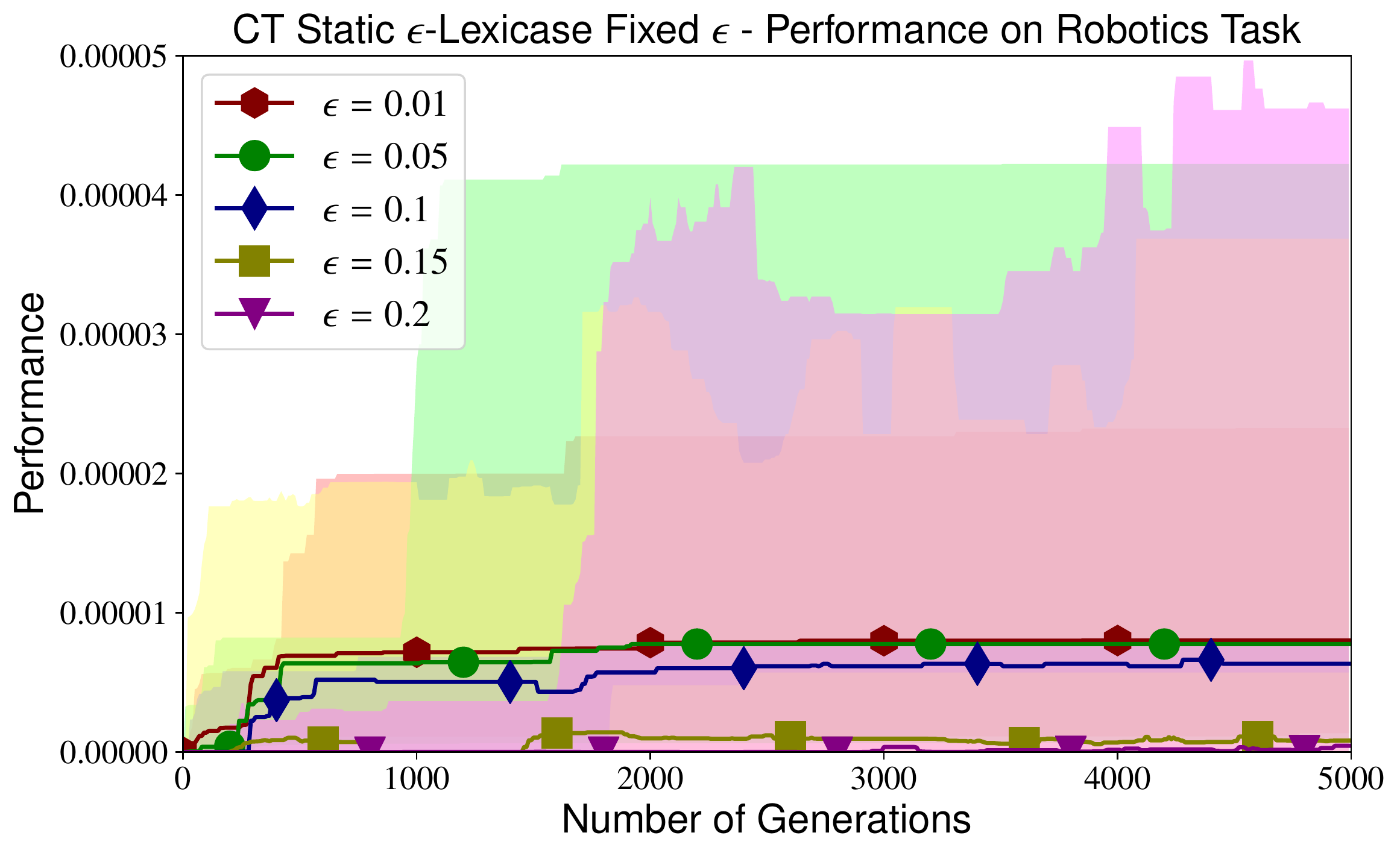}
\includegraphics[height=5cm]{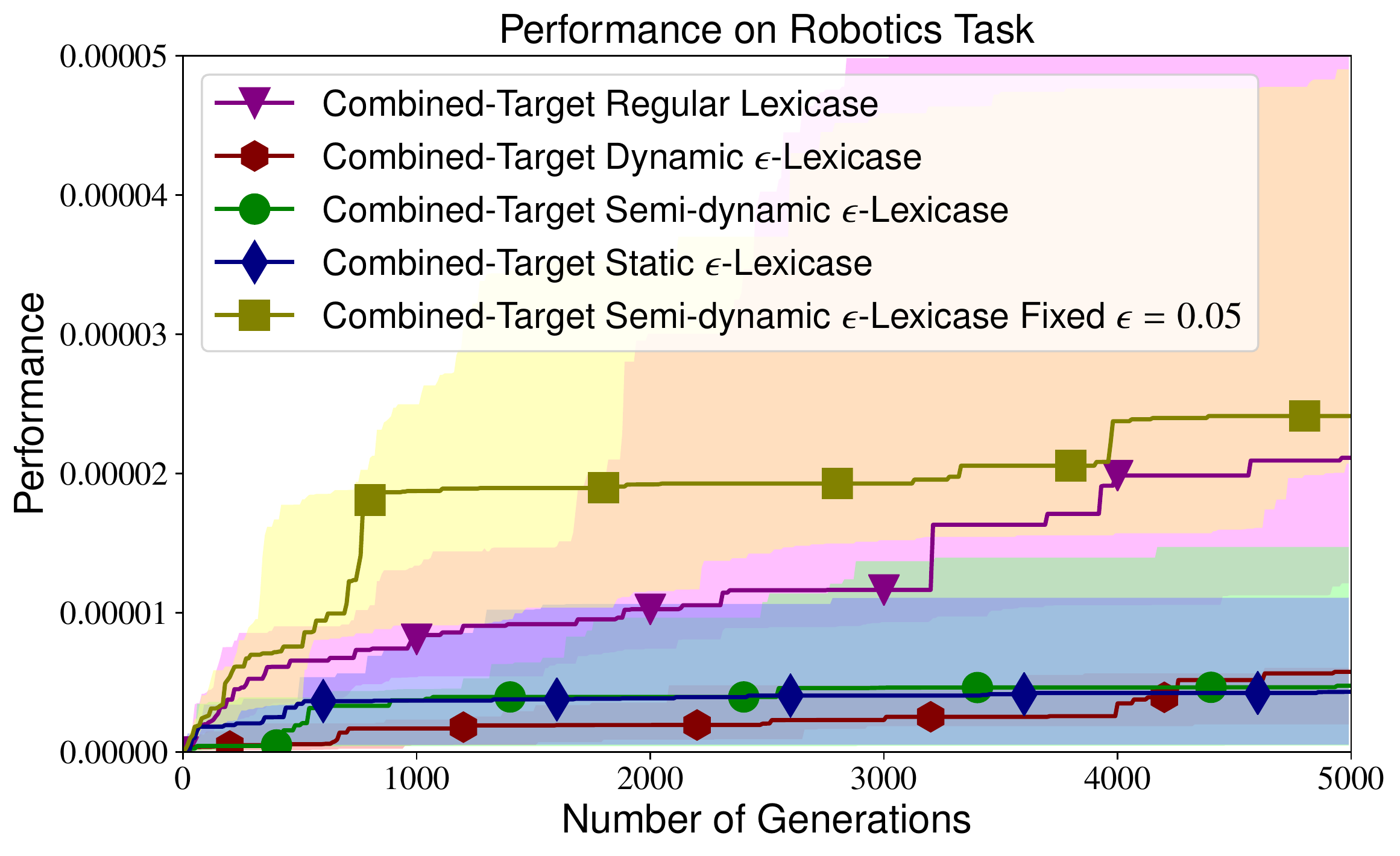}
\caption{\textbf{On the robot locomotion problem, Combined-Target Semi-dynamic $\epsilon$-Lexicase Selection with a fixed of $\epsilon=0.05$ outperforms other Combined-target Lexicase Selection variants.} Interestingly, regular Combined-Target Lexicase Selection performs better than Combined-Target $\epsilon$-Lexicase Selection variants that determine the $\epsilon$ automatically based on the MAD metric, probably because the MAD metric only works properly on objectives where most individuals in the population obtain a non-zero score. We performed 5 separate runs for each treatment and we do not show significance indicators as those are not informative with only 5 samples.}
\label{fig:elexicase_fixed_e_sweep}
\end{figure*}

\begin{figure*}[tb!]
\centering
\includegraphics[width=0.45\textwidth]{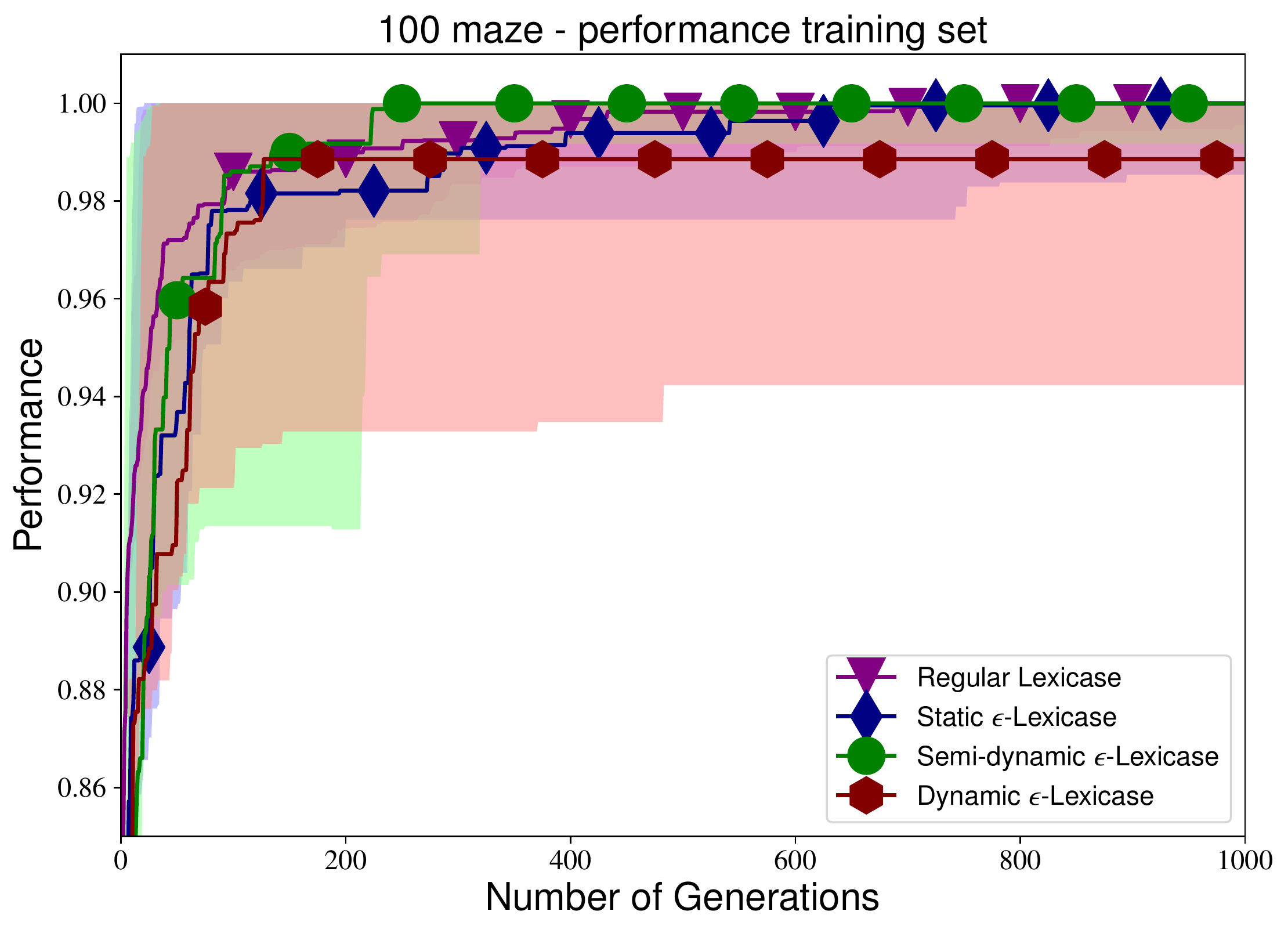}
\includegraphics[width=0.45\textwidth]{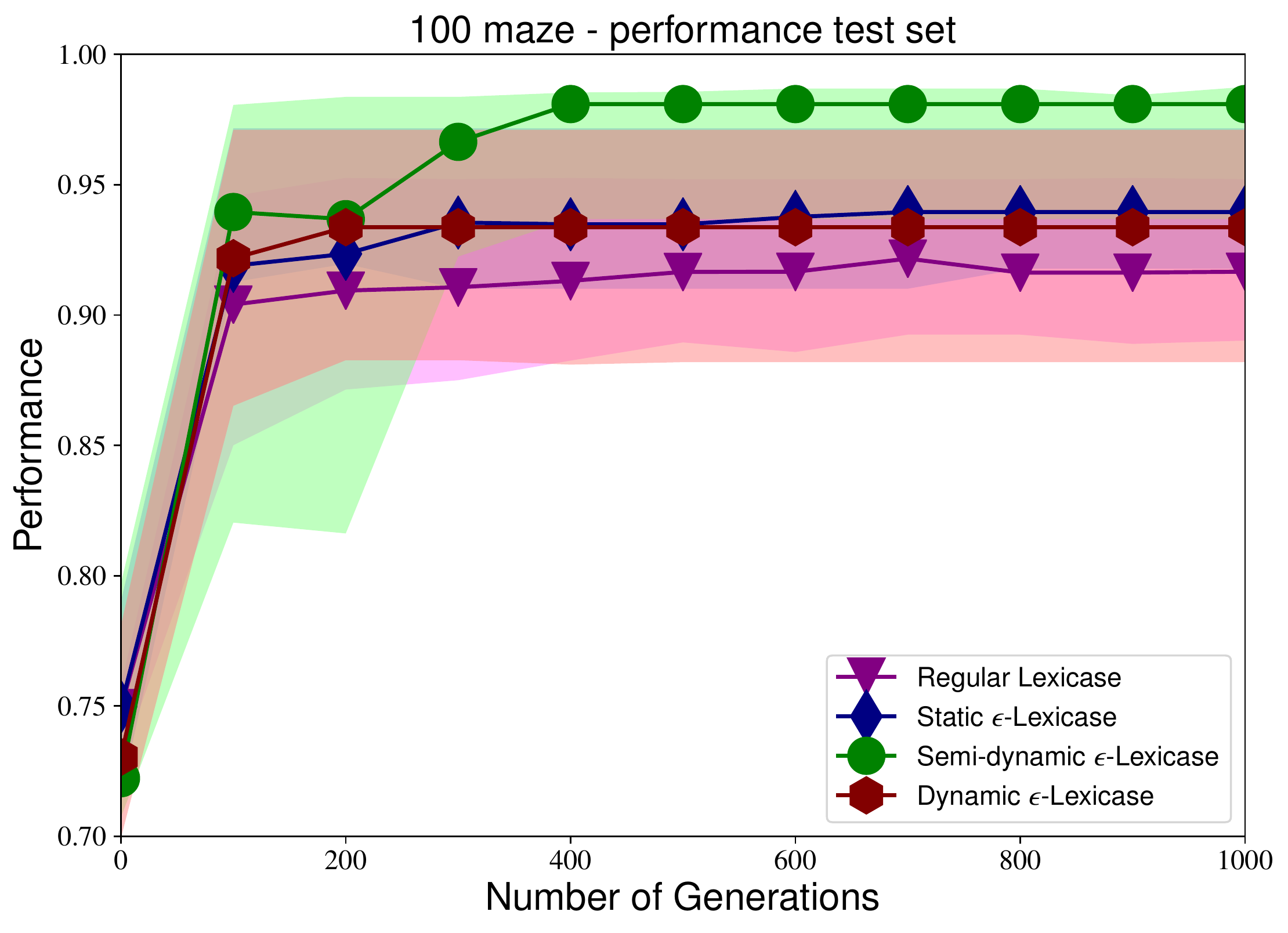}
\includegraphics[width=0.45\textwidth]{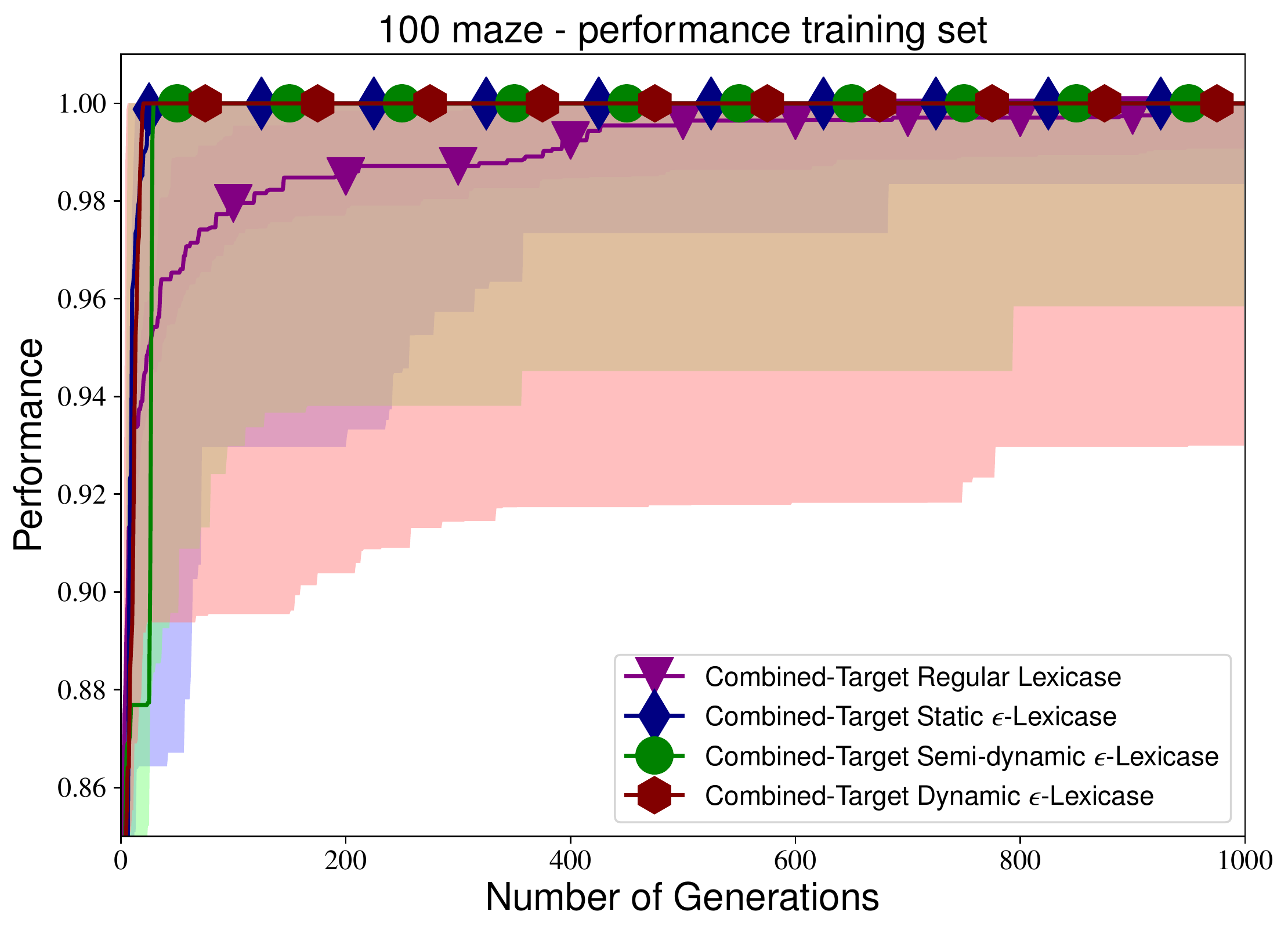}
\includegraphics[width=0.45\textwidth]{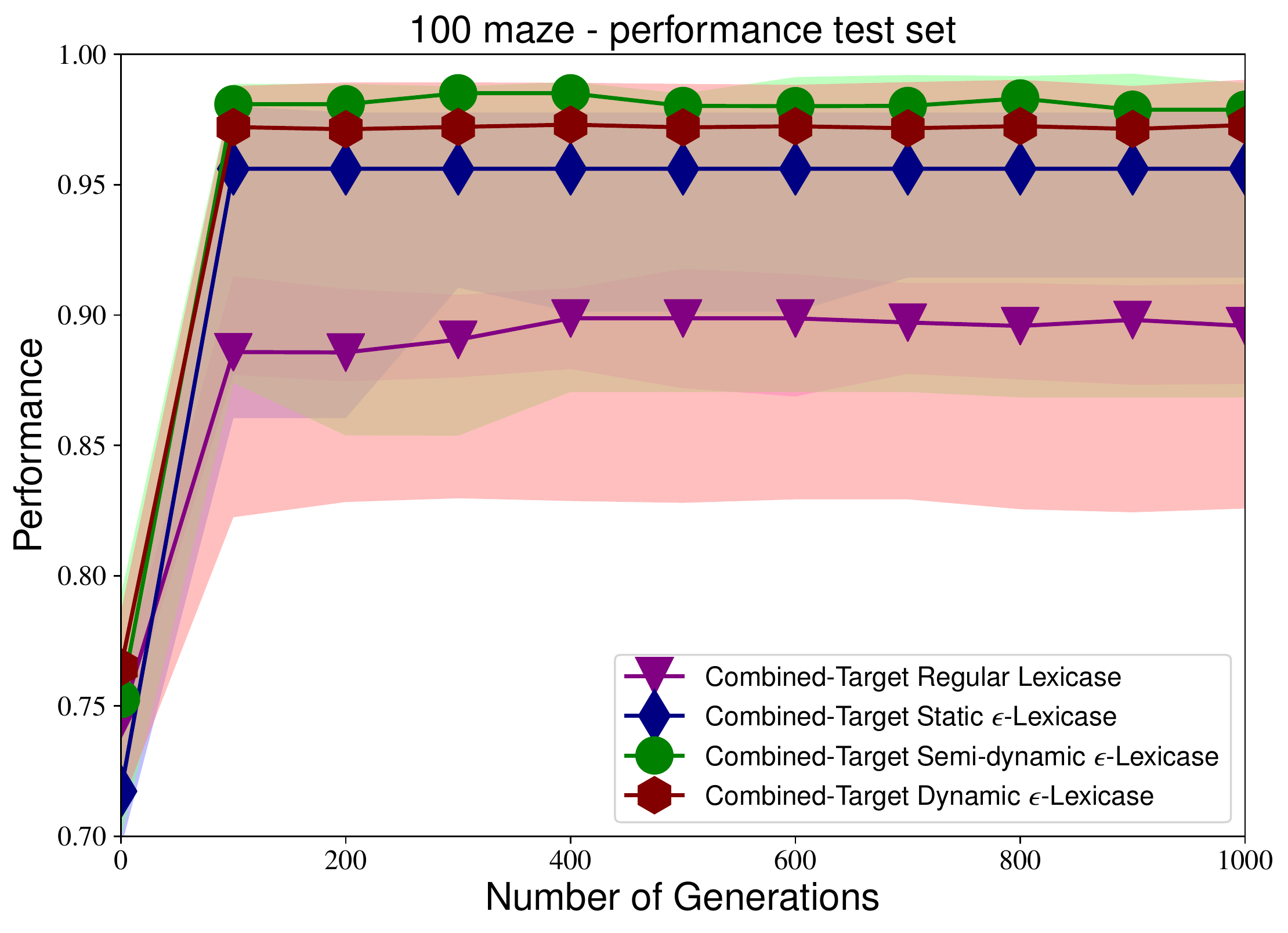}
\caption{\textbf{On the maze navigation problem, all variants of Lexicase Selection perform well.} The Combined-Target $\epsilon$-Lexicase Selection treatments seem to converge slightly faster than regular Lexicase Selection or any of the variants without the Combined-Target objective. From among the treatments that seemed to perform best, we arbitrarily chose Combined-Target dynamic $\epsilon$-Lexicase Selection for the experiments presented in the main paper. We performed 5 separate runs for each treatment and we do not show significance indicators as those are not informative with only 5 samples.}
\label{fig:elexicase_maze}
\end{figure*}

\subsection{Lexicase Selection variants performance}
\label{sec:lexicase_variants}

Because there exist many different versions of the Lexicase Selection algorithm (i.e. all variants of $\epsilon$-Lexicase Selection, main paper Sec.~\ref{sec:e_lexicase_selection}), and because it was not obvious which variant would perform best on the problems presented in the main paper, we performed preliminary experiments to decide which Lexicase Selection variant we would compare against. Because these experiments are computationally expensive, and because there were many different parameters to test, we ran only five replicates for each treatment. While this is insufficient to obtain statistical certainty about which variant performs best on each problem, they present us with a rough estimate of the performance of the different variants, and it prevents us from choosing a variant that performs pathologically poorly on any of our test problems. After determining the most promising variant, we would then run a full-scale experiment with 30 replicates, the results of which are reported in the main paper.

Because it is a computationally expensive problem, we limited the preliminary experiments on the simulated robot locomotion task to \numprint{5000} generations. Because the performance of $\epsilon$-Lexicase Selection seemed pathologically poor in our initial experiments, we started by examining the effect of the Combined-Target objective on the different Lexicase Selection variants. The results show that, without the Combined-Target objective, no variant is able to obtain a score greater than 0 within \numprint{5000} generations (Fig.~\ref{fig:elexicase_combined_target}). However, with the Combined-Target objective, performance values increase properly. As such, we added the Combined-Target objective to all Lexicase Selection variants in subsequent experiments.

The reason why non of the Lexicase Selection variants is able to obtain a score greater than 0 without the Combined-Target objective is likely a combination of factors. First, regular Lexicase Selection is strongly biased towards selecting only individuals that perform best on one of the objectives, because ties between different individuals are unlikely in the space of real numbers. Second, while the different variants of $\epsilon$-Lexicase Selection should have been able to avoid this issue with a proper $\epsilon$, when $\epsilon$ is determined automatically it is set based on the Median Absolute Deviation (MAD) metric, which becomes 0 when the majority of values is 0, which is the case for many of the objectives in the multimodal locomotion task. With a MAD metric of 0, $\epsilon$-Lexicase Selection reverts back to behaving like regular Lexicase Selection, and thus performs poorly. We did not try static $\epsilon$ variants without the Combined-Target objective, but we hypothesize that they would work slightly better than any of the automatic methods. A different measure of deviation could potentially have worked better, but we believe that such experiments are outside the scope of this paper. 

When we compare different variants of Lexicase Selection with the Combined-Target objective against each other, we see that a static $\epsilon$ of 0.05 outperforms all variants which automatically determine $\epsilon$ (Fig.~\ref{fig:elexicase_fixed_e_sweep}). Once again, this is likely because the MAD metric becomes 0 when the majority of individuals score 0 on a particular objective. Interestingly, the second best performing variant is regular Lexicase Selection. The main difference between regular Lexicase Selection and the Lexicase variants it outperforms is that regular Lexicase Selection will have an $\epsilon$ of 0 on all objectives, while the variants it outperforms all attempt to determine $\epsilon$ automatically based on the MAD score, meaning their $\epsilon$ will only be 0 on objectives on which the majority of the population receives a score of 0, i.e. on the hard objectives. As a result, these variants that determine $\epsilon$ automatically are more likely to preserve diverse individuals that obtain some performance on the easy objectives, but they will be elitist on the hard objectives. This skewed selection pressure towards easy objectives is likely why these variants of Lexicase Selection appear to perform worse than regular Lexicase Selection.

On the simulated robot maze-navigation task, all Lexicase Selection variants seem to perform well, including regular Lexicase Selection (Fig.~\ref{fig:elexicase_maze}). This is likely because of the way the maze navigation task is presented (i.e. each maze is its own test case), which closely resembles the problems that Lexicase Selection was originally designed to solve. That said, in combination with the Combined-Target objective, the $\epsilon$-Lexicase Selection variants seem to outperform regular Lexicase Selection in terms of how fast they converge to near optimal performance. This is likely because, $\epsilon$-Lexicase Selection is able to maintain a greater diversity of individuals during the first few generations, when most mazes are only partially solved, and thus when most objective scores are still real-valued numbers different from 1. Given that there is little indication that any one of the Combined-Target $\epsilon$-Lexicase Selection variants performs better than the others, we arbitrarily chose dynamic Combined-Target $\epsilon$-Lexicase Selection as the variant to compare against,  though any of the other variants would probably have resulted in similar conclusions. Because the variants in which $\epsilon$ is automatically determined performed so well, we did not do a sweep over fixed values for $\epsilon$.

\begin{figure*}[tb!]
\centering
\includegraphics[width=0.49\textwidth]{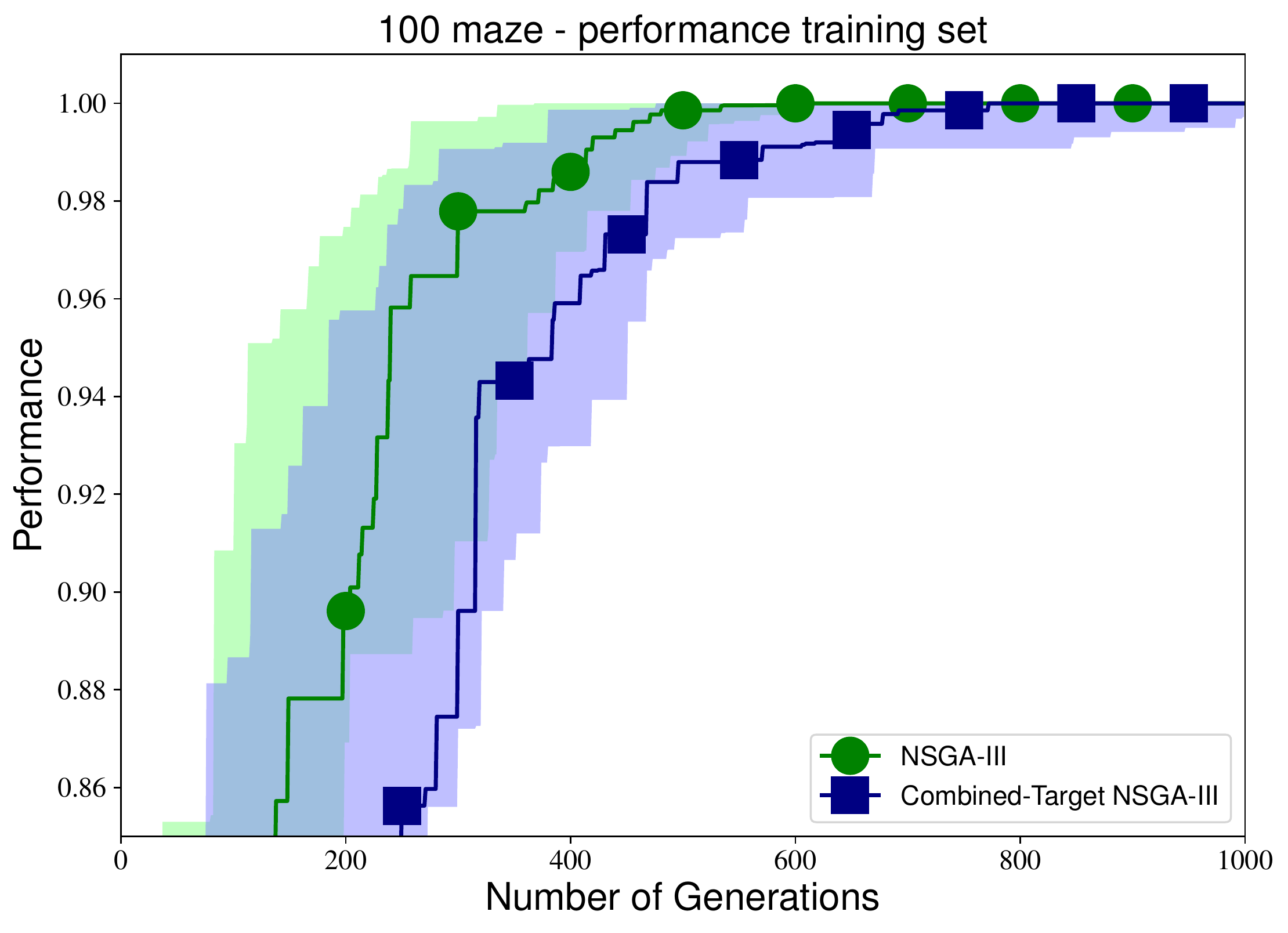}
\includegraphics[width=0.49\textwidth]{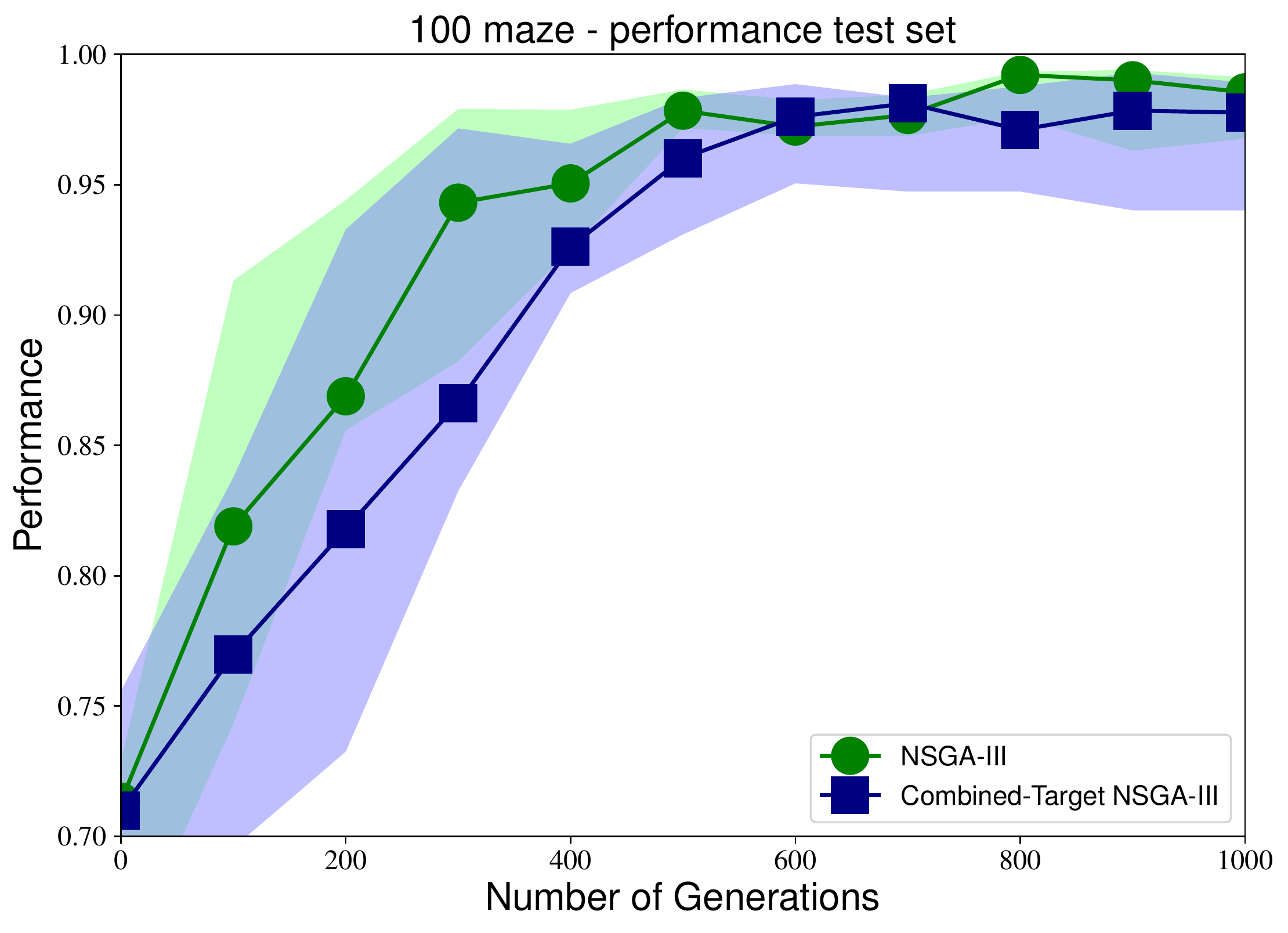}
\caption{\textbf{On the maze domain, adding the Combined-Target objective to NSGA-III seems to slightly degrade its performance.} For this reason, plain NSGA-III has been included in the experiments presented in the main paper. We performed 5 separate runs for each treatment and we do not show significance indicators as those are not informative with only 5 samples.}
\label{fig:nsga_combined_target}
\end{figure*}

\subsection{NSGA-III Combined-Target objective}
\label{sec:nsga_iii_combined_target}

On both the simulated multimodal robot locomotion domain and on the simulated robot maze navigation domain we tested whether NSGA-III would benefit from the Combined-Target objective in the same way as NSGA-II. The results on the multimodal locomotion domain are presented in the main paper (main paper Fig.~\ref{fig:naive}), and suggest that NSGA-III can benefit from the Combined-Target objective, but not as much as NSGA-II. On the maze domain, the difference is similarly small, and while it seems that the Combined-Target objective may slightly hurt the performance of NSGA-III, the sample size of five different seeds is not large enough to make any firm conclusions (Fig.~\ref{fig:nsga_combined_target}). That said, based on these results, we compare CMOEA against NSGA-III without the Combined-Target objective on the maze domain.

\subsection{NSGA-III Normalization}
\label{sec:nsga_iii_normalization}

The paper that introduces NSGA-III proposes an automatic method for normalizing all objectives based on what they refer to as \emph{extreme points} \cite{deb2013evolutionary}. Before determining the extreme points they ensure that the smallest value on each objective is 0 by finding the smallest value on each objective and then subtracting it for all individuals. From there, the extreme point for a particular axis $a$ is defined as the point $x$ that minimizes the following equation:

\begin{equation}
ASF( \vec{x}, \vec{w}^a) = \underset{i=1}{\operatorname{max}} (x_i/w^a_i)
\end{equation}

Where $\vec{w^a}$ is a unit vector pointing in the direction of axis $a$ with all values of 0 replaced with the small number of $10^{-6}$ (after which it is no longer a unit vector). These points are then supposed to define a hyper-plane that intersects all axis such that the intercepts of this hyper-plane with each axis can be used to normalize all objectives. The problem with this method is that, especially when the number of objectives is large, it is not unlikely that the extreme points for different axis is actually the same point. For example, given the points $[2, 2, 0]$, $[1, 0, 1]$, and $[0, 2, 2]$, the point $[1, 0, 1]$ is the extreme point for all three axis, and the corresponding hyper-plane is undefined. The NSGA-III paper~\cite{deb2013evolutionary}, does not suggest a solution for this problem.

\begin{figure}[tb!]
\centering
\includegraphics[width=0.49\textwidth]{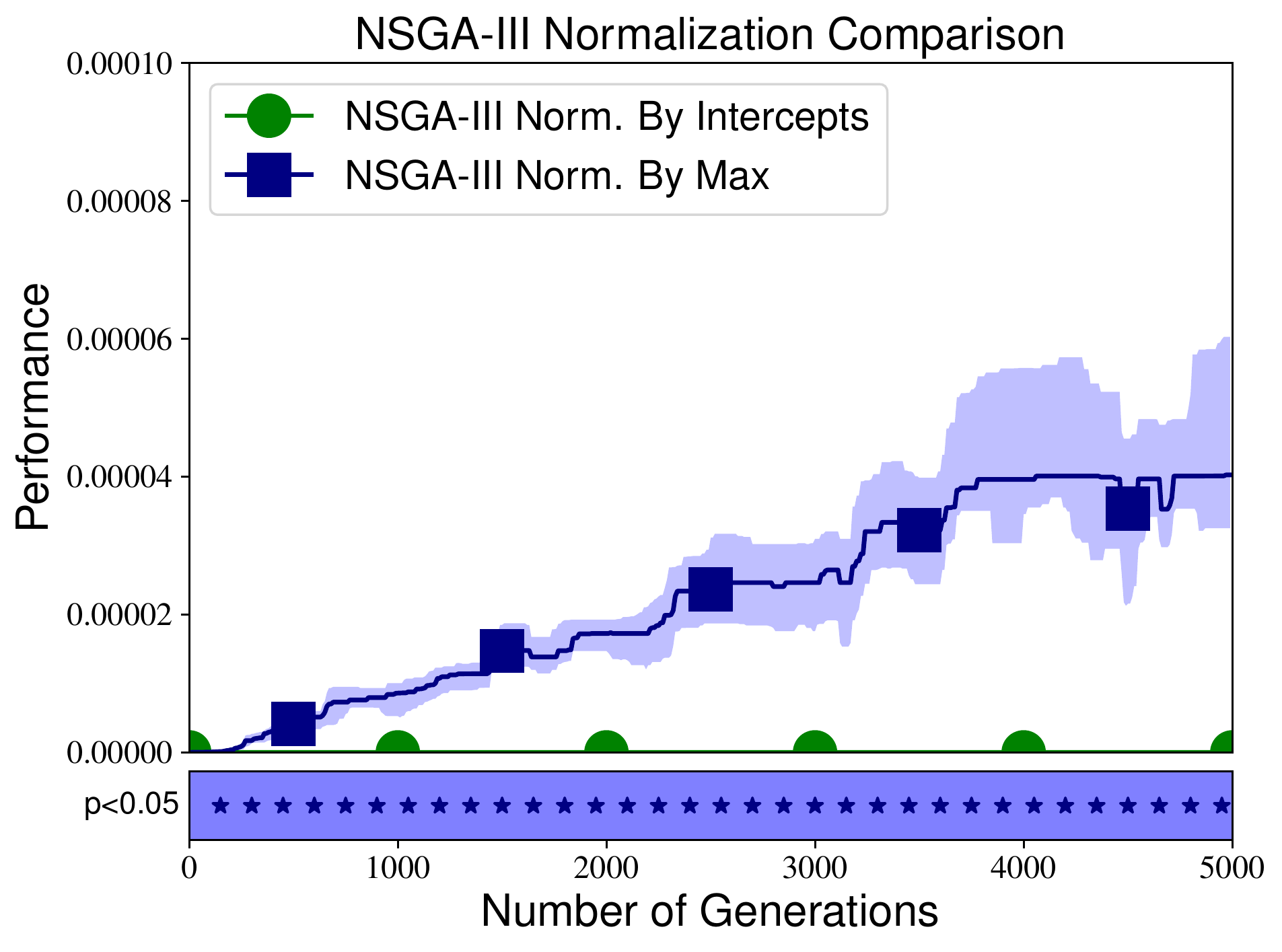}
\caption{\textbf{On the multimodal locomotion problem, our backup of normalizing by dividing by the maximum on each objective (after subtracting the minimum), outperforms the intercept-based method described in~\cite{deb2013evolutionary} with this backup.} The poor performance of the intercept method is likely because the intercepts are frequently, but not always, undefined, meaning the algorithm will constantly change its normalization procedure, thus leading to unpredictable and probably disruptive selection pressures. We performed 30 separate runs for each treatment.}
\label{fig:nsga_iii_intercepts}
\end{figure}

We resolved this problem by defining a backup normalization procedure which subtracts the smallest value on each objective among all individuals in the population from that objective for all individuals, and then divides by the largest value after subtraction (i.e. standard, per objective normalization, based on the minimum and maximum values found in the population), which we use whenever the intercepts are undefined. This raises the question whether the intercept-based normalization with the backup actually provides any benefits relative to always applying the backup normalization procedure. Our preliminary experiments demonstrate that always normalizing by dividing by the maximum value substantially outperforms the intercept-based method with backup on the multimodal locomotion task (Fig.~\ref{fig:nsga_iii_intercepts}). This is likely because the intercept-based method with backup constantly changes its normalization procedure, as the hyper-plane will switch between being defined and being undefined, thus constantly changing the selection pressures in a disruptive way. Based on these results, the backup normalization procedure was used as the default normalization procedure in the main paper.

\end{document}